%% file: main.tex
\documentclass{article}

\PassOptionsToPackage{numbers, compress}{natbib}
\usepackage[preprint]{neurips_2026}
\usepackage{xspace}
\usepackage[table]{xcolor}
\usepackage{etoc}

\usepackage{xcolor}
\usepackage{listings}

\lstdefinestyle{promptstyle}{
    basicstyle=\ttfamily\scriptsize,
    backgroundcolor=\color{gray!5},
    frame=single,
    rulecolor=\color{gray!35},
    breaklines=true,
    breakatwhitespace=false,
    columns=fullflexible,
    keepspaces=true,
    showstringspaces=false,
    captionpos=b,
    mathescape=true,
    tabsize=2
}

\newcommand{\diffdown}[1]{\textcolor{red!75!black}{\scriptsize$_{-#1\downarrow}$}}
\newcommand{\diffup}[1]{\textcolor{green!55!black}{\scriptsize$_{+#1\uparrow}$}}
\usepackage{graphicx}
\usepackage{siunitx}
\usepackage{bm}
\usepackage{tikz}
\usetikzlibrary{calc}
\usetikzlibrary{positioning}
\usetikzlibrary{arrows.meta}
\usepackage{amssymb}
\usepackage{enumitem}
\usepackage{algorithm}
\usepackage{amsthm}
\usepackage{makecell}
\usepackage{multirow}

\usepackage{algpseudocode}
\theoremstyle{plain}

\usepackage{natbib}
\theoremstyle{definition}
\usepackage{amsthm}
\usepackage{subcaption}

\newcommand{\our}{VICO\xspace}

\usepackage[utf8]{inputenc} 
\usepackage[T1]{fontenc}    
\usepackage{url}            
\usepackage{booktabs}       
\usepackage{amsfonts}       
\usepackage{nicefrac}       
\usepackage{microtype}      
\usepackage{xcolor}         

\usepackage{multirow}
\usepackage{float}
\usepackage[utf8]{inputenc} 
\usepackage[T1]{fontenc}    
\usepackage{url}            
\usepackage{booktabs}       
\usepackage{amsfonts}       
\usepackage{nicefrac}       
\usepackage{microtype}      
\usepackage{mathtools}
\usepackage{natbib}
\usepackage{xcolor}
\usepackage{url}
\usepackage{amsmath}
\usepackage{amssymb}
\usepackage{amsthm}
\usepackage{graphicx}
\usepackage{nicefrac}
\usepackage{appendix}
\usepackage{enumitem}
\usepackage{xspace}
\usepackage{pifont}
\usepackage{wrapfig,lipsum,booktabs}

\usepackage{amsfonts}

\usepackage{float}  
\usepackage{booktabs}
\usepackage{xcolor}
\usepackage{caption}
\usepackage{tcolorbox}

\usepackage{wrapfig}
\usepackage{caption}
\usepackage{subcaption}
\usepackage{array}
\definecolor{darker}{rgb}{0.15,0.45,0.4}
\definecolor{green}{HTML}{009B55}
\usepackage[colorlinks, urlcolor=darker, citecolor=darker, linkcolor=darker]{hyperref} 
\usepackage{colortbl}

\usepackage{physics}

\theoremstyle{definition}

\theoremstyle{remark}

\numberwithin{equation}{section}

\usepackage{algorithmicx}
\usepackage{algcompatible}
\usepackage[utf8]{inputenc} 
\usepackage[T1]{fontenc}    
\usepackage{hyperref}       
\usepackage{url}      
\usepackage{booktabs}       
\usepackage{amsfonts}       
\usepackage{nicefrac}       
\usepackage{microtype}      
\usepackage{xcolor}         
\usepackage{xcolor}    
\definecolor{lightgreen}{rgb}{0.56,0.93,0.56} 
\usepackage{amsmath}        
\usepackage{cleveref}   
\input{macro}

\title{\our: Visual Environments Co-Evolving for Vision-Language Model Reasoning}

\author{%
      \normalfont
      Meng Lu$^{1,8}$,
      Ligeng Zhu$^{2}$,
      Olivia Xiao$^{3}$,
      Yuchen Zhuang$^{4}$,
      Zihan Wang$^{5}$,
      Kuncheng Wu$^{6}$, \\
      \normalfont
      Bangya Liu$^{5}$,
      Yu Wang$^{7}$,
      Charles Fleming$^{8,*}$,
      Wenqi Shi$^{9,*}$,
      Xuan Wang$^{1,*}$ \\[0.4em]
      $^{1}$Virginia Tech \quad
      $^{2}$NVIDIA \quad
      $^{3}$UChicago \quad
      $^{4}$Georgia Tech \\
      $^{5}$Abaka AI \quad
      $^{6}$UC Berkeley \quad
      $^{7}$UGA \quad
      $^{8}$Cisco \quad
      $^{9}$WashU
}

\begin{document}

\maketitle
\footnotetext[1]{Equal correspondence to: Charles Fleming, Wenqi Shi, Xuan Wang.}
\input{tex/0-absract}
\input{tex/1-intro}

\input{tex/2-related}
\input{tex/3-method}
\input{tex/3-2-method}
\input{tex/4-experiments}
\input{tex/6-conclusion}


\bibliographystyle{abbrv}
\bibliography{refs}

\newpage

\input{tex/5-appendix-draft}

\clearpage
\newpage
\newpage
\input{checklist.tex}






\newpage

\end{document}

%% file: macro.tex
\usepackage{pifont}
\newcommand{\cmark}{\ding{51}}
\newcommand{\xmark}{\ding{55}}

\usepackage{colortbl}

\definecolor{mygreen}{RGB}{202,216,198}
\definecolor{myblue}{RGB}{198,208,219}

\ifdefined\DeclareUnicodeCharacter\else
  \usepackage{newunicodechar}
  \newcommand{\DeclareUnicodeCharacter}[2]{%
    \begingroup\lccode`\~="#1\relax
    \lowercase{\endgroup\newunicodechar{~}}{#2}}
  \newunicodechar{–}{--}
  \newunicodechar{—}{---}
  \newunicodechar{’}{'}
\fi

\DeclareUnicodeCharacter{2016}{\ensuremath{\Vert}} %
\DeclareUnicodeCharacter{2264}{\ensuremath{\leq}}  %
\DeclareUnicodeCharacter{2265}{\ensuremath{\geq}}  %
\DeclareUnicodeCharacter{2026}{\dots}              %
\DeclareUnicodeCharacter{00B2}{\textsuperscript{2}}%
\DeclareUnicodeCharacter{03BB}{\ensuremath{\lambda}} %
\DeclareUnicodeCharacter{03C3}{\ensuremath{\sigma}}  %
\DeclareUnicodeCharacter{221A}{\ensuremath{\sqrt}}

\usepackage[utf8]{inputenc}
\usepackage{textcomp} %

\DeclareUnicodeCharacter{03B1}{\ensuremath{\alpha}}   %
\DeclareUnicodeCharacter{03B2}{\ensuremath{\beta}}    %
\DeclareUnicodeCharacter{03B3}{\ensuremath{\gamma}}   %
\DeclareUnicodeCharacter{03B4}{\ensuremath{\delta}}   %
\DeclareUnicodeCharacter{03B5}{\ensuremath{\epsilon}} %
\DeclareUnicodeCharacter{03B6}{\ensuremath{\zeta}}    %
\DeclareUnicodeCharacter{03B7}{\ensuremath{\eta}}     %
\DeclareUnicodeCharacter{03B8}{\ensuremath{\theta}}   %
\DeclareUnicodeCharacter{03B9}{\ensuremath{\iota}}    %
\DeclareUnicodeCharacter{03BA}{\ensuremath{\kappa}}   %
\DeclareUnicodeCharacter{03BB}{\ensuremath{\lambda}}  %
\DeclareUnicodeCharacter{03BC}{\ensuremath{\mu}}      %
\DeclareUnicodeCharacter{03BD}{\ensuremath{\nu}}      %
\DeclareUnicodeCharacter{03BE}{\ensuremath{\xi}}      %
\DeclareUnicodeCharacter{03BF}{o}                     %
\DeclareUnicodeCharacter{03C0}{\ensuremath{\pi}}      %
\DeclareUnicodeCharacter{03C1}{\ensuremath{\rho}}     %
\DeclareUnicodeCharacter{03C2}{\ensuremath{\varsigma}}%
\DeclareUnicodeCharacter{03C3}{\ensuremath{\sigma}}   %
\DeclareUnicodeCharacter{03C4}{\ensuremath{\tau}}     %
\DeclareUnicodeCharacter{03C5}{\ensuremath{\upsilon}} %
\DeclareUnicodeCharacter{03C6}{\ensuremath{\phi}}     %
\DeclareUnicodeCharacter{03C7}{\ensuremath{\chi}}     %
\DeclareUnicodeCharacter{03C8}{\ensuremath{\psi}}     %
\DeclareUnicodeCharacter{03C9}{\ensuremath{\omega}}   %

\DeclareUnicodeCharacter{0391}{A}                     %
\DeclareUnicodeCharacter{0392}{B}                     %
\DeclareUnicodeCharacter{0393}{\ensuremath{\Gamma}}   %
\DeclareUnicodeCharacter{0394}{\ensuremath{\Delta}}   %
\DeclareUnicodeCharacter{0395}{E}                     %
\DeclareUnicodeCharacter{0396}{Z}                     %
\DeclareUnicodeCharacter{0397}{H}                     %
\DeclareUnicodeCharacter{0398}{\ensuremath{\Theta}}   %
\DeclareUnicodeCharacter{0399}{I}                     %
\DeclareUnicodeCharacter{039A}{K}                     %
\DeclareUnicodeCharacter{039B}{\ensuremath{\Lambda}}  %
\DeclareUnicodeCharacter{039C}{M}                     %
\DeclareUnicodeCharacter{039D}{N}                     %
\DeclareUnicodeCharacter{039E}{\ensuremath{\Xi}}      %
\DeclareUnicodeCharacter{039F}{O}                     %
\DeclareUnicodeCharacter{03A0}{\ensuremath{\Pi}}      %
\DeclareUnicodeCharacter{03A1}{P}                     %
\DeclareUnicodeCharacter{03A3}{\ensuremath{\Sigma}}   %
\DeclareUnicodeCharacter{03A4}{T}                     %
\DeclareUnicodeCharacter{03A5}{\ensuremath{\Upsilon}} %
\DeclareUnicodeCharacter{03A6}{\ensuremath{\Phi}}     %
\DeclareUnicodeCharacter{03A7}{X}                     %
\DeclareUnicodeCharacter{03A8}{\ensuremath{\Psi}}     %
\DeclareUnicodeCharacter{03A9}{\ensuremath{\Omega}}   %

\DeclareUnicodeCharacter{2260}{\ensuremath{\neq}}     %
\DeclareUnicodeCharacter{2261}{\ensuremath{\equiv}}   %
\DeclareUnicodeCharacter{2248}{\ensuremath{\approx}}  %
\DeclareUnicodeCharacter{2243}{\ensuremath{\simeq}}   %
\DeclareUnicodeCharacter{2245}{\ensuremath{\cong}}    %
\DeclareUnicodeCharacter{223C}{\ensuremath{\sim}}     %
\DeclareUnicodeCharacter{221D}{\ensuremath{\propto}}  %
\DeclareUnicodeCharacter{2264}{\ensuremath{\leq}}     %
\DeclareUnicodeCharacter{2265}{\ensuremath{\geq}}     %
\DeclareUnicodeCharacter{2208}{\ensuremath{\in}}      %
\DeclareUnicodeCharacter{2209}{\ensuremath{\notin}}   %

\DeclareUnicodeCharacter{2282}{\ensuremath{\subset}}  %
\DeclareUnicodeCharacter{2283}{\ensuremath{\supset}}  %
\DeclareUnicodeCharacter{2286}{\ensuremath{\subseteq}}%
\DeclareUnicodeCharacter{2287}{\ensuremath{\supseteq}}%
\DeclareUnicodeCharacter{2229}{\ensuremath{\cap}}     %
\DeclareUnicodeCharacter{222A}{\ensuremath{\cup}}     %

\DeclareUnicodeCharacter{2227}{\ensuremath{\wedge}}   %
\DeclareUnicodeCharacter{2228}{\ensuremath{\vee}}     %
\DeclareUnicodeCharacter{22A5}{\ensuremath{\perp}}    %
\DeclareUnicodeCharacter{2234}{\ensuremath{\therefore}} %
\DeclareUnicodeCharacter{2235}{\ensuremath{\because}} %

\DeclareUnicodeCharacter{2192}{\ensuremath{\rightarrow}}      %
\DeclareUnicodeCharacter{2190}{\ensuremath{\leftarrow}}       %
\DeclareUnicodeCharacter{21D2}{\ensuremath{\Rightarrow}}      %
\DeclareUnicodeCharacter{21D0}{\ensuremath{\Leftarrow}}       %
\DeclareUnicodeCharacter{21D4}{\ensuremath{\Leftrightarrow}}  %
\DeclareUnicodeCharacter{21A6}{\ensuremath{\mapsto}}          %

\DeclareUnicodeCharacter{221E}{\ensuremath{\infty}}   %
\DeclareUnicodeCharacter{2202}{\ensuremath{\partial}} %
\DeclareUnicodeCharacter{2207}{\ensuremath{\nabla}}   %
\DeclareUnicodeCharacter{220F}{\ensuremath{\prod}}    %
\DeclareUnicodeCharacter{2211}{\ensuremath{\sum}}     %
\DeclareUnicodeCharacter{222B}{\ensuremath{\int}}     %
\DeclareUnicodeCharacter{2206}{\ensuremath{\Delta}}   %
\DeclareUnicodeCharacter{221A}{\ensuremath{\sqrt{}}}  %

\DeclareUnicodeCharacter{00B1}{\ensuremath{\pm}}      %
\DeclareUnicodeCharacter{00D7}{\ensuremath{\times}}   %
\DeclareUnicodeCharacter{00B7}{\ensuremath{\cdot}}    %
\DeclareUnicodeCharacter{00B0}{\ensuremath{^\circ}}   %

\DeclareUnicodeCharacter{2016}{\ensuremath{\Vert}}    %

\DeclareUnicodeCharacter{2026}{\dots}                 %
\DeclareUnicodeCharacter{2032}{\ensuremath{'}}        %
\DeclareUnicodeCharacter{2033}{\ensuremath{''}}       %

\DeclareUnicodeCharacter{2200}{\ensuremath{\forall}}  %
\DeclareUnicodeCharacter{2203}{\ensuremath{\exists}}  %

\DeclareUnicodeCharacter{00B2}{\textsuperscript{2}}   %
\DeclareUnicodeCharacter{00B3}{\textsuperscript{3}}   %

\DeclareUnicodeCharacter{2212}{\ensuremath{-}}

%% file: tex/0-absract.tex
\begin{abstract}

Reinforcement learning with verifiable rewards (RLVR) has become a standard recipe for post-training vision-language models (VLMs), but it typically assumes a static training environment. 
As the actor improves, fixed tasks drift out of its learning frontier: many become trivial, others remain unsolvable; and the learning signal collapses. 
We argue that \emph{VLM post-training should evolve the visual environment alongside the actor}, not just the actor itself. 
We propose \our{}, a co-evolutionary framework in which an \emph{actor} and an \emph{Environment-as-Rewriter} (EnvRewriter) are trained jointly: the EnvRewriter edits verifiable image-side structures, such as scene graphs, chart tables, or protected region masks, and re-renders them to produce label-valid training samples whose difficulty is calibrated to the actor's current ability through a pass-rate-based reward. 
This loop continuously realigns task difficulty with actor capability without any additional human annotation. 
Across nine multimodal benchmarks spanning mathematical reasoning and visually grounded understanding, \our{}-8B improves over its base model by up to \textbf{+5.0\%} on out-of-domain tasks, surpasses the strongest self-evolution and text-editing co-evolution baselines by \textbf{+4.3\%} and \textbf{+8.4\%} respectively, and stays comparable to chart-specialized RLVR methods using \textbf{16–160× fewer labeled samples}. 
By shifting from human-labeled supervision to image-editing co-evolution, \our{} offers a scalable path beyond static-corpus RLVR for visual reasoning.

\end{abstract}


%% file: tex/1-intro.tex
\section{Introduction}
\label{sec:introduction}

\begin{wrapfigure}{r}{0.62\textwidth}
    \centering
    \vspace{-1.5ex}
    \includegraphics[width=0.99\linewidth]{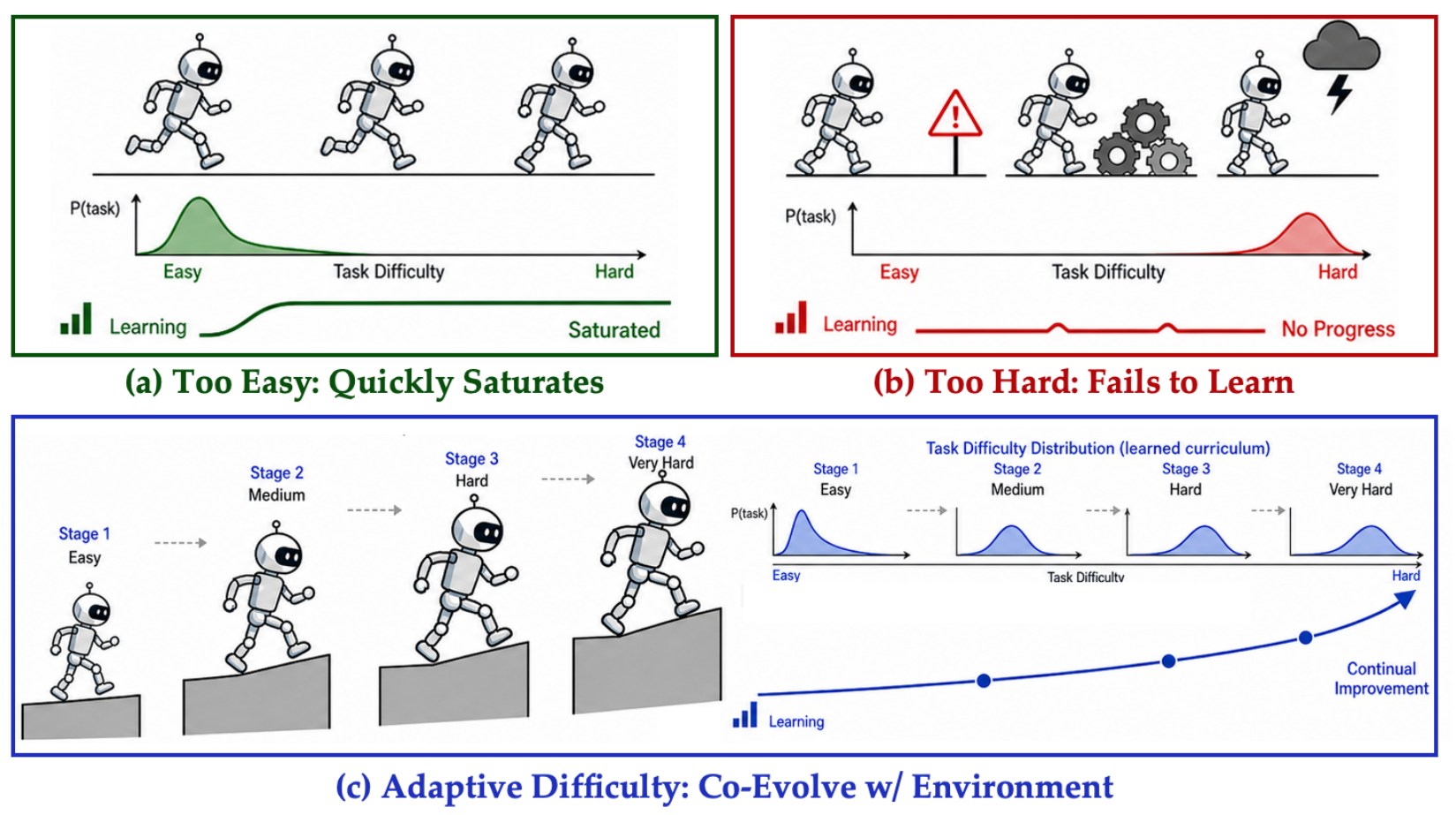}
    \caption{\textbf{Co-evolution with adaptive visual environments and actors leads to better training.}}
    \label{fig:ticker}
    \vspace{-1ex}
\end{wrapfigure}
Large Vision-Language Models (VLMs) have recently improved through reinforcement learning with verifiable rewards (RLVR)~\citep{Guo_2025}, where models learn from automatically checkable outcomes rather than dense human rationales~\citep{shao2024deepseekmathpushinglimitsmathematical}.
This recipe has made self-evolution an attractive paradigm: rather than relying on costly human supervision, a model can generate tasks, collect rollouts, and improve from its own experience~\citep{li2026mmzeroselfevolvingmultimodelvision,wang2025visionzero}. 
Yet most self-evolution pipelines share a structural limitation: the actor keeps changing, but the visual environment stays fixed. 
As the actor improves, fixed tasks drift out of its learning frontier, where many become trivial while others remain unsolvable. 
In both regimes, the rollout group provides little useful contrast (Figure~\ref{fig:ticker}): the empirical pass rate $\hat{p}$ moves toward $1$ or $0$, the Group Relative Policy Optimization (GRPO) reward variance, proportional to $\hat{p}(1-\hat{p})$, collapses, and the learning signal weakens precisely when the actor is ready to learn more~\citep{guo2025genenvdifficultyalignedcoevolutionllm}. 
Crucially, $\hat{p}$ already tells us exactly how difficult each task is for the current actor, yet existing methods use it only for normalization or filtering, discarding a signal that could otherwise guide the environment itself. 
This raises the central question of our work: \emph{can the visual environment adapt to the actor using the same rollout feedback that trains the actor?} 



Answering this question (\textit{i.e.}, adapting the visual environment), however, requires more than simply generating new images. Useful training samples must satisfy two constraints simultaneously.
First, they should be \emph{verifiable}: edits should not introduce ambiguous questions or incorrect labels, otherwise the verifiable reward signal that underpins RLVR breaks down.
Second, they should be \emph{capability-aligned}: edits should move tasks toward the actor's current learning frontier rather than making them arbitrarily harder or easier. 
Human annotation~\citep{masry2022chartqabenchmarkquestionanswering,lu2024mathvistaevaluatingmathematicalreasoning} satisfies both but does not scale, while offline augmentation with stronger teachers~\citep{chen2023sharegpt4vimprovinglargemultimodal,liu2023visualinstructiontuning} reduces annotation cost but produces a static distribution that cannot track the actor's evolving weaknesses. 
What is missing is an automatic visual environment that adapts to the actor while preserving the validity of the label.

We propose \our{}, a co-evolution framework that addresses both requirements through joint training of two VLMs. 
The \emph{actor} learns to answer visual reasoning questions, while the \emph{EnvRewriter} learns to perform verifiable structured edits to match the actor's current ability. 
For each rewritten task, we sample a group of actor rollouts and reuse the resulting pass rate $\hat{p}$ as a shared learning signal: 
binary rollout rewards update the actor through GRPO, while $\hat{p}$  updates the EnvRewriter through an $\alpha$-curriculum reward that peaks near $\hat{p}=0.5$, steering edits into the actor's \emph{learnable zone}. 
We then evaluate \our{} across three representative visual reasoning domains, including compositional reasoning, chart understanding, and general visual question answering, and against three categories of baselines: text-only rewriting, static-environment self-evolution, and offline augmentation with stronger teachers, including GPT-5~\citep{openai_gpt5_2025} and Nano Banana~\citep{fortin2025gemini25flashimage}. Across eight out-of-distribution benchmarks, \our{} exceeds the strongest self-evolution baseline by +4.3\% on average. Notably, even when stronger image generators are scaled up, static baselines plateau, indicating that the gains stem from the iterative actor–environment loop rather than from data generation quality alone.
Our contributions are as follows:
\begin{itemize}
    \item \textbf{An actor–environment co-evolution paradigm for VLM reasoning.} We introduce a framework in which the visual environment evolves alongside the actor, using the rollout pass rate $\hat{p}$ as a shared learning signal. This continually realigns task difficulty with actor capability and overcomes the saturation that limits fixed-environment self-evolution.
    \item \textbf{A verifiable image-editing training framework.} We design a pipeline that couples structured visual editing with RLVR, converting edited images into valid training samples through deterministic rendering and programmatic relabeling, with no human annotation.
    \item \textbf{Strong empirical performance across diverse benchmarks.} Across nine multimodal benchmarks, VICO surpasses RLVR baselines by +15.5\% on average, exceeds the strongest self-evolution baseline by +4.3\% on out-of-domain tasks, and outperforms the proprietary Gemini-2.5-Pro by +2.5\%.
    \item \textbf{Annotation-efficient training.} VICO is competitive with chart-specialized RLVR methods using 16×–160× fewer labeled samples, demonstrating that adaptive environments can substitute for large-scale human or teacher-model annotation.
\end{itemize}

%% file: tex/2-related.tex
\vspace{-5pt}
\section{Related Work}
\label{sec:related_work}

\paragraph{Reinforcement Learning for Vision-Language Reasoning.}
RLVR has become a standard post-training recipe for language and vision-language models, replacing dense human rationales with rule- or program-checkable outcomes~\citep{lambert2025tulu3pushingfrontiers,shao2024deepseekmathpushinglimitsmathematical}. 
Recent multimodal extensions of RLVR target multi-step visual reasoning through cross-modal symbolic formalization~\citep{yang2025r1onevisionadvancinggeneralizedmultimodal}, scaled rule-based reward training~\citep{meng2025mmeurekaexploringfrontiersmultimodal}, and combined SFT-RL pipelines~\citep{deng2025openvlthinkercomplexvisionlanguagereasoning}. 
Subsequent work refines RLVR through efficient visual processing~\citep{yang2025visionthinksmartefficientvision}, weakness-driven problem generation~\citep{liang2025swsselfawareweaknessdrivenproblem}, curiosity-driven exploration~\citep{dai2025cdecuriositydrivenexplorationefficient}, and multi-turn world-model reasoning~\citep{wang2025vagenreinforcingworldmodel}.
While these methods differ in reward design, exploration strategy, and training schedule, they share a common assumption: a fixed pool of visual tasks. 
\emph{\our{} instead treats the visual environment as a learnable component that co-adapts with the actor}, aiming to unlock new capabilities rather than refine existing ones over a static distribution~\citep{yue2025doesreinforcementlearningreally}.

\paragraph{Self-Evolving VLMs.} Self-evolving VLMs reduce reliance on human-curated supervision by letting models generate, solve, or refine their own training data. MM-Zero bootstraps multimodal reasoning from zero labeled data by proposing and solving visual problems~\citep{li2026mmzeroselfevolvingmultimodelvision}, VisPlay self-generates question-answer pairs from raw images for RL training~\citep{he2025visplayselfevolvingvisionlanguagemodels}, and Vision-Zero frames self-improvement as visual self-play with verifiable outcomes~\citep{wang2025visionzero}. Related efforts apply self-evolution to visual games~\citep{xie2025playgeneralizelearningreason,tong2025gamerlsynthesizingmultimodalverifiable}, perception~\citep{zhang2025viperempoweringselfevolutionvisual}, image-quality assessment~\citep{wen2026selfevolvingvisionlanguagemodelsimage}, self-refinement~\citep{deng2025selfrefinementvisionlanguagemodelstriangular}, and OCR hallucination mitigation~\citep{he2025seeingbelievingmitigatingocr}. Across these methods, the actor evolves but the visual environment remains fixed, leaving the training distribution increasingly misaligned with the actor's current ability.

\paragraph{Co-Evolving VLMs and Environments.} A more recent line of co-evolution work allows the environment to change with the model: GenEnv trains an environment policy for difficulty-aligned planning tasks~\citep{guo2025genenvdifficultyalignedcoevolutionllm}, V-Zero perturbs the textual side of visual tasks while keeping the image pool fixed~\citep{wang2026vzeroselfimprovingmultimodalreasoning}, C2-Evo co-evolves multimodal data and model training through model-driven generation~\citep{chen2025c2evocoevolvingmultimodaldata}, and Active-Zero retrieves from a large external image pool in an offline augmentation setting~\citep{he2026activezeroselfevolvingvisionlanguage}. 
\our{} differs from this line in two ways: it edits the visual substructure of each image rather than perturbing text or retrieving from a fixed pool, directly adapting the modality where reasoning is grounded; and it uses the actor's own pass rate as the supervision for environment updates, providing a difficulty-calibration loop tightly coupled to the actor's learning trajectory.

%% file: tex/3-method.tex
\input{figures/fig-main-figure}

\section{Methodology}
\label{sec:methodology}

\our{} views visual reasoning training as an \emph{actor-environment curriculum process} rather than a single model self-optimization problem. We maintain two policies: 
an \emph{actor} $\pi_a$ (the vision-language model being trained) and an \emph{EnvRewriter} $\pi_R$ (the environment that produces training samples). 
Unlike standard RLVR with a fixed training source, \our{} enables both to co-evolve: $\pi_a$ learns to solve visual reasoning tasks whose difficulty is shaped by the current EnvRewriter, while $\pi_R$ is rewarded for producing edited images whose difficulty is aligned with the actor's current capability — targeting the ``learnable zone'' where the learning signal is most effective.

\subsection{Data-Evolving Paradigm: From Static RLVR to Adaptive Co-evolution} 
\label{sec:paradigm}
Standard RLVR minimizes a loss $\mathcal{L}(\theta)$ over a static dataset $\mathcal{D}_{\text{static}} = \{(x, q, a)\}$, where $\theta$ denotes the parameters of the actor. In contrast, \our{} implements a  \emph{Data-Evolving Paradigm}. At each round $t$, the training dataset $\mathcal{D}_t$ is generated dynamically by the EnvRewriter $\pi_R$, based on the actor’s current pass rate. This establishes a feedback loop 
(Figure~\ref{fig:main-figure}) in which the EnvRewriter’s goal is not to beat the actor, but to identify its ``breaking points'' and convert them into a training signal.

\subsection{Seed Visual Data and Verifiable Sample Generation}
\label{sec:method3.2-data-creation}
We start \our{} from a small seed visual dataset and let the EnvRewriter
derive additional training examples by editing the structured or protected
representation of each seed sample in subsequent rounds. The key requirement is that every edited sample remains \emph{label-valid} while its difficulty changes. We therefore avoid unconstrained pixel-level generation. Instead, the answer can either be recomputed programmatically from the edited structure or preserved by construction through constrained editing. Table~\ref{tab:verifiable_editing_protocol} summarizes the editing interface for each domain.

\begin{table}[t]
\centering
\footnotesize
\setlength{\tabcolsep}{4pt}
\renewcommand{\arraystretch}{1.12}
\caption{%
\textbf{\our{} keeps edited samples verifiable across domains.}
Structured domains use deterministic recomputation, while natural images use constrained answer preservation; neither needs judging or relabeling.
}
\label{tab:verifiable_editing_protocol}
\begin{tabular}{@{}p{0.12\linewidth}p{0.43\linewidth}p{0.39\linewidth}@{}}
\toprule
\textbf{Domain} & \textbf{Editing Interface} & \textbf{Label-Validity Rule} \\
\midrule
CLEVR~\citep{johnson2016clevrdiagnosticdatasetcompositional}
& Edit the scene graph: attributes, objects, shapes, or relations.
& Re-execute the question program on the edited graph; reject rendering or ambiguity failures. \\

ChartQA~\citep{masry2022chartqabenchmarkquestionanswering}
& Edit the CSV table under a two-tier rule: Tier~1 has a unique answer cell; Tier~2 has no unique cell mapping.
& Tier~1 computes $a'=\mathrm{table}'[r,c]$; Tier~2 preserves $a'=a$. Reject schema, rendering, or invariance failures. \\

ZwZ~\citep{wei2026zoomingzoomingregiontoimagedistillation}
& Edit only the complement of protected target masks, e.g., background or non-target distractors.
& Preserve $(q',a')=(q,a)$; reject edits whose mask overlaps protected targets or fails rendering. \\
\bottomrule
\end{tabular}
\vspace{-2ex}
\end{table}

We experiment with three types of data: $2{,}000$ CLEVR with scene graphs and question programs, $2{,}000$ ChartQA with their underlying CSV tables, and $1{,}000$ ZwZ with answer-relevant region annotations. The actor–environment co-evolution loop is shared across all three domains; only the verifiable editing interface depends on the structure available in each dataset. More details are provided in Appendix~\ref{appendix:programmatic-data-generation}. We further analyze seed-sampling stability in Appendix~\ref{app:multi-seed-stability}; three ChartQA runs with different seed samples fluctuate by less than $2\%$.

\subsection{Rewards and Co-Evolutionary Optimization}
\label{sec:reward_and_training}


\paragraph{Actor Reward and Update.}
The EnvRewriter generates a training instance $(x', q', a')$ with a verified answer $a'$, where $a'$ is programmatically computed from the modified structured representation (Appendix~\ref{appendix:programmatic-data-generation}). This verifiable answer enables rule-based rewards, avoiding neural reward models that are prone to reward hacking. The actor's response receives a reward $+1$ if it both produces the correct answer (exact match against $a'$) and follows the required \texttt{<think>}/\texttt{<answer>} format; otherwise, it receives $-1$ (Appendix~\ref{appendix:reward_details})~\citep{zhou2025r1zerosahamomentvisual}. The actor $\pi_a$ is updated to maximize this reward via GRPO~\citep{shao2024deepseekmathpushinglimitsmathematical} on edited triples sampled from $\mathcal{D}_a$.



\paragraph{EnvRewriter Reward and Update.}
The core innovation for $\pi_R$ is a difficulty-calibration reward that targets a success-rate band around $\alpha = 0.5$. For each edited image $x'$, the actor generates $K$ rollouts, and we compute the empirical pass rate $\hat{p}_{\text{final}}$ as the fraction of correct rollouts. The EnvRewriter reward is a bell-shaped function of $\hat{p}_{\text{final}}$ peaked at $\alpha$ (Appendix~\ref{appendix:reward_details}), which discourages edits that are already mastered ($\hat{p}_{\text{final}} \to 1$) or hopeless ($\hat{p}_{\text{final}} \to 0$). We additionally apply a difficulty filter that excludes edits with $|\hat{p}_{\text{final}} - \alpha| > k_{\min}$ from EnvRewriter updates, where $k_{\min} \in (0, \tfrac{1}{2}]$ is the maximum allowed deviation from $\alpha$, to prevent overfitting to transient spikes; a theoretical justification is provided in Section~\ref{sec:theory}. The EnvRewriter $\pi_R$ is updated via Reward-Weighted Regression (RWR) on a weighted SFT set of edit trajectories, with stability ensured by a KL penalty to the initial EnvRewriter $\pi_R^{(0)}$ and a maximum KL threshold $\delta_{\max}$. Appendix~\ref{appendix:algorithm} Algorithm~\ref{alg:coevolve} summarizes the full optimization loop. 

%% file: figures/fig-main-figure.tex
\begin{figure}
  \centering
  \includegraphics[width=\linewidth]{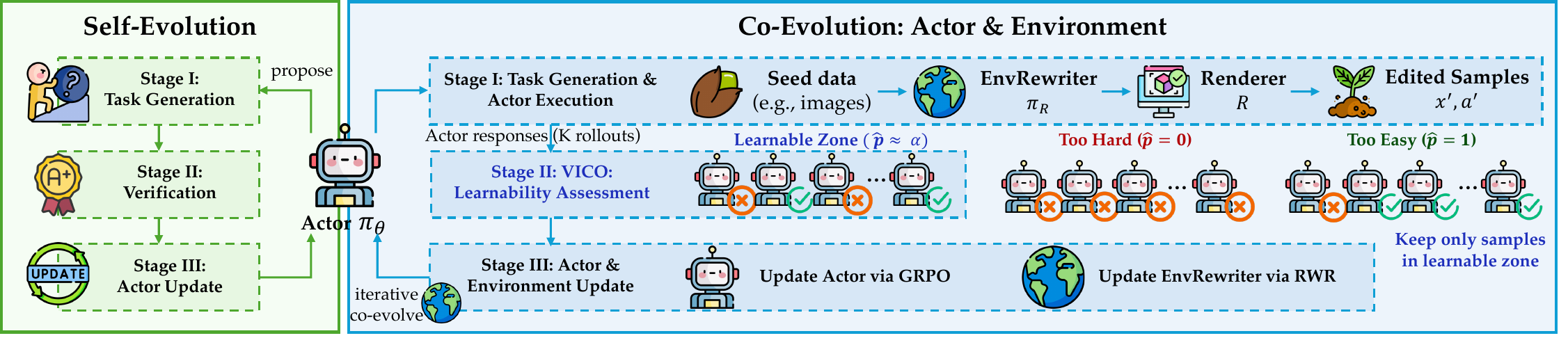}
  \vspace{-2.5ex}
  \caption{\textbf{\our{} turns self-evolution into an adaptive actor--environment loop.} The EnvRewriter creates visual tasks, the Actor solves them, and shared rollouts update both policies: task rewards train the Actor, while difficulty-aligned rewards train the EnvRewriter.}
  \label{fig:main-figure}
\end{figure}

%% file: tex/3-2-method.tex
\section{Theoretical Analysis of Difficulty-Calibrated \our{}}
\label{sec:theory}

We provide a simple analysis to clarify two properties of \our{}'s
co-evolution mechanism. First, edits with pass rates close to
$\alpha=0.5$ provide the strongest learning signal for the actor.
Second, as the actor matures, the optimal target can shift away from
$0.5$, which explains the saturation observed in our experiments.
Full derivations are deferred to Appendix~\ref{appendix:theory-proofs}.

\paragraph{Intermediate difficulty maximizes the actor's learning signal.}
Consider a stylized bandit setting where the actor receives a binary reward
$r\in\{0,1\}$ with success probability $p$ on an edited image $x'$. Under a
REINFORCE-style update~\citep{Williams1992} with a variance-minimizing
baseline~\citep{kaledin2022variancereductionpolicygradientmethods}, and under
a standard bounded-score assumption for policy-gradient updates, the squared gradient norm satisfies
\begin{equation}
  \mathbb{E}\big[\|g(x', r)\|^2\big] \;\propto\; p(1-p),
  \label{eq:variance-bound}
\end{equation}
up to constants independent of $p$. Since $p(1-p)$ is concave and maximized at
$p=1/2$, the gradient signal is strongest for edits of intermediate difficulty.
Equivalently, maximizing $p(1-p)$ is the same as minimizing
$(p-\tfrac{1}{2})^2$. This supports adopting the
$\alpha$-Curriculum reward~\citep{guo2025genenvdifficultyalignedcoevolutionllm}
$R_{\text{rewriter}}(\hat{p})=\exp(-\beta(\hat{p}-\alpha)^2)$ with
$\alpha=0.5$ as the difficulty-calibration objective for the EnvRewriter.

\paragraph{A drift-corrected target predicts saturation.}
The target $\alpha=0.5$ is optimal only when edits are costless. As the actor
improves, its pass rate $\hat{p}_{\text{orig},t}$ on the unedited image rises.
Forcing the edited sample back to $\hat{p}=0.5$ may then require increasingly
aggressive edits. These edits can move the training distribution away from the
natural image manifold~\citep{alemohammad2023selfconsuminggenerativemodelsmad}.
We model this trade-off by adding a quadratic drift penalty:
\begin{equation}
  \Delta\text{Acc}_t
  =
  C_t p(1-p)
  -
  \lambda \kappa_t
  (p-\hat{p}_{\text{orig},t})^2,
  \label{eq:drift-objective}
\end{equation}
where $C_t>0$ denotes the actor's remaining learning headroom, and
$\lambda\kappa_t\geq 0$ controls the drift cost. Maximizing this concave
objective gives
\begin{equation}
  p^*_t
  =
  \tfrac{1}{2}
  +
  \rho_t
  \left(\hat{p}_{\text{orig},t}-\tfrac{1}{2}\right),
  \quad
  \rho_t
  =
  \frac{\lambda \kappa_t}{C_t+\lambda \kappa_t}
  \in [0,1],
  \label{eq:optimal_target}
\end{equation}
so the optimal target is a convex combination of $0.5$ and
$\hat{p}_{\text{orig},t}$. When the actor has large headroom
($\rho_t\to 0$), the optimal target remains near $0.5$. As headroom shrinks or
drift cost increases ($\rho_t\to 1$), the optimum moves toward the unedited pass
rate. Continued editing then enters a drift-dominated regime, where stronger
perturbations bring smaller gains and may eventually hurt performance. This
analysis predicts diminishing per-round gains, fewer sweet-zone samples, and
eventual saturation, which we verify in Appendix~\ref{appendix:RQ9}. 

%% file: tex/4-experiments.tex
\section{Experiments}
\label{sec:exp}
\subsection{Experimental Setups}

\input{tables/tab-main-results-table}

\paragraph{Training Setup.}
For the Actor $\pi_a$, we train via GRPO~\citep{shao2024deepseekmathpushinglimitsmathematical} with $n=8$ on-policy rollouts and a learning rate $1\!\times\!10^{-6}$. The EnvRewriter $\pi_R$ is updated via reward-weighted LoRA SFT with $\beta=4.0$ and $\alpha=0.5$ (\citep{guo2025genenvdifficultyalignedcoevolutionllm}). We use AdamW for both policies, run \our{} for $T=5$ rounds, and use up to $8\!\times\!\text{A100-80GB}$ GPUs with verl~\citep{Sheng_2025}. We train Qwen3-VL-8B~\citep{bai2025qwen3vltechnicalreport} on three visual datasets, and Qwen2.5-VL-7B~\citep{bai2025qwen25vltechnicalreport} and InternVL3-8B~\citep{zhu2025internvl3exploringadvancedtraining} on chart understanding to assess backbone generalizability. We validate \our{} on three visual reasoning domains: CLEVR~\citep{johnson2016clevrdiagnosticdatasetcompositional} for compositional reasoning, ChartQA~\citep{masry2022chartqabenchmarkquestionanswering} for chart reasoning, and ZwZ~\citep{wei2026zoomingzoomingregiontoimagedistillation} for natural-image VQA. Full hyperparameters, two-stage training, dataset examples, and multi-seed analyzes are in Appendix~\ref{appendix:training-details}.

\paragraph{Tasks and Datasets.} 
We evaluate on nine benchmarks under identical zero-shot settings (temperature $0.7$, top-$p$ $1.0$, max response length $3{,}824$ tokens). For mathematical reasoning, we use MathVista~\citep{lu2024mathvistaevaluatingmathematicalreasoning}, MathVision~\citep{wang2024measuringmultimodalmathematicalreasoning}, MathVerse~\citep{zhang2024mathversedoesmultimodalllm}, WeMath~\citep{qiao2024wemathdoeslargemultimodal}, and LogicVista~\citep{xiao2024logicvistamultimodalllmlogical}. For chart reasoning, we use ChartQA~\citep{masry2022chartqabenchmarkquestionanswering}. For general vision-centric understanding, we use MMMU~\citep{yue2024mmmumassivemultidisciplinemultimodal}, MMVP~\citep{zhang2024mmvpmultimodalmocapdataset}, and BLINK~\citep{fu2024blinkmultimodallargelanguage}. The \textit{Average} column in Table~\ref{tab:main_results_table} is the equally weighted average of the mathematical-reasoning average and the vision-centric average. 

\paragraph{Baselines.} 
We compare \our{} against three categories of strong baselines reflecting different post-training paradigms: 
(1) \textbf{RLVR-trained reasoning models} trained with RLVR on human-labeled data---R1-OneVision-7B~\citep{yang2025r1onevisionadvancinggeneralizedmultimodal}, MM-Eureka-Qwen-7B~\citep{meng2025mmeurekaexploringfrontiersmultimodal}, OpenVLThinker-7B~\citep{deng2025openvlthinkercomplexvisionlanguagereasoning}; 
(2) \textbf{Self-evolution models} that evolve the model in a fixed environment---MM-Zero-8B~\citep{li2026mmzeroselfevolvingmultimodelvision}, three VisionZero-7B variants (CLEVR, Chart, Real-world), and VisionZero-InternVL3-8B~\citep{wang2025visionzero}; 
(3) \textbf{Co-evolution baseline} V-Zero~\citep{wang2026vzeroselfimprovingmultimodalreasoning}, which perturbs question-answer pairs without editing the visual environment. 
These three categories isolate the effect of \emph{visual-environment co-evolution} from data-rich RL post-training and model-side self-evolution. Implementation details are in Appendix~\ref{appendix:training-details}.

\begin{table}[t]
\centering
\footnotesize
\caption{\textbf{\our{} improves VLM backbones with far less training data.}
The same pipeline improves Qwen2.5-VL-7B and InternVL3-8B, while using $16\times$--$160\times$ less data than chart-specialized RL models. Overall average uses the seven displayed benchmarks; bold marks the best result and underline marks the second-best. \textit{{*}Average is computed over reported columns only.}}
\label{tab:backbone_generality}
\resizebox{\textwidth}{!}{%
\begin{tabular}{l|c|ccccc|cc|c}
\toprule
\textbf{Baselines ($\downarrow$)} & \textbf{Seed Data}
 & \multicolumn{5}{c|}{\textbf{Mathematical Reasoning}}
 & \multicolumn{2}{c}{\textbf{Vision-Centric}}
 & \textbf{Overall}\\
\cmidrule(lr){3-7} \cmidrule(lr){8-9}
 \textbf{Datasets ($\rightarrow$)}& & MathVista & MathVerse
 & ChartQA & WeMath & LogicVista & MMMU & BLINK
 & avg. \\
\midrule
\multicolumn{10}{l}{\textit{Chart-Specialized Methods}} \\
\midrule
Chart-R1-7B~\citep{chen2026chartr1chainofthoughtsupervisionreinforcement}
& $\sim$33k  & -    & -    & \textbf{91.0} & -    & -    & -    & -    & - \\
Bespoke-MiniChart-7B~\citep{movva2025enhancingscientificvisualquestion}
& $\sim$91k  & -    & -    & \underline{89.5} & -    & -    & -    & -    & - \\
ECD (Qwen2.5-VL-7B)~\citep{yang2025effectivetrainingdatasynthesis}
& $\sim$320k & -    & -    & 85.3 & -    & -    & -    & -    & - \\
\midrule
\multicolumn{10}{l}{\textit{Qwen2.5-VL-7B Family}} \\
\midrule
Qwen2.5-VL-7B (base)
& --        & 68.2 & 39.6 & 76.4 & 60.7 & \textbf{47.2} & \textbf{54.7} & 55.2 & 57.4 \\
OpenVLThinker-7B~\citep{deng2025openvlthinkercomplexvisionlanguagereasoning}
& $\sim$50k & 70.2 & 47.9 & 59.9 & 36.5 & 44.3 & 36.1 & \textbf{61.2} & 50.9 \\
MM-Eureka-Qwen-7B~\citep{meng2025mmeurekaexploringfrontiersmultimodal}
& 15k       & \underline{73.0} & 50.3 & 58.2 & 36.2 & 42.9 & 50.1 & \underline{58.3} & 52.7 \\
R1-OneVision-7B~\citep{yang2025r1onevisionadvancinggeneralizedmultimodal}
& 10k       & 64.1 & 46.4 & 73.4 & \underline{61.8} & 44.5 & 46.4 & 48.7 & 55.0 \\
VisionZero-Qwen7B (CLEVR)~\citep{wang2025visionzero}
& 2000      & 72.2 & \underline{53.2} & 86.3 & 49.8 & \underline{45.8} & 53.0 & 56.0 & 59.5 \\
VisionZero-Qwen7B (Chart)
& 1000      & 72.4 & 52.5 & \underline{87.2} & 51.2 & 43.5 & 53.6 & 56.8 & \underline{59.6} \\
VisionZero-Qwen7B (Real-world)
& 1000      & \textbf{73.1} & 52.1 & 86.3 & 50.8 & 42.6 & 52.5 & 57.2 & 59.2 \\
C\textsuperscript{2}-Evo~\citep{chen2025c2evocoevolvingmultimodaldata}
& --        & 50.2 & 19.1 & -    & -    & -    & -    & -    & 34.7\textsuperscript{*} \\
\rowcolor{blue!8}
\textbf{\our{}-Qwen2.5-VL-7B (Chart)}
& 2000      & 72.8 & \textbf{54.9} & \textbf{88.0} & \textbf{67.6} & 44.0 & \underline{53.8} & 53.7 & \textbf{62.1} \\
\midrule
\multicolumn{10}{l}{\textit{InternVL3-8B Family}} \\
\midrule
InternVL3-8B (base)
& --        & 60.4 & 32.2 & \underline{69.9} & \underline{61.3} & 34.6 & 55.3 & \underline{61.2} & \underline{53.6} \\
MM-Eureka-InternVL-8B
& 15k       & \underline{62.4} & 32.1 & 54.9 & 58.2 & 38.9 & 51.6 & 61.0 & 51.3 \\
VisionZero-InternVL3-8B
& 2000      & 62.2 & \underline{32.9} & 53.0 & 28.7 & \textbf{41.8} & \textbf{56.9} & \textbf{63.3} & 48.4 \\
\rowcolor{blue!8}
\textbf{\our{}-InternVL3-8B (Chart)}
& 2000      & \textbf{68.9} & \textbf{57.6} & \textbf{80.6} & \textbf{63.4} & \underline{40.2} & \underline{55.7} & 58.5 & \textbf{60.7} \\
\bottomrule
\end{tabular}%
}
\vspace{-8pt}
\end{table}

\subsection{Main Results}
\textbf{\our{}-actor consistently improves over the base model and outperforms three categories of baselines: RL-trained reasoning models, self- and co-evolution methods.} Table~\ref{tab:main_results_table} presents the main comparison. For the Chart Understanding variant, we report the OOD average over the eight benchmarks excluding ChartQA, which is its in-domain benchmark. For the CLEVR and ZwZ variants, all nine evaluation benchmarks are treated as out-of-domain. Notably, \our{} attains these improvements using only 2000 seed inputs, which is $16\times$–$160\times$ fewer than the chart-specialized RLVR baselines that depend on large amounts of human or model-generated CoT annotations (Table~\ref{tab:backbone_generality}). The strongest variant, \our{}-8B (Chart), achieves the best OOD average accuracy of $64.9\%$ (excluding in-domain ChartQA), improving over the base model by $+5.0\%$. Against RL-trained reasoning baselines, \our{} surpasses them by $+15.5\%$ on average. Against the strongest self-evolution baseline (MM-Zero), the three \our{} variants surpass it by $+4.3\%$ on the OOD benchmarks. Against the text-editing co-evolution baseline (V-Zero), \our{} outperforms it by $+8.4\%$ on the overlapping benchmarks. Beyond open-source baselines, \our{}-8B also outperforms Gemini-2-Flash and Gemini-2.5-Pro by $+5.2\%$ and $+2.5\%$ on the OOD benchmarks, showing that an 8B model trained through co-evolution can compete with much larger proprietary VLMs.

\textbf{\our{} has strong generalization across OOD domains and models.}
As shown in Table~\ref{tab:main_results_table}, \our{}-8B variants gain $+5.2\%$ on average across OOD benchmarks.
The transfer is strongest in mathematical reasoning: without math-specific training, \our{}-8B (CLEVR), \our{}-8B (Chart), and \our{}-8B (ZwZ) improve over Qwen3-VL-8B by $+5.5\%$, $+5.4\%$, and $+5.3\%$ on MathVista, MathVision, MathVerse, WeMath, and LogicVista.
This suggests that \our{} learns transferable reasoning skills rather than domain-specific edit shortcuts.
By comparison, existing RLVR baselines transfer less reliably: R1-OneVision-7B and MM-Eureka-Qwen-7B improve on one benchmark, but Qwen2.5-based RLVR baselines drop by $-7.3\%$ on average relative to their base models.
\our{} also transfers across backbones. Table~\ref{tab:backbone_generality} shows that \our{}-Qwen2.5-VL-7B (Chart) and \our{}-InternVL3-8B (Chart) improve over their base models by $+4.7\%$ and $+7.1\%$ on average.

\begin{table}[t]
\centering
\begin{minipage}[t]{0.45\linewidth}
\vspace{0pt}
\centering
\footnotesize
\setlength{\tabcolsep}{6pt}
\renewcommand{\arraystretch}{1.2}
\captionof{table}{%
\textbf{Visual edits outperform text-only perturbations.} With the same CLEVR seed source, image editing in \our{} improves the overall average by $+6.1\%$, compared with $+2.4\%$ for text edits. The largest gap appears on ChartQA ($+9.4\%$ vs. $+0.9\%$).}
\label{tab:image_vs_text_editing}
\resizebox{\linewidth}{!}{%
\begin{tabular}{@{}lccc@{}}
\toprule
\textbf{Benchmark} & \textbf{Base} & \makecell{\textbf{\our{}-Qwen3VL-8B}\\\textbf{(CLEVR, text-edit)}} & \makecell{\textbf{\our{}-Qwen3VL-8B}\\\textbf{(CLEVR, image-edit)}} \\
\midrule
\multicolumn{4}{@{}l}{\textit{Mathematical reasoning}} \\
\midrule
\quad MathVista   & 67.7 & 70.3\diffup{2.6}   & \textbf{72.8}\diffup{5.1} \\
\quad MathVision  & 31.5 & 39.0\diffup{7.5}   & \textbf{41.0}\diffup{9.5} \\
\quad MathVerse   & 50.6 & 53.9\diffup{3.3}   & \textbf{56.0}\diffup{5.4} \\
\quad WeMath      & 72.0 & 73.1\diffup{1.1}   & \textbf{76.8}\diffup{4.8} \\
\quad LogicVista  & 55.1 & 54.1\diffdown{1.0} & \textbf{57.8}\diffup{2.7} \\
\midrule
\multicolumn{4}{@{}l}{\textit{Chart reasoning}} \\
\midrule
\quad ChartQA     & 76.9 & 77.8\diffup{0.9}   & \textbf{86.3}\diffup{9.4} \\
\midrule
\rowcolor{gray!10}
\textbf{Overall avg.} & 59.0 & 61.4\diffup{2.4} & \textbf{65.1}\diffup{6.1} \\
\bottomrule
\end{tabular}%
}
\end{minipage}%
\hfill
\begin{minipage}[t]{0.48\linewidth}
\vspace{0pt}
\centering
\footnotesize
\setlength{\tabcolsep}{6pt}
\renewcommand{\arraystretch}{1.2}
\captionof{table}{%
\textbf{Iterative co-evolution beats static augmentation.} Strong offline generators such as GPT-5 and Nano Banana improve ChartQA accuracy, but they remain below actor-conditioned visual co-evolution, which reaches $89.1\%$.}
\label{tab:static_agumentation_offline-generation-baseline}
\resizebox{\linewidth}{!}{%
\begin{tabular}{@{}lccc@{}}
\toprule
\textbf{Setting} & \makecell{\textbf{Iter.}\\\textbf{loop}} & \makecell{\textbf{Visual}\\\textbf{edit}} & \textbf{ChartQA} \\
\midrule
Base (Qwen3-VL-8B)                                       & --      & --      & 76.9 \\
\midrule
\multicolumn{4}{l}{\textit{Static offline augmentation (no loop)}} \\
\midrule
\quad Offline-Random (self-gen)                          & \xmark  & \cmark  & 78.6\diffup{1.7}  \\
\quad Offline-GPT-5                                      & \xmark  & \cmark  & 80.1\diffup{3.2}  \\
\quad Offline Nano Banana                                 & \xmark  & \cmark  & 81.9\diffup{5.0}  \\
\midrule
\multicolumn{4}{l}{\textit{RLVR-only (no environment adaptation)}} \\
\midrule
\quad Supervised GRPO                                    & \xmark  & \xmark  & 79.4\diffup{2.5}  \\
\quad Direct GRPO                                        & \xmark  & \xmark  & 83.8\diffup{6.9}  \\
\quad \our{}-Qwen3VL-8B (Chart, text-edit)               & \cmark  & \xmark  & 77.2\diffup{0.3}  \\
\rowcolor{gray!10}
\quad \textbf{\our{}-Qwen3VL-8B (Chart, image-edit)}     & \cmark  & \cmark  & \textbf{89.1}\diffup{12.2} \\
\bottomrule
\end{tabular}%
}
\end{minipage}
\end{table}

\textbf{Editing images is more effective than rewriting text alone.}
We compare visual-environment editing with text-only perturbation in Table~\ref{tab:image_vs_text_editing}. Within the same \our{} environment, image editing yields a $+6.1\%$ gain in the evaluated reasoning benchmarks, while text-only perturbation improves by only $+2.4\%$. The gap is especially clear on ChartQA, where image editing gains $+9.4\%$ versus only $+0.9\%$ for text-only perturbation. This finding is further supported by independent evidence: in the separately developed V-Zero framework, text-only perturbation also improves mathematical reasoning by only $+1.8\%$ over its base model. Together, these results show that the key bottleneck is the editing modality itself: effective co-evolution requires modifying the visual environment, not just rewriting the text.

\input{figures/fig-combined-figure}
\textbf{\our{} trains stably and improves across training rounds.}
Figure~\ref{fig:combined_figure}(a) shows that the step-wise GRPO training score increases smoothly without collapse or divergence. Across rounds, Figure~\ref{fig:combined_figure}(b) shows that \our{} steadily improves on ChartQA. Accuracy rises from $76.9\%$ at the base model to $89.1\%$ at R4, giving a total gain of $+12.2\%$. Each round contributes a positive gain. This improvement also appears at the sample level. Appendix
Table~\ref{appendix-tab:capability_unlock} shows that R1 unlocks $13.9\%$ test samples, and the cumulative unlocked fraction grows to $16.1\%$ by R4. These results show that \our{} accumulates new correct behaviors across rounds rather than reflecting round-specific fluctuations.

\textbf{VICO's Gains concentrate on reasoning-heavy question types.} 
Figure~\ref{fig:combined_figure}(d) breaks down ChartQA accuracy by task type. The largest gains appear in multi-step reasoning, including ratio/percentage ($+29.8\%$), value lookup ($+20.1\%$), and difference ($+11.9\%$). In contrast, visually simpler tasks improve less, such as comparison ($+7.1\%$) and counting ($+8.2\%$). The yes/no category slightly drops ($-2.6\%$), likely because it is already saturated. These results suggest that \our{} mainly improves reasoning-intensive skills rather than simply boosting easy visual recognition patterns.

\textbf{\our{} turns errors into correct answers with limited regression.} 
To understand where the gain of round 4 comes from, we compare each sample's correctness between the base model and our round 4 actor on the ChartQA test set (Appendix Table~\ref{tab:flip}). \our{} converts 365 samples from wrong to right ($14.6\%$), while only 64 samples flip from right to wrong ($2.6\%$). This gives an improvement-to-regression ratio of $\mathbf{5.7\times}$. The net gain of 301 samples corresponds to $+12.0\%$, closely matching the aggregate accuracy gain of $+12.2\%$. Qualitatively, wrong-to-right cases often show explicit step-by-step decomposition, whereas the base model tends to give single-shot answers; examples are provided in Appendix~\ref{appendix:cot-analysis}. These results show that the gains mainly come from newly acquired correct behaviors, not from trading off many previously correct samples.

\subsection{Co-Evolution Dynamics, Capability Emergence, and Ablations}
We then examine whether the actor--environment loop behaves as intended and whether this loop is the source of the gains.

\input{tables/tab-rewriter-init}
\textbf{Both Rewriter SFT and the co-evolution loop are required.}
Table~\ref{tab:rewriter_init} separates the roles of supervised initialization
and iterative co-evolution. Actor SFT alone improves ChartQA by only $+2.5\%$,
so stronger actor initialization cannot explain the full gain. Running the
co-evolution loop with a vanilla EnvRewriter gives only $+1.7\%$, because the
Rewriter cannot effectively produce executable edit plans. Thus, the loop alone
is also insufficient without a valid environment policy. The full $+12.2\%$
gain appears only when the EnvRewriter is SFT-initialized and then optimized
through the iterative actor--environment loop. This shows that Rewriter SFT
provides valid environment edits, while co-evolution turns those edits into
actor-aligned training signals.

\textbf{The EnvRewriter learns bidirectional difficulty calibration.}
We compare two domains with opposite initial actor competence: ChartQA starts easy for the actor ($\hat{p}_{\text{orig}}\approx 0.85$), while ZwZ starts hard ($\hat{p}_{\text{orig}}\approx 0.21$). The EnvRewriter adapts in opposite directions. It makes ChartQA harder and lowers $\hat{p}_{\text{final}}$, while making ZwZ easier and raising $\hat{p}_{\text{final}}$. In both cases, $\hat{p}_{\text{final}}$ enters $[0.35,0.65]$ within the first few rounds
(Figure~\ref{fig:bidirectional_calibration}). This shows actor-relative calibration instead of random perturbation.


\textbf{Stronger edits reflect learned decisions, not longer instructions.}
We next ask whether the EnvRewriter becomes a stronger editor across rounds. We measure edit intensity as
$|\hat{p}_{\text{orig}}-\hat{p}_{\text{final}}|$, the pass-rate shift caused by
an edit for the current actor. Edit intensity increases from $0.22$ to $0.51$, showing that the EnvRewriter changes task difficulty more effectively.
This increase is not due to longer outputs: the average instruction length remains nearly unchanged ($973 \to 981$ tokens) (Figure~\ref{fig:rewriter-edit-length} in the Appendix). In contrast, the actor's response length grows by
$+49\%$ (Figure \ref{fig:combined_figure}(c)), consistent with longer reasoning traces on harder tasks. These trends provide behavioral evidence of co-adaptation: the EnvRewriter produces more targeted edits, while the actor adapts with more elaborate reasoning. Additional analyzes are provided in Appendix~\ref{appendix:appendix_editing_intensity_saturation} and
Appendix~\ref{appendix:cot-analysis}.

\textbf{The iterative visual co-evolution loop drives the gains.}
%
We ablate whether the improvement comes from the co-evolution loop,
the stronger offline data or RLVR-only training. First, we remove the loop
and compare it against static offline augmentation. As shown in
Table~\ref{tab:static_agumentation_offline-generation-baseline}, stronger
teachers improve the fixed corpus only modestly: Qwen3-VL-8B self-generation,
GPT-5, and Nano Banana reach $78.64\%$, $80.10\%$, and $81.91\%$ on ChartQA.
All remain below $82\%$. In contrast, \our{}-8B (Chart) reaches $89.12\%$,
exceeding the strongest static method by $+7.21\%$. Scaling the offline teacher
from self-generation to Nano Banana gives only $+3.27\%$, while adding the
iterative loop gives $+7.21\%$. Second, we keep the RLVR setup fixed but remove Rewriter-driven environment adaptation. Across all \our{} variants, co-evolution outperforms Direct GRPO by
$+3.7\%$ and Supervised GRPO by $+4.8\%$ on average. These results show that
\textit{the gains come from VICO's iterative, difficulty-aligned actor-environment
 co-evolution}.

\textbf{\our{} is robust against imperfect image editing.}
Automatic image editors may produce blank images, unchanged images, or
image-label mismatches. We therefore stress-test \our{} by corrupting $20\%$ of
the R1 training data. As shown in Table~\ref{tab:20-noise}, the overall
average performance drops by less than $1\%$. The noisy model still remains well
above the baseline model. This shows that \our{} does not require a flawless image
editor: failed edits do not dominate training, and the
remaining valid samples are sufficient for stable co-evolution. Additional details in
Appendix~\ref{appendix:20-noise}.

\textbf{Image-generator capability does not replace co-evolution.}
Our editing strategy must control difficulty while keeping labels verifiable.
Simple perturbations, such as noise, rotation, or blur, can lower the Actor's pass rate, but they mostly add perceptual noise and may shift samples away from the task distribution.
Strong image generators produce better images, yet unconstrained generation can still cause answer drift or ambiguous labels.
In contrast, \our{} edits verifiable visual structures and preserves supervision through programmatic relabeling or answer-preserving editing.
Thus, strong generators are useful backends, but not replacements for \our{}: the core benefit is the actor-conditioned loop that selects, verifies, and re-aligns edits using rollout feedback.
See Appendix~\ref{appendix:case_dynamic_editing}.

\input{figures/fig-saturation-evidence}



\input{figures/fig-bidirectional-calibration}

\input{tables/tab-20-noise}











%% file: tables/tab-main-results-table.tex
\begin{table*}[t]
\centering
\footnotesize
\renewcommand\arraystretch{0.93}
\caption{\textbf{\our{} achieves the strongest overall performance among comparable open-source VLMs.}
We report accuracy on nine multimodal reasoning benchmarks; \textbf{bold} marks the best result and \underline{underline} marks the second-best among non-commercial models. }
\resizebox{1.0\linewidth}{!}{%
\begin{tabular}{l @{\hskip2.5pt} | c c c c c c | c | c c c | c | c}
\toprule
\textbf{Baselines ($\downarrow$)} & \multicolumn{7}{c|}{\textbf{Mathematical Reasoning}} & \multicolumn{4}{c|}{\textbf{General Vision-Centric}} & \textbf{All} \\
\cmidrule(lr){2-8} \cmidrule(lr){9-12} \cmidrule(lr){13-13}
\textbf{Datasets ($\rightarrow$)}
  & MathVista & MathVision & ChartQA & MathVerse & WeMath & LogicVista & avg.
  & MMMU & MMVP & BLINK & avg.
  & avg. \\
\midrule
\multicolumn{13}{l}{\textit{Commercial VLMs (For reference)}} \\
\midrule
GPT-4o            & 61.4 & 30.4 & 80.7 & 50.2 & 40.0 & 45.9 & 51.4 & 65.1 & 86.3 & 60.0 & 70.47 & 61.0 \\
GPT-5             & 80.2 & 41.1 & 85.9 & 68.1 & 64.7 & 49.1 & 64.9 & 68.6 & 84.7 & 74.8 & 76.0 & 70.4 \\
Gemini-2-Flash    & 73.4 & 41.3 & 77.0 & 54.4 & 57.1 & 56.2 & 56.9 & 52.6 & 83.0 & 59.6 & 65.1 & 62.5 \\
Gemini-2.5-Pro    & 72.8 & 29.6 & 80.1 & 67.0 & 68.9 & 52.7 & 61.9 & 54.8 & 75.7 & 78.1 & 69.5 & 65.7 \\
Gemini-3.1-pro    & 83.3 & 34.7 & 84.1 & 67.8 & 65.9 & 57.4 & 65.5 & 60.0 & 89.7 & 83.5 & 77.7 & 71.6 \\
\midrule
\multicolumn{13}{l}{\textit{Open-Source VLMs}} \\
\midrule
Qwen2.5-VL-7B     & 68.2 & 25.6 & 76.4 & 39.6 & 60.7 & 47.2 & 53.0 & 54.7 & 76.8 & 55.2 & 62.2 & 57.6 \\
Qwen3-VL-8B       & 67.7 & 31.5 & 76.9 & 50.6 & 72.0 & 55.1 & 59.0 & 55.8 & 77.7 & 68.9 & 67.5 & 63.2 \\
InternVL3-8B      & 60.4 & 21.3 & 69.9 & 32.2 & 61.3 & 34.6 & 46.6 & 55.3 & 64.3 & 61.2 & 60.3 & 53.5 \\
InternVL3-14B     & 74.1 & 33.8 & \underline{87.3} & 44.2 & 68.9 & 51.6 & 60.0 & 56.4 & 76.30 & 60.2 & 64.3 & 62.2 \\
\midrule
\multicolumn{13}{l}{\textit{RL-Trained Reasoning Baselines}} \\
\midrule
OpenVLThinker-7B          & 70.2 & 25.3 & 59.9 & 47.9 & 36.5 & 44.3 & 47.4 & 36.1 & 71.7 & 61.2 & 50.0 & 48.7 \\
MM-Eureka-Qwen-7B         & 73.0 & 26.9 & 58.2 & 50.3 & 36.2 & 42.9 & 47.9 & 50.1 & 73.0 & 58.3 & 51.7 & 49.8 \\
MM-Eureka-InternVL-8B     & 62.4 & 22.1 & 54.9 & 32.1 & 58.2 & 38.9 & 44.8 & 51.6 & 74.3 & 61.0 & 50.0 & 47.4 \\
R1-OneVision-7B           & 64.1 & 24.1 & 73.4 & 46.4 & 61.8 & 44.5 & 52.4 & 46.4 & 61.3 & 48.7 & 52.3 & 52.3 \\
\midrule
\multicolumn{13}{l}{\textit{Self-evolution}} \\
\midrule
MM-Zero (Qwen3-VL-8B)              & 67.7 & 39.6 & 80.8 & 45.1 & 73.4 & 55.4 & 60.3 & 58.7 & 77.3 & 67.5 & 67.8 & 64.1 \\
VisionZero-Qwen2.5-VL-7B (CLEVR)     & 72.2 & 28.4 & 86.3 & 53.2 & 49.8 & 45.8 & 56.0 & 53.0 & 79.5 & 56.0 & 58.0 & 57.0 \\
VisionZero-Qwen2.5-VL-7B (Chart)     & 72.4 & 28.9 & 87.2 & 52.5 & 51.2 & 43.5 & 56.0 & 53.6 & 79.1 & 56.8 & 58.1 & 57.0 \\
VisionZero-Qwen2.5-VL-7B (RW)        & 73.1 & 28.5 & 86.3 & 52.1 & 50.8 & 42.6 & 55.6 & 52.5 & 79.2 & 57.2 & 57.8 & 56.7 \\
VisionZero-InternVL3-8B            & 62.2 & 24.2 & 53.0 & 32.9 & 28.7 & 41.8 & 40.5 & 56.9 & 69.0 & 63.3 & 47.2 & 43.9 \\
\midrule
\multicolumn{13}{l}{\textit{Co-evolution}} \\
\midrule
V-Zero (Qwen2.5-VL-7B)
  & 69.6 & 27.0 & -    & 42.6 & -    & 48.6 & 47.0
  & 58.6 & -    & 66.9 & 62.8 & 54.9 \\
\rowcolor{magenta!12} \our{}-Qwen3-VL-8B (CLEVR)
  & 72.8 & \textbf{41.0} & 86.3 & \textbf{56.0} & \textbf{76.8} & 57.8 & \underline{65.1}
  & 64.9 & \textbf{80.3} & 69.9 & \underline{71.7} & \underline{68.4} \\
\rowcolor{magenta!3}  \quad Direct GRPO
  & 71.6\diffdown{1.2} & 37.8\diffdown{3.2} & 78.1\diffdown{8.2}
  & 54.0\diffdown{2.0} & 71.2\diffdown{5.6} & 54.7\diffdown{3.1}
  & 61.2\diffdown{3.9}
  & 64.1\diffdown{0.8} & 78.3\diffdown{2.0} & 66.1\diffdown{3.8}
  & 69.5\diffdown{2.2}
  & 65.4\diffdown{3.0} \\
\rowcolor{magenta!3}  \quad Supervised GRPO
  & 69.4\diffdown{3.4} & 36.9\diffdown{4.1} & 78.6\diffdown{7.7}
  & 54.1\diffdown{1.9} & 72.5\diffdown{4.3} & 56.0\diffdown{1.8}
  & 61.2\diffdown{3.9}
  & 54.6\diffdown{10.3} & 76.3\diffdown{4.0} & 68.3\diffdown{1.6}
  & 66.4\diffdown{5.3}
  & 63.8\diffdown{4.6} \\
\midrule
\rowcolor{teal!15} \our{}-Qwen3-VL-8B (Chart)
  & \textbf{78.4} & 39.4 & \textbf{89.1} & \underline{55.0} & \underline{74.5} & 56.5 & \textbf{65.5}
  & \textbf{65.2} & \underline{79.7} & \underline{70.5} & \textbf{71.8} & \textbf{68.6} \\
\rowcolor{teal!4}  \quad Direct GRPO
  & 70.9\diffdown{7.5} & 37.9\diffdown{1.5} & 83.8\diffdown{5.3}
  & 52.7\diffdown{2.3} & 73.0\diffdown{1.5} & 57.0\diffup{0.5}
  & 62.5\diffdown{3.0}
  & \underline{65.0}\diffdown{0.2} & 77.7\diffdown{2.0} & 65.8\diffdown{4.7}
  & 69.5\diffdown{2.3}
  & 66.0\diffdown{2.6} \\
\rowcolor{teal!4}  \quad Supervised GRPO
  & 75.6\diffdown{2.8} & 36.1\diffdown{3.3} & 79.4\diffdown{9.7}
  & 51.8\diffdown{3.2} & 70.1\diffdown{4.4} & 56.0\diffdown{0.5}
  & 61.5\diffdown{4.0}
  & 56.8\diffdown{8.4} & 76.7\diffdown{3.0} & 68.7\diffdown{1.8}
  & 67.4\diffdown{4.4}
  & 64.4\diffdown{4.2} \\
\midrule
\rowcolor{blue!12} \our{}-Qwen3-VL-8B (Real VQA)
  & \underline{75.8} & \underline{40.8} & 86.2 & 54.1 & 73.2 & \textbf{59.6} & 65.0
  & 61.5 & 79.3 & \textbf{71.5} & 70.8 & 67.9 \\
\rowcolor{blue!3}  \quad Direct GRPO
  & 71.9\diffdown{3.9} & 37.1\diffdown{3.7} & 79.8\diffdown{6.4}
  & 51.0\diffdown{3.1} & 72.1\diffdown{1.1} & 58.3\diffdown{1.3}
  & 61.7\diffdown{3.3}
  & 53.7\diffdown{7.8} & 77.7\diffdown{1.6} & 57.2\diffdown{14.3}
  & 62.9\diffdown{7.9}
  & 62.3\diffdown{5.6} \\
\rowcolor{blue!3}  \quad Supervised GRPO
  & 70.6\diffdown{5.2} & 37.6\diffdown{3.2} & 77.0\diffdown{9.2}
  & 53.0\diffdown{1.1} & 71.2\diffdown{2.0} & \underline{58.9}\diffdown{0.7}
  & 61.4\diffdown{3.6}
  & 54.2\diffdown{7.3} & 76.3\diffdown{3.0} & 58.8\diffdown{12.7}
  & 63.1\diffdown{7.7}
  & 62.2\diffdown{5.7} \\
\bottomrule
\end{tabular}%
}
\vspace{-5pt}
\label{tab:main_results_table}
\end{table*}

%% file: figures/fig-combined-figure.tex
\begin{figure}[t]
  \centering
  \includegraphics[width=\linewidth]{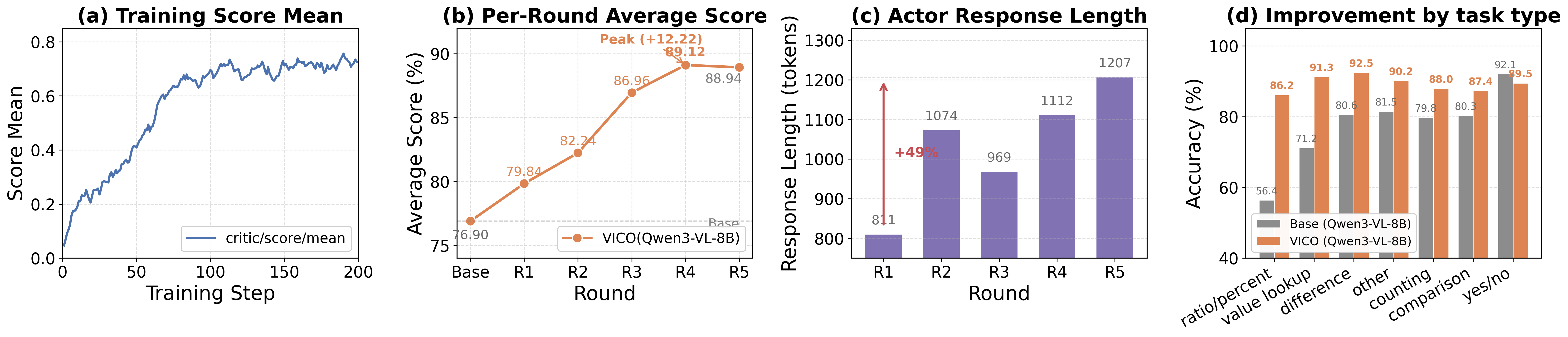}
  \caption{\textbf{\our{} trains stably and induces an emergent curriculum.}
  (a) GRPO reward improves smoothly. (b) ChartQA accuracy peaks at Round 4 with a $+12.2$ gain. (c) Actor responses grow by $+49\%$, suggesting harder tasks and longer reasoning. (d) Gains concentrate on reasoning-heavy question types.}
  \label{fig:combined_figure}
\end{figure}

%% file: tables/tab-rewriter-init.tex
\begin{wraptable}{r}{0.5\linewidth}
\centering
\vspace{-3ex}
\caption{%
\textbf{Rewriter initialization is not enough; co-evolution unlocks the full gain.}
Actor SFT (w/o co-training) and a vanilla EnvRewriter give modest gains ($+2.5\%$ and $+1.7\%$), while an SFT-initialized EnvRewriter optimized in the actor--environment loop reaches $+12.2\%$.
}
\label{tab:rewriter_init}
\small
\begin{tabular}{lc}
\toprule
\textbf{Baselines} & \textbf{ChartQA ($\Delta$)} \\
\midrule
Base (Qwen3-VL-8B)
& 76.9 \\
Actor SFT only
& 79.4 \diffup{2.5} \\
Co-evolution (w/ Vanilla Re-writer)
& 78.6 \diffup{1.7} \\
\rowcolor{gray!10}
Co-evolution (\our)
& \textbf{89.1} \diffup{12.2} \\
\bottomrule
\end{tabular}
\vspace{-1ex}
\end{wraptable}

%% file: figures/fig-saturation-evidence.tex

%% file: figures/fig-bidirectional-calibration.tex

\begin{figure}[t]
  \centering
  \begin{subfigure}[b]{0.45\linewidth}
    \centering
    \includegraphics[width=0.99\linewidth]{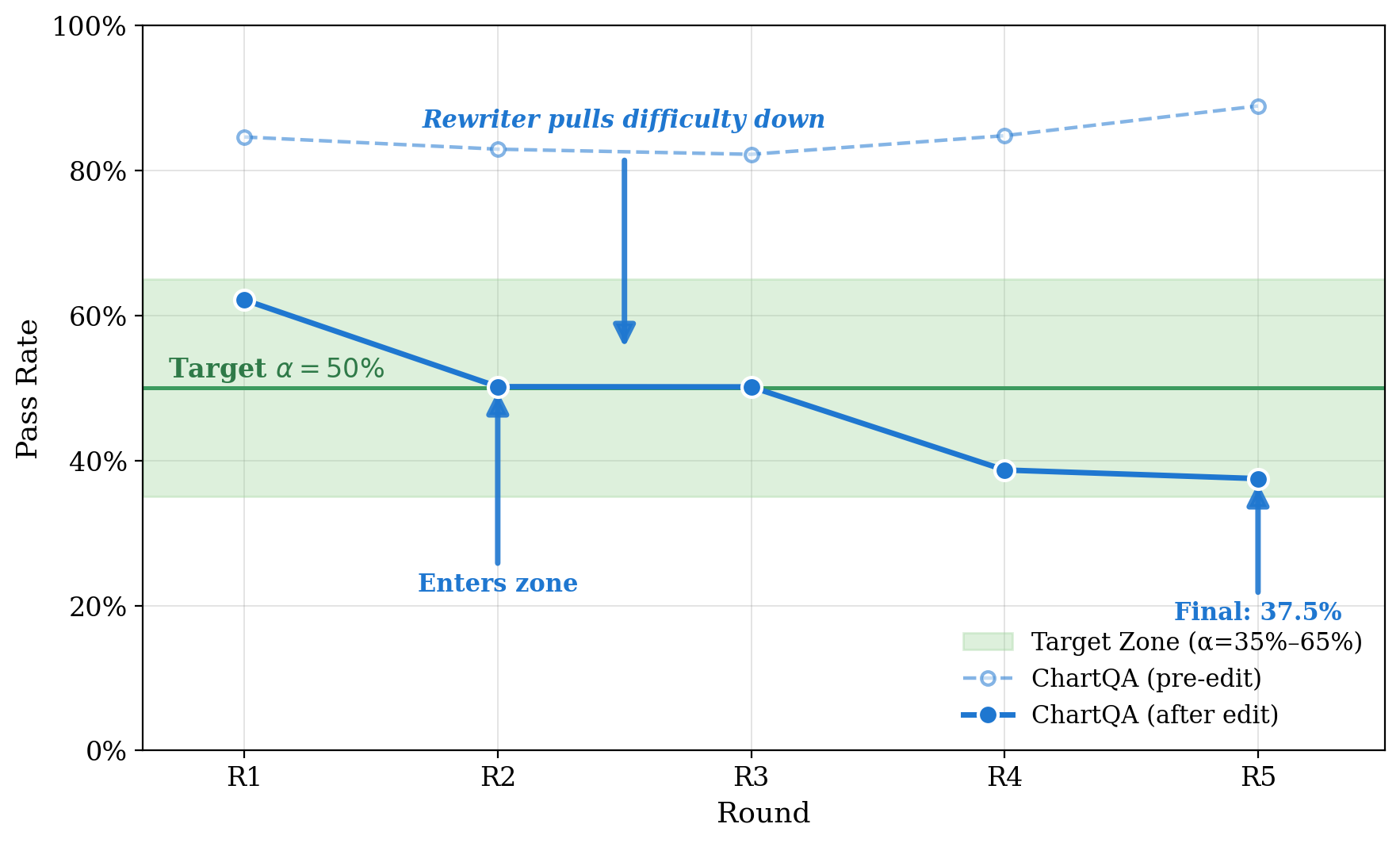}
    \caption{Strong ChartQA Actor: edits lower pass rate toward the sweet zone.}
    \label{fig:bidirectional_chartqa}
  \end{subfigure}
  \hfill
  \begin{subfigure}[b]{0.45\linewidth}
    \centering
    \includegraphics[width=0.99\linewidth]{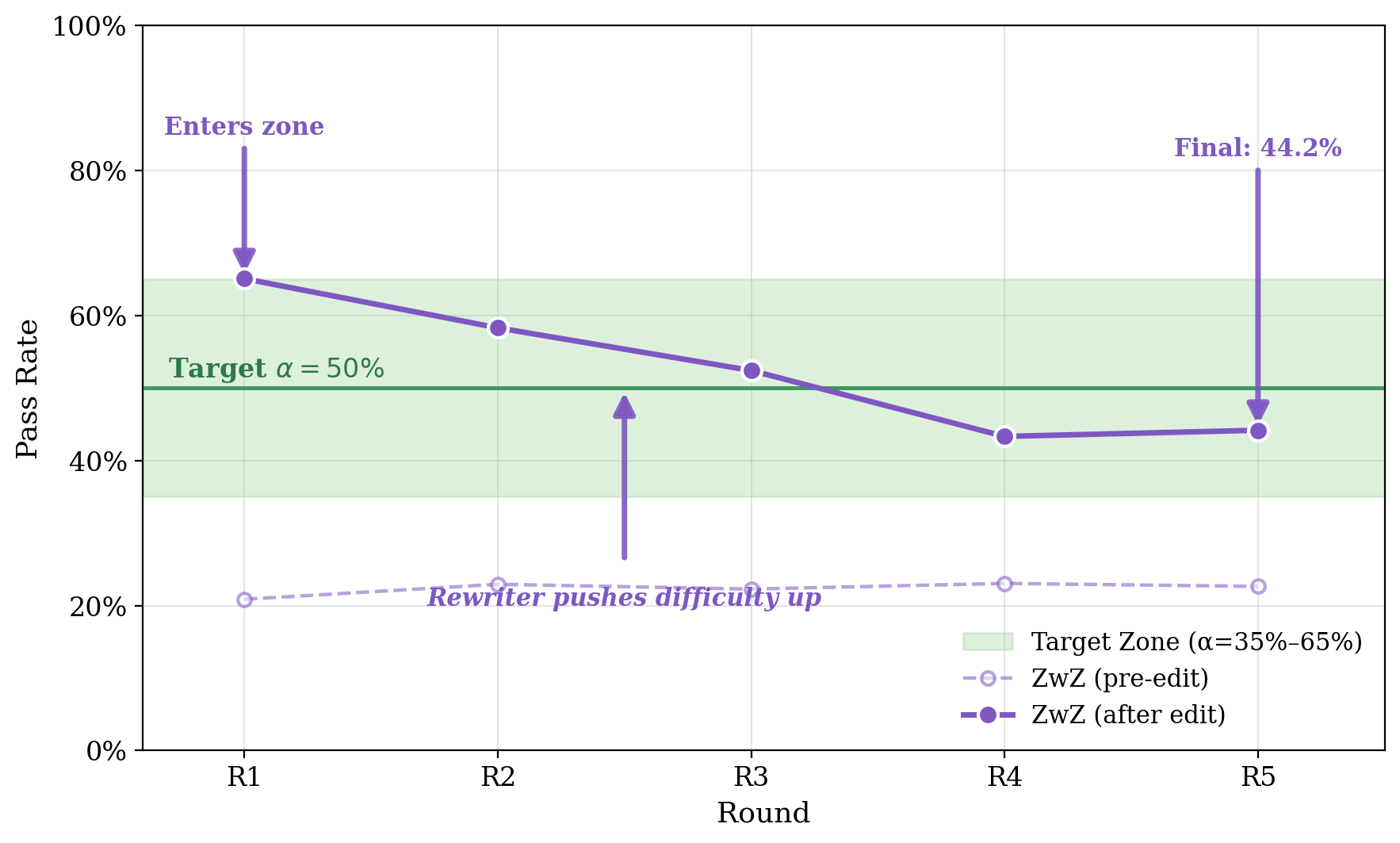}
    \caption{Weak ZwZ Actor: edits raise pass rate toward the sweet zone.}
    \label{fig:bidirectional_zwz}
  \end{subfigure}
  \caption{
    \textbf{The EnvRewriter calibrates bi-directional difficulty.} 
    On ChartQA, edits lower a strong Actor's pass rate; on ZwZ, edits raise a weak Actor's pass rate, moving both toward the sweet zone.}
  \label{fig:bidirectional_calibration}
\end{figure}


%% file: tables/tab-20-noise.tex
\begin{table}[t]
\centering
\caption{\textbf{\our{} remains robust under imperfect editing.} Training with 20\% failed-edit images causes only a slight drop, showing that the co-evolution loop does not need a flawless image editor.}
\label{tab:20-noise}
\resizebox{\textwidth}{!}{%
\begin{tabular}{lcccccccccc c}
\toprule
\textbf{Baselines ($\downarrow$)}
 & \multicolumn{6}{c}{\textbf{Mathematical Reasoning}}
 & \multicolumn{4}{c}{\textbf{Vision-Centric}}
 & \textbf{Overall} \\
\cmidrule(lr){2-7} \cmidrule(lr){8-11}
 & MathVista & MathVision & ChartQA & MathVerse & WeMath & avg.
 & MMMU & MMVP & BLINK & avg.
 &  avg.\\
\midrule
Qwen3-VL-8B
 & 67.7 & 31.5 & 76.9 & 50.6 & 72.0 & 59.8
 & 55.8 & 77.7 & 68.9 & 67.5
 & 63.6 \\
Round1 (clean)
 & 73.6 & 35.9 & 85.2 & 53.6 & 74.0 & 64.5
 & 60.2 & 78.3 & 69.0 & 69.2
 & 66.8 \\
\rowcolor{blue!8}
\textbf{Round1 (noise-20\%)}
 & \textbf{72.9} & \textbf{35.5} & \textbf{84.8} & \textbf{52.2} & \textbf{72.7} & \textbf{63.6}
 & \textbf{58.4} & \textbf{78.0} & \textbf{68.9} & \textbf{68.5}
 & \textbf{66.0} \\
\bottomrule
\end{tabular}
}
\end{table}

%% file: tex/6-conclusion.tex
\section{Conclusion}
\label{sec:conclusion}
We present \our{}, a co-evolution framework that reframes visual reasoning training as a dynamic, data-evolving co-evolution between an actor and an EnvRewriter, rather than a model-evolving process over a static corpus. 
By repurposing the rollout pass rate as a shared learning signal, \our{} jointly trains an actor and an EnvRewriter that edits verifiable visual substructures toward the actor's learning frontier, producing difficulty-calibrated training samples without any additional human annotation. 
Across nine multimodal benchmarks, \our{} consistently outperforms strong RLVR, self-evolution, and co-evolution baselines, generalizes to out-of-distribution tasks unseen during training, and remains stable across multiple VLM backbones, exceeding the strongest self-evolution baseline by +4.3\% on out-of-domain tasks while using up to 160× fewer labeled samples. 
Together, these results suggest that the bottleneck in visual reasoning post-training is \emph{not the size of the static corpus but the fixed nature of the environment itself}, and that difficulty-aligned visual editing offers a scalable path forward in domains where high-quality human annotation is costly to obtain.



%% file: tex/5-appendix-draft.tex
\appendix
\newpage

\renewcommand{\topfraction}{0.95}
\renewcommand{\bottomfraction}{0.85}
\renewcommand{\textfraction}{0.05}
\renewcommand{\floatpagefraction}{0.85}
\setlength{\textfloatsep}{8pt plus 2pt minus 2pt}
\setlength{\floatsep}{6pt plus 2pt minus 2pt}
\setlength{\intextsep}{6pt plus 2pt minus 2pt}



\newpage

\section{Limitations, Future Work, and Impact Statement}
\label{sec:limitation}

\paragraph{Limitations}

Although \our{} shows strong empirical gains across multimodal reasoning benchmarks, several limitations remain.
First, \our{} relies on a verifiable image-side structure for editing. In CLEVR and ChartQA, this structure is provided by scene graphs or chart tables; in natural-image VQA, it is approximated by protected region masks. Extending the framework to visual domains without such structures or reliable region annotations remains an important direction.

Second, \our{} introduces additional compute costs because each round estimates actor pass rates with multiple rollouts and updates both the actor and the EnvRewriter. This cost is higher than that of one-shot static augmentation, although it enables the training distribution to adapt to the actor's evolving capabilities.

Third, our evaluation focuses on visual reasoning tasks with verifiable answers. Open-ended generation, dialog, and real-time interactive environments are outside the current scope. Future work can study adaptive difficulty targets, broader visual domains, and tighter integration with more general-purpose image editors.

\paragraph{Future work}
Future work can extend \our{} to more complex multimodal tasks beyond static
image reasoning. We plan to broaden the framework to tasks that require
richer visual grounding, multi-image comparison, diagram reasoning, and
tool-assisted multimodal problem solving. These settings require the environment
to edit not only visual appearance but also higher-level task structure while
preserving verifiable supervision.

\paragraph{Impact Statement}
\our{} aims to reduce the reliance on large-scale human annotation for VLM post-training by turning existing visual data into adaptive, label-valid training samples. This may make visual reasoning training more scalable and accessible, especially in domains where high-quality annotations are expensive. At the same time, automatic image editing can introduce artifacts, biases, or label drift if used without verification. Our framework mitigates these risks through deterministic relabeling, answer-preserving constraints, and filtering, but careful validation remains necessary before applying similar systems to high-stakes domains.

\section{Implementation Details}
\label{appendix:training-details}

\subsection{\our{}: Algorithm Details}
\label{appendix:algorithm}

 \begin{algorithm}[h]
 \caption{\textbf{\our{} jointly updates the Actor and EnvRewriter from shared rollouts.} Each round generates edited tasks, evaluates Actor rollouts, updates the Actor with task rewards, and updates the EnvRewriter with difficulty-calibration rewards.}
 \label{alg:coevolve}
 \small
 \begin{algorithmic}
 \footnotesize
 \STATE {\bfseries Require:} Actor $\pi_a$, EnvRewriter $\pi_R$, seed dataset $\mathcal{D}_{\text{seed}} = \{(x_i, q_i, a_i, \sigma_i)\}$, number of rounds $T$, rollouts per sample $K$, target pass rate $\alpha$, difficulty filter threshold $k_{\min}$, RWR temperature $\lambda$
\STATE {\bfseries Ensure:} Trained actor $\pi_a$, EnvRewriter $\pi_R$
 \STATE Initialize actor pool $\mathcal{D}_a \gets \varnothing$, EnvRewriter pool $\mathcal{D}_R \gets \varnothing$
 \FOR{round $t = 1, \dots, T$}
     \STATE \textcolor{blue}{\textbf{Stage 1: Online Generation \& Interaction}}
     \STATE Sample edit batch: $\mathcal{T}_t \gets \{(x'_i, q'_i, a'_i)\}$ where $u_i \sim \pi_R(\cdot \mid x_i, \sigma_i)$, $x'_i = \mathcal{R}(u_i(\sigma_i))$
     \FOR{each edited triple $(x'_i, q'_i, a'_i) \in \mathcal{T}_t$}
         \STATE Sample $K$ rollouts $\{\hat{a}_{i,k}\}_{k=1}^K$ from $\pi_a(\cdot \mid x'_i, q'_i)$
         \STATE Compute actor reward $R_{\text{actor}}(\hat{a}_{i,k}, a'_i)$ for each rollout
         \STATE Compute pass rate $\hat{p}_i \gets k_i / K$
         \STATE Compute EnvRewriter reward $R_{\text{rewriter}}(\hat{p}_i)$
     \ENDFOR
     \STATE \textcolor{blue}{\textbf{Stage 2: Joint Update}}
     \STATE Update $\pi_a$ via GRPO on $\mathcal{T}_t$ to maximize $\mathbb{E}[R_{\text{actor}}]$
     \STATE Filter: $\tilde{\mathcal{T}}_t \gets \{(x_i, \sigma_i, u_i) : |\hat{p}_i - \alpha| \leq k_{\min}\}$
     \STATE Build weighted SFT set: $\tilde{\mathcal{D}}_R^t \gets \{(x_i, \sigma_i, u_i, w_i) : w_i \propto \exp(\lambda R_{\text{rewriter}}(\hat{p}_i))\}$
     \STATE Update $\pi_R$ via RWR on $\tilde{\mathcal{D}}_R^t$ to maximize $\mathbb{E}[R_{\text{rewriter}}]$
     \STATE \textcolor{blue}{\textbf{Stage 3: Aggregation}}
     \STATE $\mathcal{D}_a \gets \mathcal{D}_a \cup \text{Valid}(\mathcal{T}_t)$
     \STATE $\mathcal{D}_R \gets \mathcal{D}_R \cup \tilde{\mathcal{D}}_R^t$
 \ENDFOR
 \end{algorithmic}
 \end{algorithm}

\our{} trains an Actor and an EnvRewriter in an iterative actor--environment loop. The Actor learns from verifiable rollout rewards, while the EnvRewriter learns from the Actor's empirical pass rate on its edited samples. The two models share a common pool of rollouts: each rollout simultaneously yields a task-level reward used to update the Actor and a difficulty signal used to update the EnvRewriter, enabling co-evolution without doubling the inference cost. The complete training loop is summarized in Algorithm~\ref{alg:coevolve} and described below.

\paragraph{Inputs and initialization}
The pipeline starts from a seed visual dataset $\mathcal{D}_{\text{seed}}=\{(x,q,a,\sigma)\}$, where $x$ is the image, $q$ is the question, $a$ is the answer, and $\sigma$ is a verifiable substructure when available, such as a CLEVR scene graph or a ChartQA data table. We initialize the Actor $\pi_a^{(0)}$ and the EnvRewriter $\pi_R^{(0)}$ from the same pretrained VLM family. The EnvRewriter is first SFT-initialized to produce executable edit commands. Four hyperparameters control the loop: the number of rounds $T$, the rollouts per sample $K$, the target pass rate $\alpha$ (we use $\alpha = 0.5$), the difficulty filter threshold $k_{\min}$ that discards edits whose pass rate is too far from $\alpha$, and the RWR temperature $\lambda$ that controls how sharply the EnvRewriter is biased toward high-reward edits. Two online pools, the Actor pool $\mathcal{D}_a$ and the EnvRewriter pool $\mathcal{D}_R$, are initialized empty and grow throughout training.

\paragraph{Stage 1: Online generation and interaction}
At the start of each round $t$, the EnvRewriter proposes a structured edit command $u_i \sim \pi_R(\cdot \mid x_i, \sigma_i)$ over the verifiable substructure $\sigma_i$ for every seed instance. A domain renderer or editor produces the edited image $x'_i=\mathcal{R}(u_i(\sigma_i))$, and the answer $a'_i$ is either computed programmatically from the edited structure or preserved by construction under the domain-specific validity rules in Appendix~\ref{appendix:programmatic-data-generation}. Thus, every retained sample is label-valid without model-based labeling. For each edited triple $(x'_i, q'_i, a'_i)$, the Actor samples $K$ responses, and each response receives the rule-based Actor reward defined in Appendix~\ref{appendix:reward_details}. We then compute the empirical pass rate $\hat{p}_i = k_i / K$, where $k_i$ is the number of correct rollouts, and map it to the difficulty-calibration reward $R_{\text{rewriter}}$, which peaks when $\hat{p}_i$ matches the target $\alpha$.

\paragraph{Stage 2: Joint update}
The two models are updated from the same batch of rollouts, but with different reward signals. Valid edited triples $(x'_t,q'_t,a'_t)$ are used to update the Actor $\pi_a$ via GRPO to maximize $\mathbb{E}[R_{\text{actor}}]$, leveraging the within-group advantage normalization that makes GRPO robust to scale differences across edits. The EnvRewriter $\pi_R$ is updated via Reward-Weighted Regression (RWR): we first apply a difficulty filter, retaining only edits whose pass rate falls within $[\alpha - k_{\min}, \alpha + k_{\min}]$, and then construct a weighted SFT set in which each retained edit transition is weighted by $\exp(\lambda R_{\text{rewriter}})$. Fine-tuning $\pi_R$ on this set biases it toward edits that land close to the target band. We additionally apply a KL penalty to the initial EnvRewriter and cap the per-step KL to prevent the editor from drifting into degenerate or off-distribution edit distributions.

\paragraph{Stage 3: Aggregation}
After each round, valid edited triples (those that render successfully and pass label-consistency checks) are appended to the Actor pool $\mathcal{D}_a$, while the weighted edit transitions are appended to the EnvRewriter pool $\mathcal{D}_R$. Both pools persist across rounds, so the Actor sees an increasingly diverse set of edited tasks, and the EnvRewriter learns from an increasingly informative history of difficulty signals. The updated Actor and EnvRewriter then become the starting point for the next round.


\paragraph{Why this differs from static RLVR?}
The key distinction from static RLVR is that the empirical pass rate $\hat{p}$ is not used only as a filtering statistic. It is also reused as an actor-relative supervision signal for the EnvRewriter, allowing the visual environment to track the Actor's current capability. Crucially, the loop requires no new human-labeled QA: the seed representation $\sigma_i$ provides the editable interface, and labels for edited samples are either recomputed or preserved through explicit validity constraints.

\subsection{Actor GRPO Training Details}
\label{appendix:actor-grpo}

For the Actor $\pi_a$, we use GRPO with rule-based rewards. For each edited prompt, we sample multiple on-policy rollouts and score them with exact-match answer verification plus a format check. Unless otherwise stated, rollout sampling uses temperature $0.7$, top-$p=1.0$, and a maximum response length of $3{,}824$ tokens, matching the zero-shot evaluation setting in the main text.

In the ChartQA multi-round runs, each round trains on approximately $1500$--$2500$ EnvRewriter-generated modified training images, together with a fixed regularization set of $2{,}000$ original ChartQA samples. The resulting Actor checkpoint initializes the next round.

\subsection{EnvRewriter SFT and RWR Details}
\label{appendix:rewriter-training}

The EnvRewriter is trained in two stages. First, we initialize it with
format-aware SFT on verified edit demonstrations generated by GPT-5.
This stage teaches the model to produce structured, executable edit commands, use the available edit operations, and propose edits that adjust task difficulty while preserving label verifiability. Second, during co-evolution, we update the learned EnvRewriter using reward-weighted regression over valid edit transitions collected through interaction with the Actor.

For each executed edit, the current Actor produces $K$ rollouts on the resulting sample. The empirical pass rate $\hat{p}$ is used to compute the edit's RWR weight according to the bell-shaped difficulty objective in Appendix~\ref{appendix:reward_details}. These weights are computed from the outcome of each edit turn. Invalid or unexecutable edits, as well as edits outside the RWR difficulty band, are excluded from the RWR target set. This encourages the EnvRewriter to concentrate its probability mass on valid edits near the Actor's learnable zone. Actor training data are constructed separately: samples that satisfy execution and label-validity requirements may be retained even when their pass rates fall outside the RWR difficulty band.

\paragraph{GPT-5 as a strategic trajectory generator.}
We use GPT-5 to generate editing demonstrations for training the EnvRewriter. Given a seed sample, its editable structure, the current Actor's empirical pass rate, the target difficulty band, and the domain-specific edit schema, GPT-5 proposes a structured edit command intended to adjust the task's difficulty. The command is executed by the environment, and the resulting sample is evaluated by the Actor. The execution outcome and Actor feedback can then be included in the context for subsequent edits, producing actor-conditioned editing trajectories.
GPT-5 provides the edit proposals, but it does not serve as the authority for either ground-truth labels or training rewards. Labels are obtained through domain-specific recomputation or checked for answer preservation, depending on the edit setting, while difficulty is measured using Actor rollouts. We construct the SFT dataset from executable, verified demonstrations whose edits move the Actor's pass rate closer to the target pass rate. These demonstrations supervise both the edit-command format and the choice of editing strategies.
  
\paragraph{EnvRewriter interface.}
The EnvRewriter is an LVLM policy initialized from the same model family as the Actor and SFT-initialized to produce executable edit commands. Its input
contains the image, question, answer, editable structure $\sigma$, current
Actor pass rate $\hat{p}$, target pass rate $\alpha$, and the domain-specific tool schema. Its output is a structured tool call with fields
\texttt{tool}, \texttt{params}, and \texttt{reasoning}. For example, in
ChartQA, the EnvRewriter may output a \texttt{chart\_rerender} command that
edits the underlying CSV table and re-renders the chart. The command is
executed by the domain renderer or image editor, and the resulting sample is
kept only if it passes the validity checks in Table~\ref{tab:verifiable_editing_protocol}.
Full prompts and tool schemas are provided in Appendix~\ref{appendix:rewriter-prompt}.

\subsection{Reward Functions and Difficulty Filtering}
\label{appendix:reward_details}
\label{appendix:reward-details}

\paragraph{Actor Reward.}
The Actor reward is defined as:
\begin{equation}
R_{\text{actor}}(q', o) =
\begin{cases}
+1, & \text{if } \mathrm{FormatCorrect}(o) \wedge \mathrm{AnswerCorrect}(o), \\
-1, & \text{otherwise}.
\end{cases}
\label{eq:actor_reward_appendix}
\end{equation}

\paragraph{EnvRewriter Reward.}
For each executed edit, the empirical pass rate on the resulting sample is $\hat{p} = k/K$, where $k$ is the number of correct Actor rollouts under $R_{\text{actor}}$. The EnvRewriter reward is:
\begin{equation}
R_{\text{rewriter}}(\hat{p}) = \exp\left(-\beta(\hat{p} - \alpha)^2\right),
\label{eq:rewriter_reward_appendix}
\end{equation}
where $\beta > 0$ controls sharpness. We use $\alpha = 0.5$, $\beta = 4.0$, and $k_{\min} = 0.15$ for the difficulty filter.

\paragraph{Difficulty filter.}
The reward encourages edits whose resulting samples have pass rates near the Actor's learning frontier. In practice, we keep a sweet-zone band around $\alpha=0.5$ and discard invalid or poorly calibrated edits from EnvRewriter updates. This prevents the EnvRewriter from overfitting to samples that are already solved or impossible for the current Actor.

\subsection{Hyperparameters}
\label{appendix:hyperparameters}

\begin{table}[h]
\caption{\textbf{\our{} uses one compact training recipe.} Unless stated otherwise, the same core hyperparameters are used across visual domains.}
\label{tab:hyperparameters_appendix}

\centering
\footnotesize
\setlength{\tabcolsep}{7pt}
\renewcommand{\arraystretch}{1.2}
\begin{tabular}{@{}ll@{}}
\toprule
\textbf{Component} & \textbf{Value} \\
\midrule
Actor optimizer & AdamW \\
Actor update & GRPO \\
Actor learning rate & $1\times10^{-6}$ \\
Rollouts per sample & $K=8$ for pass-rate estimation \\
Sampling temperature & $0.7$ \\
Top-$p$ & $1.0$ \\
Maximum response length & $3{,}824$ tokens \\
EnvRewriter update & Reward-weighted LoRA SFT / RWR \\
Difficulty target & $\alpha=0.5$ \\
Reward sharpness & $\beta=4.0$ \\
Sweet-zone band & $[0.35,0.65]$ unless otherwise stated \\
Training rounds & Up to $5$ rounds \\
Hardware & Up to $8\times$A100-80GB GPUs \\
Training framework & verl-based distributed RL training \\
\bottomrule
\end{tabular}
\end{table}



\begin{table}[ht]
\caption{\textbf{RWR trains the EnvRewriter with pass-rate-weighted edit transitions.}
We train LoRA on Qwen3VL-8B-Instruct using LlamaFactory. Each retained edit transition is duplicated $\lfloor 5 \cdot w(\hat{p}) \rfloor$ times based on the Actor pass rate $\hat{p}$ on the sample produced by that edit turn.
}
\label{tab:sft_rewriter_parameters}

\centering
\footnotesize
\setlength{\tabcolsep}{4pt}
\renewcommand{\arraystretch}{1.05}
\begin{tabular}{@{}p{0.27\linewidth}p{0.18\linewidth}p{0.27\linewidth}p{0.18\linewidth}@{}}
\toprule
\multicolumn{2}{c}{\textit{Model \& LoRA}} & \multicolumn{2}{c}{\textit{Optimization}} \\
\midrule
base model & Qwen3VL-8B-Instruct & learning rate & $5.0\times10^{-6}$ \\
template & qwen3\_vl & scheduler & cosine \\
finetuning type & LoRA & warmup ratio & 0.1 \\
LoRA rank & 64 & total epochs & 10 \\
LoRA alpha & 128 & per-device batch & 2 \\
LoRA target & all & grad. accum. steps & 8 \\
precision & bf16 & effective batch & 64 \\
\midrule
\multicolumn{2}{c}{\textit{Eval \& Checkpointing}} & \multicolumn{2}{c}{\textit{RWR Data Construction}} \\
\midrule
eval strategy & steps & RWR weight $\alpha$ & 0.5 \\
eval/save steps & 50 / 50 & RWR weight $\beta$ & 4.0 \\
save total limit & 3 & max sample repeats & 5 \\
load best model & true & weight formula & $\exp(-\beta(\hat{p}-\alpha)^2)$ \\
best metric & eval\_loss & number of GPUs & 4 \\
\bottomrule
\end{tabular}
\end{table}

\paragraph{RL training parameters.}
We  optimize the actor with reinforcement learning to improve multi-step visual reasoning.

\begin{table}[ht]
\caption{\textbf{GRPO trains the Actor with rollout-level verifiable rewards.}
We train Qwen3VL-8B-Instruct with verl and a vLLM rollout backend. We disable KL penalty in reward shaping and loss to avoid gradient interference, following~\citep{shao2024deepseekmathpushinglimitsmathematical}.
}
\label{tab:grpo_solver_parameters}

\centering
\footnotesize
\setlength{\tabcolsep}{4pt}
\renewcommand{\arraystretch}{1.05}
\begin{tabular}{@{}p{0.27\linewidth}p{0.18\linewidth}p{0.27\linewidth}p{0.18\linewidth}@{}}
\toprule
\multicolumn{2}{c}{\textit{Model \& Reward}} & \multicolumn{2}{c}{\textit{Rollout \& Batch}} \\
\midrule
base model & Qwen3VL-8B-Instruct & rollouts per prompt, $N$ & 8 \\
algorithm & GRPO & train batch size & 32 \\
reward manager & binary & PPO mini-batch & 8 \\
KL in reward & False & PPO micro-batch/GPU & 1 \\
KL loss coeff. & 0 & log-prob micro-batch/GPU & 1 \\
entropy coeff. & 0 & rollout backend & vLLM \\
 & & tensor model parallel & 1 \\
 & & sampling temperature & 0.7 \\
\midrule
\multicolumn{2}{c}{\textit{Optimization}} & \multicolumn{2}{c}{\textit{Sequence Limits \& Schedule}} \\
\midrule
actor learning rate & $1.0\times10^{-6}$ & max prompt length & 1{,}048 \\
scheduler & cosine & max response length & 3{,}824 \\
warmup ratio & 0.05 & max model length & 6{,}896 \\
total epochs & 3 & max batched tokens & 16{,}392 \\
precision & bf16 & truncation & left \\
grad. checkpointing & enabled & filter overlong prompts & True \\
FSDP strategy & FSDP2 & save / val freq & 50 / 20 \\
sequence parallel size & 1 & validate before training & True \\
number of GPUs & 8 & & \\
\bottomrule
\end{tabular}
\end{table}

\subsection{Compute and Hardware}
\label{appendix:compute}

All experiments use distributed training on up to $8\times$A100-80GB GPUs. We use a VERL-based implementation for GRPO training and parallel rollout generation. A detailed cost and annotation-efficiency comparison is provided in Appendix~\ref{appendix:cost-efficiency}.

\section{Data Sources and Verifiable Visual Editing}
\label{appendix:data-editing}
\label{appendix:programmatic-data-generation}

This appendix describes how \our{} produces label-valid training samples across three visual domains.
The key point is that different domains provide different forms of verifiability.

For CLEVR, \our{} edits an explicit scene graph and recomputes the answer
programmatically by executing the original question program. For ChartQA,
\our{} edits the underlying CSV table and uses a two-tier verification rule:
Tier~1 recovers the new answer by entity-anchored lookup in the edited table,
while Tier~2 keeps CSV values fixed and preserves the original answer by
construction. For natural-image VQA, where no complete symbolic state is
available, \our{} uses answer-preserving regional editing: the question and
answer remain fixed, while the EnvRewriter edits only non-target visual regions.

Thus, \our{} uses two label-validity mechanisms: deterministic recomputation
when the answer can be derived from the edited structure, and constrained answer
preservation when the answer-bearing structure must remain fixed.

\subsection{Seed Data Sources}
\label{appendix:seed-data}

\our{} uses a small seed set to initialize each visual reasoning domain.
For compositional reasoning, we sample CLEVR images with their scene graphs and question programs.
For chart reasoning, we sample ChartQA training images together with their underlying data tables.
For natural-image VQA, we sample ZwZ images, which provide answer-relevant region annotations. We use these regions to identify target objects and preserve the original question-answer pair during editing.



\subsection{Overview of the Three Editing Arms}
\label{appendix:editing-overview}
\label{appendix:editing-arms}

Table~\ref{tab:editing-arms} summarizes the three editing arms.
CLEVR supports full deterministic answer recomputation from the edited scene graph. ChartQA uses a two-tier table rule: Tier~1 updates the answer by entity-anchored lookup in the edited CSV table, while Tier~2 preserves the answer under constrained table edits. The natural-image VQA arm follows a different rule: it keeps the original question and answer unchanged and only edits regions outside the answer-relevant target objects.

\begin{table}[h]
\caption{%
\textbf{\our{} uses three editing arms for verifiable visual variation.}
CLEVR recomputes answers from scene graphs, ChartQA uses a two-tier table rule, and ZwZ preserves answers by protecting target objects.
}
\label{tab:editing-arms}

\centering
\footnotesize
\setlength{\tabcolsep}{4pt}
\renewcommand{\arraystretch}{1.08}
\begin{tabular}{@{}p{0.12\linewidth}p{0.39\linewidth}p{0.39\linewidth}@{}}
\toprule
\textbf{Arm} & \textbf{Editing interface} & \textbf{Answer rule} \\
\midrule
CLEVR & Scene graph edits over objects, attributes, and relations; rendered with Blender. & Recompute $a'$ by re-executing the question program. \\
ChartQA & CSV/table edits rendered with matplotlib. & Tier~1: entity-anchored lookup; Tier~2: keep $a'=a$ under constrained edits. \\
ZwZ & SAM3 masks with protected target objects; edit only non-target regions. & Keep $q'=q$ and $a'=a$ under target-object protection. \\
\bottomrule
\end{tabular}
\end{table}

\subsection{CLEVR: Scene-Graph Editing}
\label{appendix:clevr-editing}

For CLEVR, the editable structure $\sigma$ is the scene graph paired with each image.
Each scene graph specifies a set of objects, with each object carrying attributes such as \texttt{shape}, \texttt{color}, \texttt{material}, and \texttt{size}, as well as pairwise spatial relations such as \texttt{left-of}, \texttt{right-of}, \texttt{front}, and \texttt{behind}.
Questions are generated from question-program templates over the scene graph, so the same program can be re-executed on the edited graph to produce $a'$.

\paragraph{Edit operations.}
The EnvRewriter outputs an edit command $u$ in one of four categories:
(i) \textbf{attribute change}, which modifies the \texttt{color}, \texttt{material}, or \texttt{size} of a target object;
(ii) \textbf{category swap}, which changes an object's \texttt{shape} among \texttt{cube}, \texttt{sphere}, and \texttt{cylinder};
(iii) \textbf{entity add/remove}, which inserts or deletes objects while preserving renderability; and
(iv) \textbf{relation modification}, which repositions objects to alter spatial relations.
Each operation acts on the scene graph rather than on pixels, so $u(\sigma)$ remains a fully specified renderable scene.

\paragraph{Renderer.}
We use Blender with the original CLEVR rendering script to produce $x'=\mathcal{R}(u(\sigma))$.
Camera parameters, lighting, and material shaders are kept consistent with the original rendering pipeline to avoid stylistic drift.

\paragraph{Answer derivation.}
We retain the original question program and re-execute it on $u(\sigma)$ to obtain $a'$.
For example, if the original program is \texttt{count(filter(color=red, shape=cube))} and the edit changes one red cube into a blue cube, the new answer decreases by one.
Because program execution on a discrete scene graph is exact, CLEVR samples are label-valid by construction.

\paragraph{Failure mode.}
An edit is discarded if it makes the question ambiguous or non-executable.
For example, if an edit creates multiple possible referents for a question that assumes a unique object, the sample is removed before training.

\subsection{ChartQA: Table-Level Editing and Re-rendering}
\label{appendix:chart-editing}

For chart reasoning, the editable structure $\sigma$ is the CSV table
underlying each chart. The EnvRewriter edits the table-level representation
rather than chart pixels, and matplotlib re-renders the edited table into a new
chart image. This allows \our{} to adjust visual and numerical difficulty while
keeping label validity controlled by the table.

\paragraph{Tier construction.}
For each ChartQA seed sample, we first apply an automatic label-to-cell matching
heuristic. We scan the CSV table and match value cells against the original
answer. If exactly one cell matches, the sample is assigned to Tier~1. Let
$(r^\star,j^\star)$ denote the matched answer cell, and let $e^\star$ denote
the row entity at $r^\star$ in the original table. If no cell matches, or
multiple cells match, the sample is assigned to Tier~2. Tier~2 often includes
derived-answer cases such as ratios, percentages, or differences, or ambiguous
duplicate values.

\paragraph{Edit operations.}
For Tier~1, the EnvRewriter may modify table values, add near-match distractor
rows, increase chart density, or sharpen contrasts. After editing, the answer is
recovered by entity-anchored lookup in the edited table:
\[
a'=\mathrm{table}'[\mathrm{entity}=e^\star, j^\star].
\]
This lookup avoids relying on the original row index after row insertion or
reordering. For Tier~2, the EnvRewriter keeps CSV values fixed and only varies
rendering style or other answer-preserving visual factors. The label is
therefore kept as $a'=a$ by construction.

\paragraph{Renderer.}
We re-render charts with matplotlib using the same chart type and style
configuration as the source chart whenever possible, including chart type, color
palette, axis labels, fonts, and gridlines.

\paragraph{Failure mode.}
We discard edits that violate the table schema, remove or duplicate the Tier~1
anchor entity $e^\star$, remove the target column $j^\star$, produce no numeric
columns, fail rendering, or break the Tier~2 invariance constraint by changing
CSV values. Numerical edits are bounded to keep the rendered chart visually
plausible.

\subsection{ZwZ: Answer-Preserving Natural-Image Editing}
\label{appendix:zwz-editing}
\label{appendix:realvqa-pipeline}

For natural-image VQA, \our{} does not attempt to recompute a new answer from a full symbolic scene representation.
Instead, it preserves the original question-answer pair.
Given a seed sample $(x,q,a)$, the edited sample is $(x',q,a)$, where $q'=q$ and $a'=a$.

The key assumption is that the answer is grounded in a set of answer-relevant target objects.
ZwZ provides answer-relevant region annotations, and we use SAM3 to decompose the image into object-level masks.
Masks that overlap the answer-relevant region are treated as \textbf{target objects}.
These target objects form the protected region $\sigma^{+}$.
All other object masks and background regions form the editable complement $\sigma^{-}$.

\paragraph{Edit operations.}
The EnvRewriter receives the image, question, answer, object masks, protected target objects, and current Actor pass-rate signal.
It then proposes edits restricted to $\sigma^{-}$.
Typical edits include adding distractor objects outside the target region, modifying background clutter, changing the appearance of non-target objects, or introducing visually plausible distractors that do not alter the answer-relevant target objects.
The EnvRewriter is not allowed to edit, occlude, remove, or semantically change the protected target objects.

\paragraph{Rendering.}
The original image, the protected target-object mask, and the edit command are passed to an external image editor.
The editor modifies only the editable complement and returns the edited image $x'$.
Unlike CLEVR or ChartQA, this step is not treated as deterministic symbolic rendering.
It is an answer-preserving image-editing step under a protected-region constraint.

\paragraph{Answer rule.}
For ZwZ, the answer is not recomputed.
We keep the original question and answer:
\[
q' = q, \qquad a' = a.
\]
The Actor reward therefore compares the model's prediction against the original answer $a$.
This distinguishes natural-image VQA from the structured arms, where $a'$ is programmatically recomputed.

\paragraph{Failure mode.}
An edit is discarded if it touches the protected target objects, changes the visual evidence needed to answer the question, or introduces new content that would make the original answer ambiguous.
For example, if the question depends on counting target objects, adding another object of the same relevant category is rejected.
If the question depends on a spatial relation involving a target object, edits that change that relation are rejected.
This conservative filtering keeps the natural-image VQA arm answer-preserving rather than relabeled.

\subsection{Filtering Invalid or Ambiguous Edits}
\label{appendix:edit-filtering}

Across all editing arms, we discard edits that fail label-validity or execution checks.

First, the edited sample must be executable by the corresponding renderer or image editor.
For CLEVR, the scene graph must render successfully in Blender.
For ChartQA, the edited CSV must satisfy the table schema and render successfully with matplotlib.
For ZwZ, the image editor must preserve the protected target-object region.

Second, the edited sample must preserve a valid answer. For CLEVR, the question
program must be executable on the edited scene graph to compute $a'$. For
ChartQA Tier~1, the anchor entity $e^\star$ and target column $j^\star$ must
remain valid and addressable in the edited table so that
$a'=\mathrm{table}'[\mathrm{entity}=e^\star,j^\star]$ can be recovered by
lookup. For ChartQA Tier~2, the edit must satisfy the invariance constraint:
CSV values remain fixed and the original answer is kept as $a'=a$. For ZwZ, the
original answer remains valid because protected target regions are not edited.

Third, the edited sample must provide useful difficulty for the current Actor.
We measure this by the empirical pass rate $\hat{p}$ after Actor rollouts.
Invalid, ambiguous, unexecutable, or out-of-band edits are removed from the EnvRewriter update set.

\subsection{Dataset and Edited Sample Visualizations}
\label{appendix:dataset-examples}
\label{appendix:visualization-of-datasets}

Figures~\ref{fig:clevr_editing_examples}, \ref{fig:chartqa_editing_examples}, and \ref{fig:zwz_editing_examples} show representative visual examples for the three editing arms.
For CLEVR, edits are applied to a scene graph and the answer $a'$ is recomputed
by executing the original question program. For ChartQA, Tier~1 examples recover
$a'$ by entity-anchored lookup in the edited CSV table, while Tier~2 examples
preserve $a'=a$ through constrained table edits. For ZwZ, the original question
and answer are retained, while answer-relevant target objects are protected and
only non-target regions are edited.
Together, these examples show the two label-validity mechanisms used by \our{}: deterministic recomputation when an answer can be derived from the edited structure, and constrained answer preservation when the answer-bearing structure must remain fixed.

\subsection{Robustness to Tool Failures and Difficulty Alignment as the Bottleneck}
\label{appendix:tool-failure-discussion}

Renderer or editor failures may produce non-renderable scenes, invalid charts, or protected-region violations. In \our{}, these failures are filtered before EnvRewriter updates: invalid, label-breaking, or out-of-band edits contribute no gradient to $\pi_R$. This biases the EnvRewriter toward edit plans that are executable, label-valid, and useful for the current Actor. Appendix~\ref{appendix:editor-robustness} further stress-tests this assumption by corrupting 20\% of R1 training data and shows that performance drops by less than one point.

\section{Image Editing Case Studies}
\label{appendix: case study}


\subsection{CLEVR Scene-Graph Editing Examples}
\label{appendix:clevr-case-study}
\begin{figure*}[t]
  \centering

  \begin{subfigure}[t]{0.235\textwidth}
    \centering
    \includegraphics[width=\linewidth]{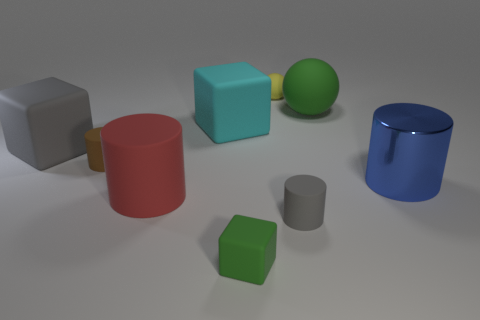}
    \caption{Case 1: original}
    \label{fig:clevr_case1_ori}
  \end{subfigure}
  \hfill
  \begin{subfigure}[t]{0.235\textwidth}
    \centering
    \includegraphics[width=\linewidth]{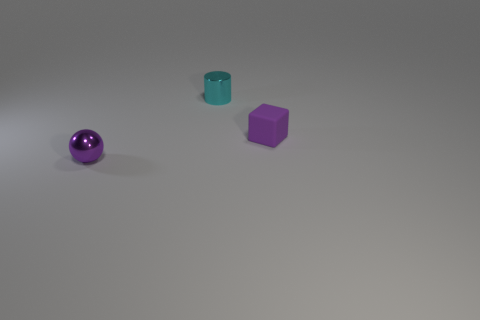}
    \caption{Case 2: original}
    \label{fig:clevr_case2_ori}
  \end{subfigure}
  \hfill
  \begin{subfigure}[t]{0.235\textwidth}
    \centering
    \includegraphics[width=\linewidth]{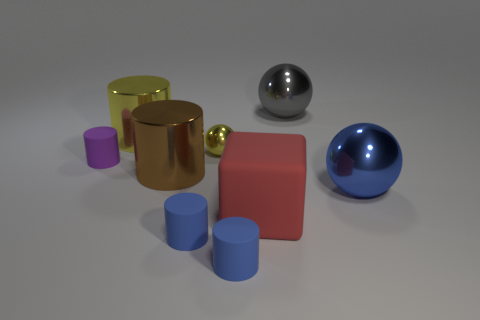}
    \caption{Case 3: original}
    \label{fig:clevr_case3_ori}
  \end{subfigure}
  \hfill
  \begin{subfigure}[t]{0.235\textwidth}
    \centering
    \includegraphics[width=\linewidth]{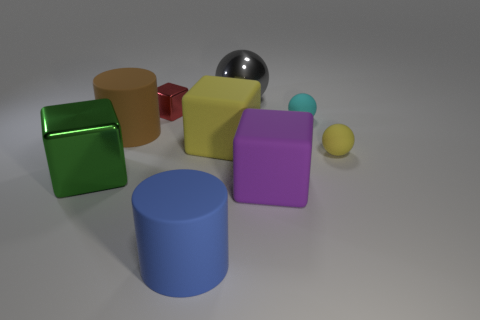}
    \caption{Case 4: original}
    \label{fig:clevr_case4_ori}
  \end{subfigure}

  \vspace{0.6em}

  \begin{subfigure}[t]{0.235\textwidth}
    \centering
    \includegraphics[width=\linewidth]{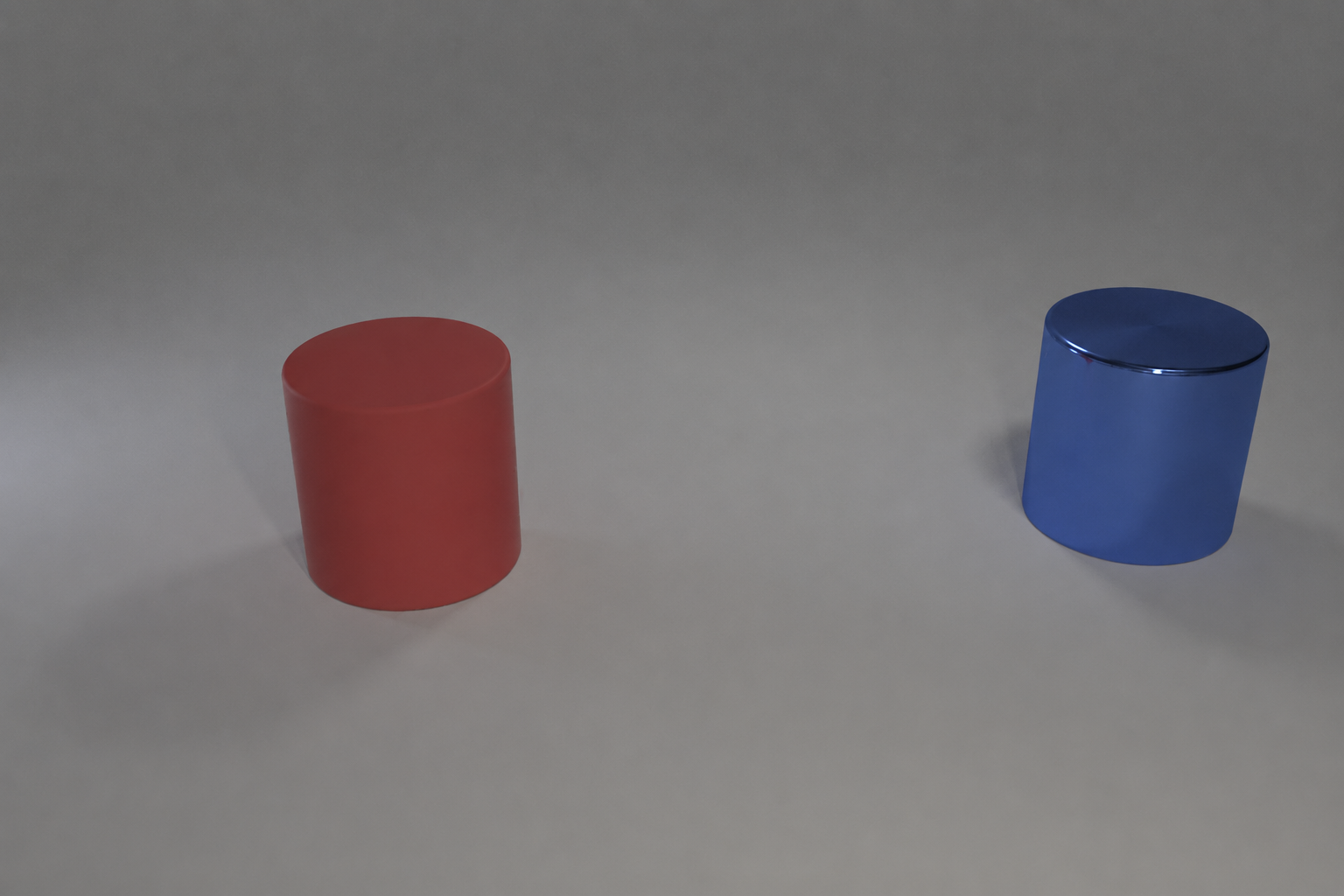}
    \caption{Case 1: easier}
    \label{fig:clevr_case1_easier}
  \end{subfigure}
  \hfill
  \begin{subfigure}[t]{0.235\textwidth}
    \centering
    \includegraphics[width=\linewidth]{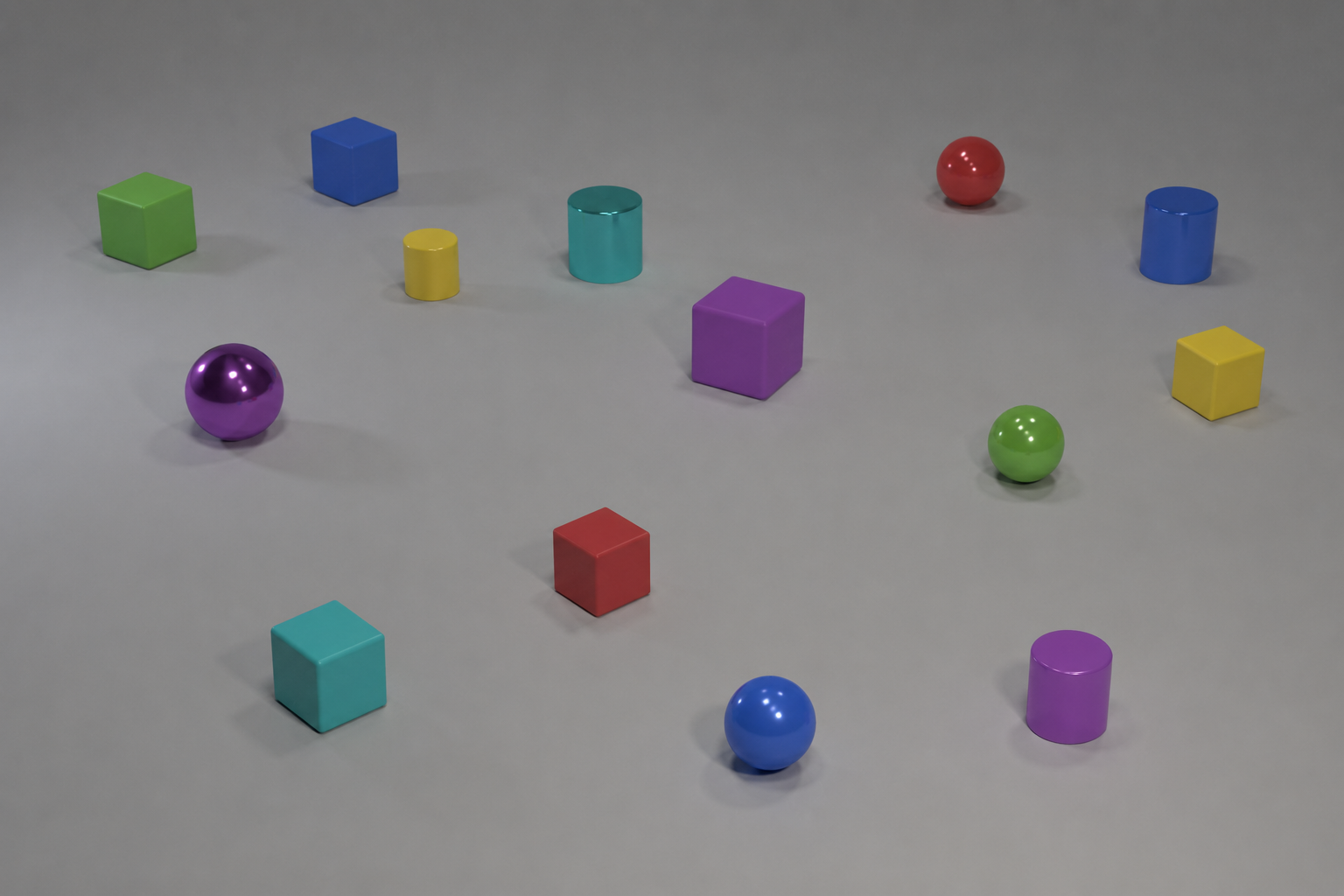}
    \caption{Case 2: harder}
    \label{fig:clevr_case2_harder}
  \end{subfigure}
  \hfill
  \begin{subfigure}[t]{0.235\textwidth}
    \centering
    \includegraphics[width=\linewidth]{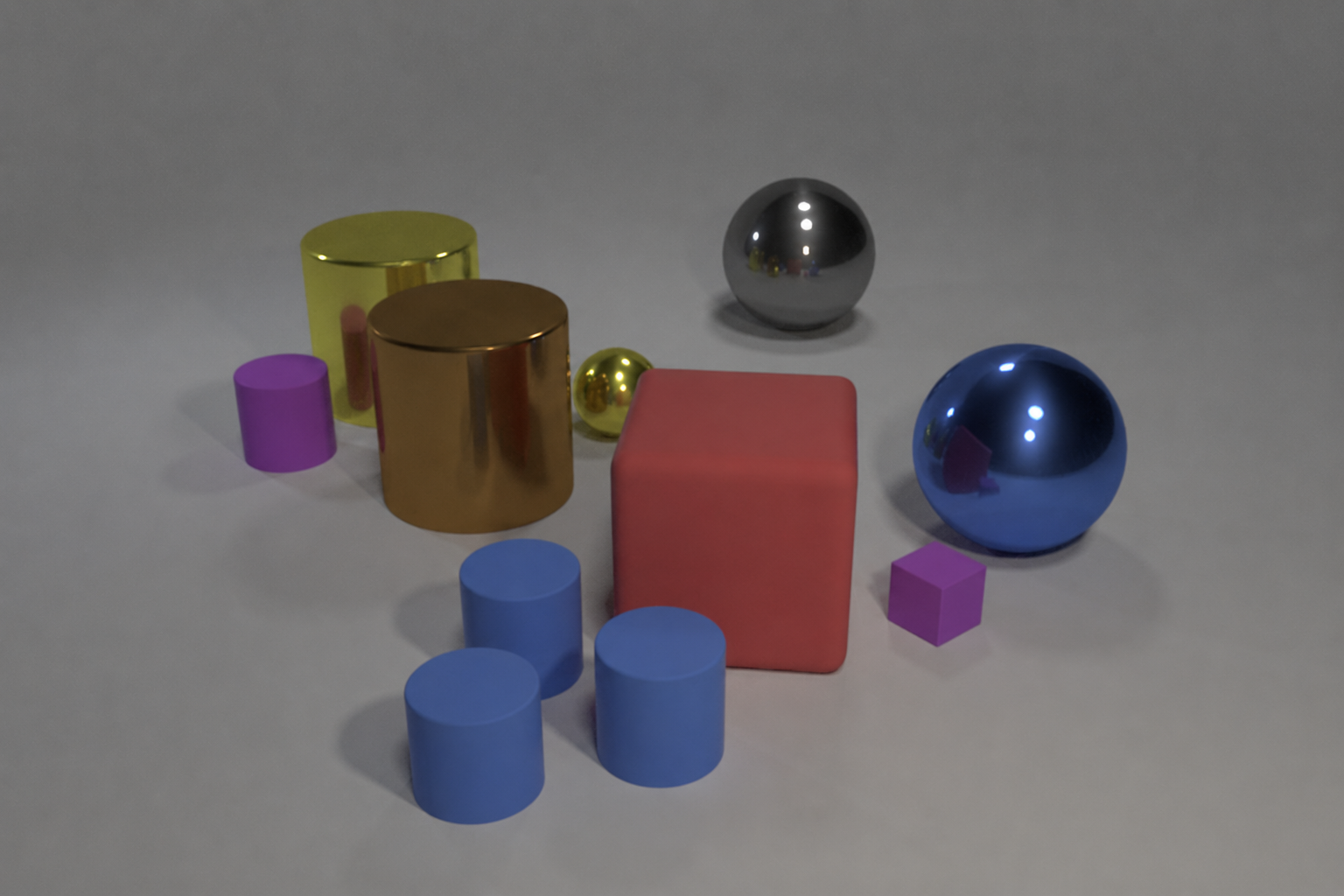}
    \caption{Case 3: harder}
    \label{fig:clevr_case3_harder}
  \end{subfigure}
  \hfill
  \begin{subfigure}[t]{0.235\textwidth}
    \centering
    \includegraphics[width=\linewidth]{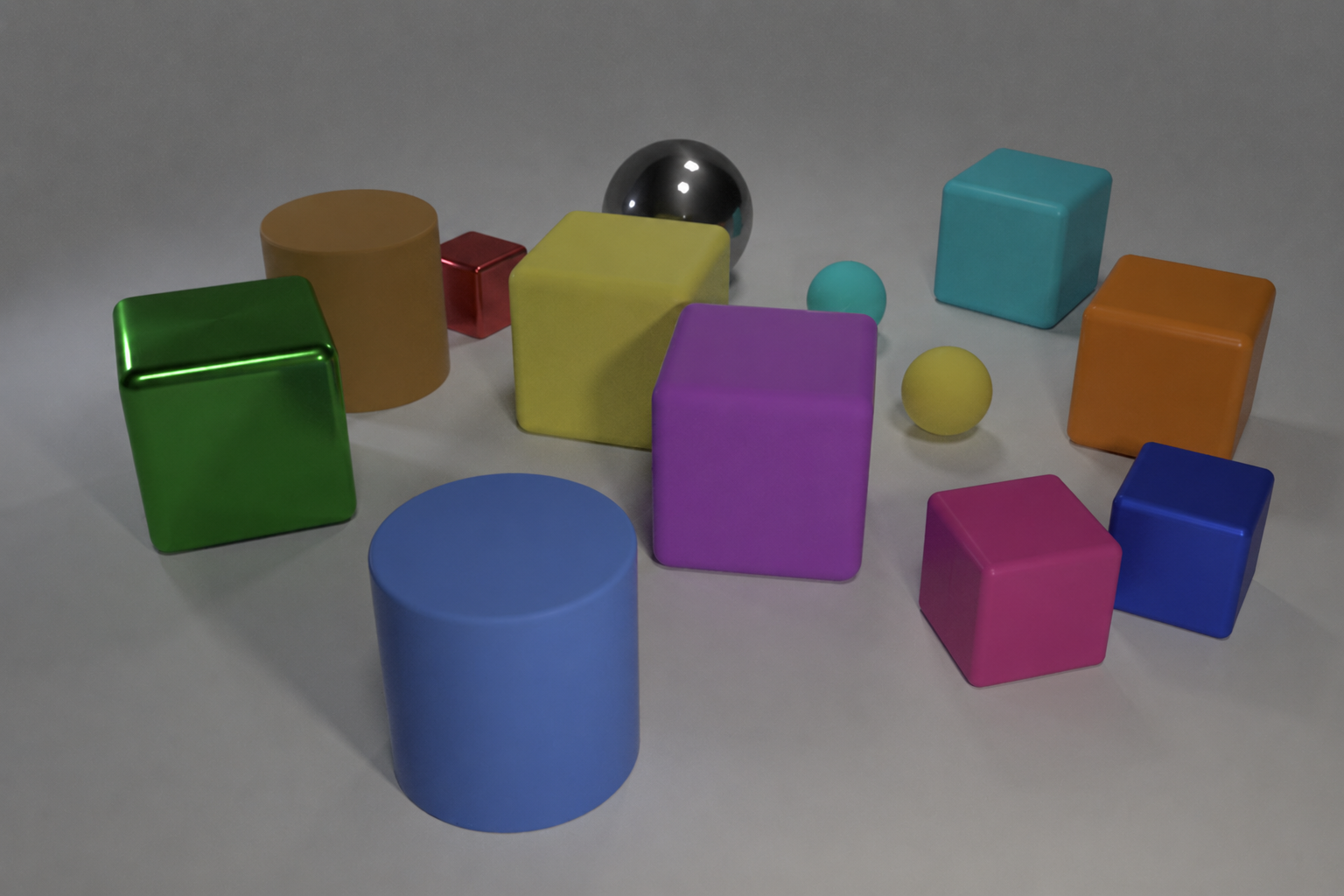}
    \caption{Case 4: harder}
    \label{fig:clevr_case4_harder}
  \end{subfigure}

  \caption{\textbf{CLEVR edits change difficulty while preserving verifiable labels.}
Each column shows one case before and after editing. The EnvRewriter modifies the scene graph, re-renders the image, and recomputes $a'$ by executing the original question program on the edited graph.}
  \label{fig:clevr_editing_examples}
\end{figure*}

Figure~\ref{fig:clevr_editing_examples} shows representative CLEVR edits.
Each example starts from a seed scene and applies a structured edit to object attributes, object categories, object counts, or spatial relations.
Because the edit is applied to the scene graph rather than directly to pixels, the modified scene remains renderable, and the answer can be recomputed by program execution.
These examples illustrate how \our{} produces label-valid visual perturbations instead of relying on model-based relabeling.

\subsection{ChartQA Table-Level Editing Examples}
\label{appendix:chartqa-case-study}
\begin{figure*}[t]
  \centering

  \begin{subfigure}[t]{0.31\textwidth}
    \centering
    \includegraphics[width=\linewidth]{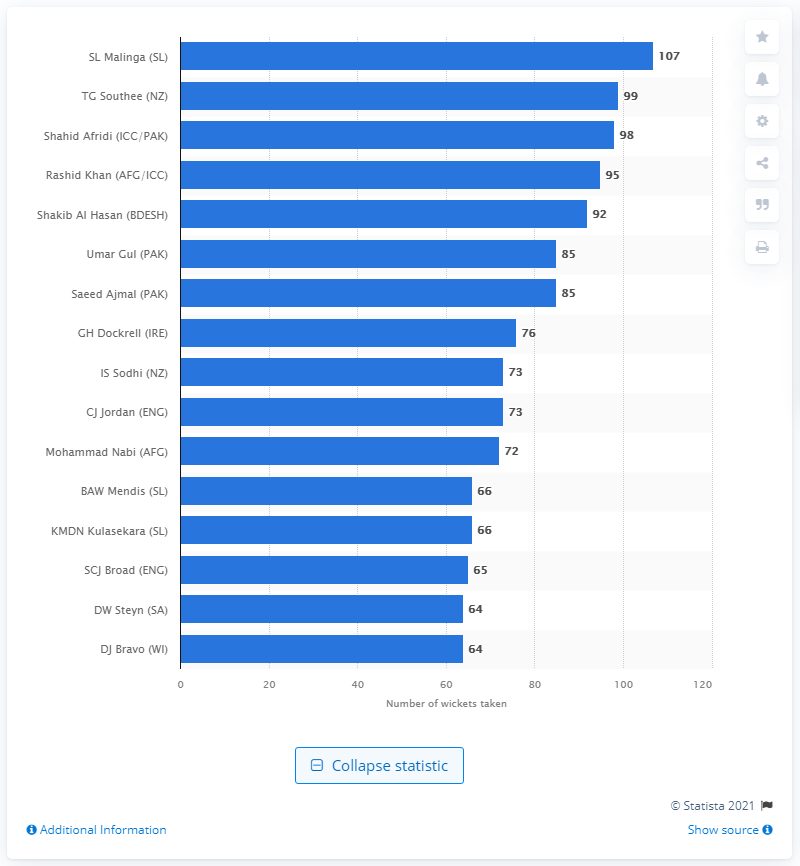}
    \caption{Case 1: original}
    \label{fig:zwz_case1_ori}
  \end{subfigure}
  \hfill
  \begin{subfigure}[t]{0.31\textwidth}
    \centering
    \includegraphics[width=\linewidth]{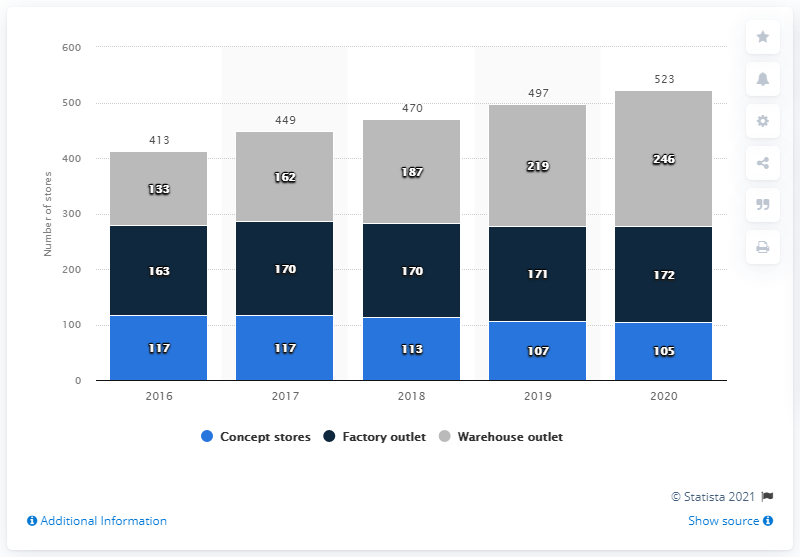}
    \caption{Case 2: original}
    \label{fig:zwz_case2_ori}
  \end{subfigure}
  \hfill
  \begin{subfigure}[t]{0.31\textwidth}
    \centering
    \includegraphics[width=\linewidth]{figures/chartqa-case-3.png}
    \caption{Case 3: original}
    \label{fig:zwz_case3_ori}
  \end{subfigure}

  \vspace{0.6em}

  \begin{subfigure}[t]{0.31\textwidth}
    \centering
    \includegraphics[width=\linewidth]{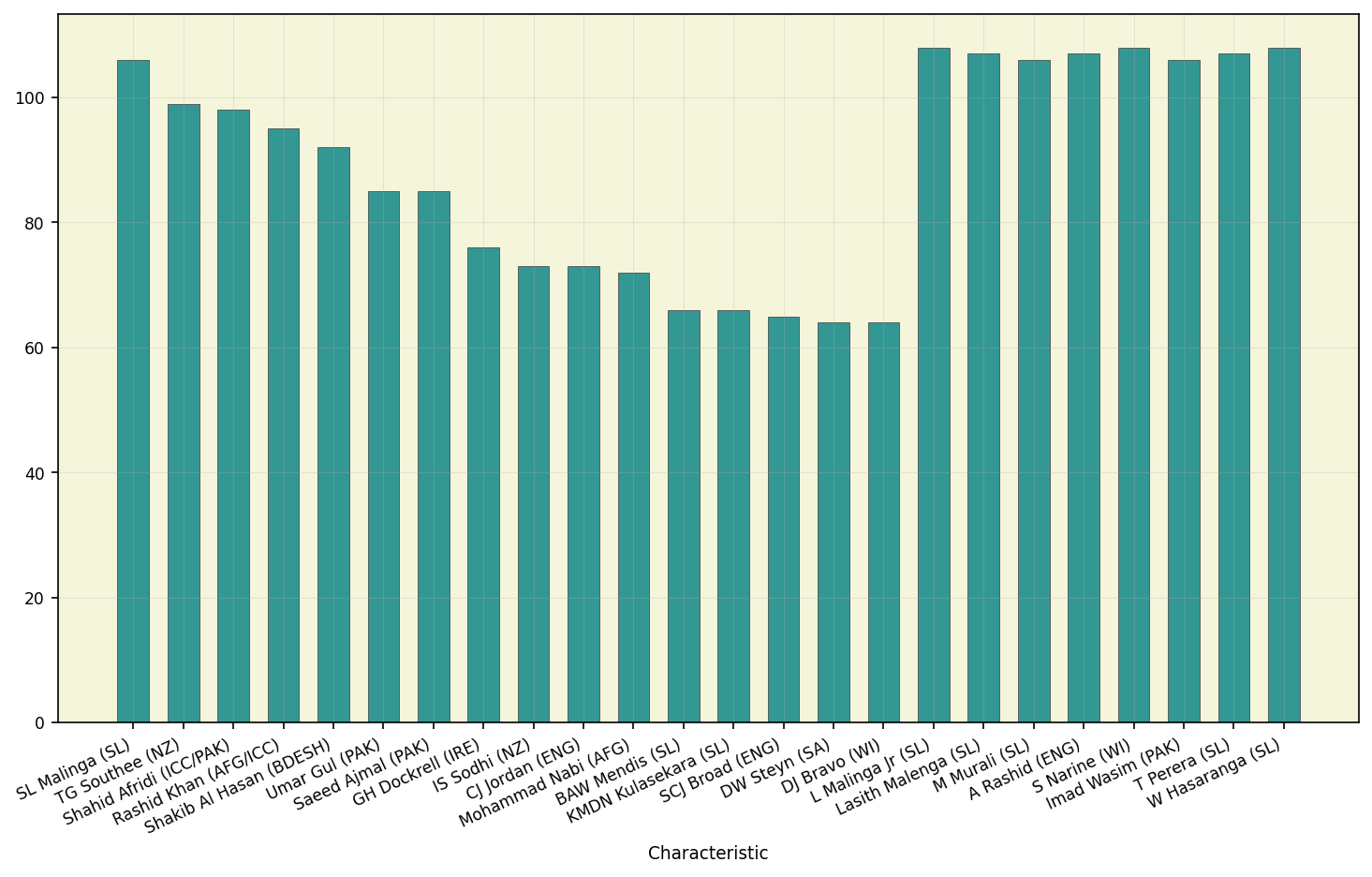}
    \caption{Case 1: edited}
    \label{fig:chartqa_case1_edit}
  \end{subfigure}
  \hfill
  \begin{subfigure}[t]{0.31\textwidth}
    \centering
    \includegraphics[width=\linewidth]{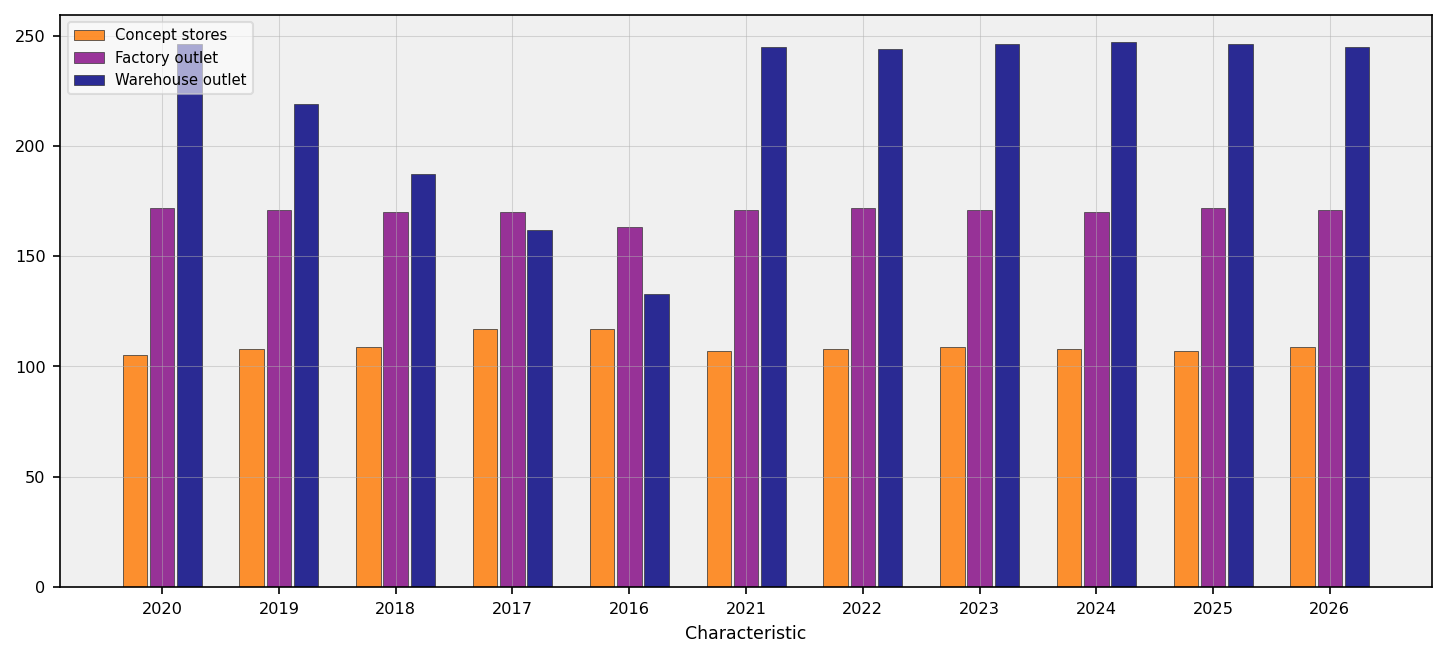}
    \caption{Case 2: edited}
    \label{fig:chartqa_case2_edit}
  \end{subfigure}
  \hfill
  \begin{subfigure}[t]{0.31\textwidth}
    \centering
    \includegraphics[width=\linewidth]{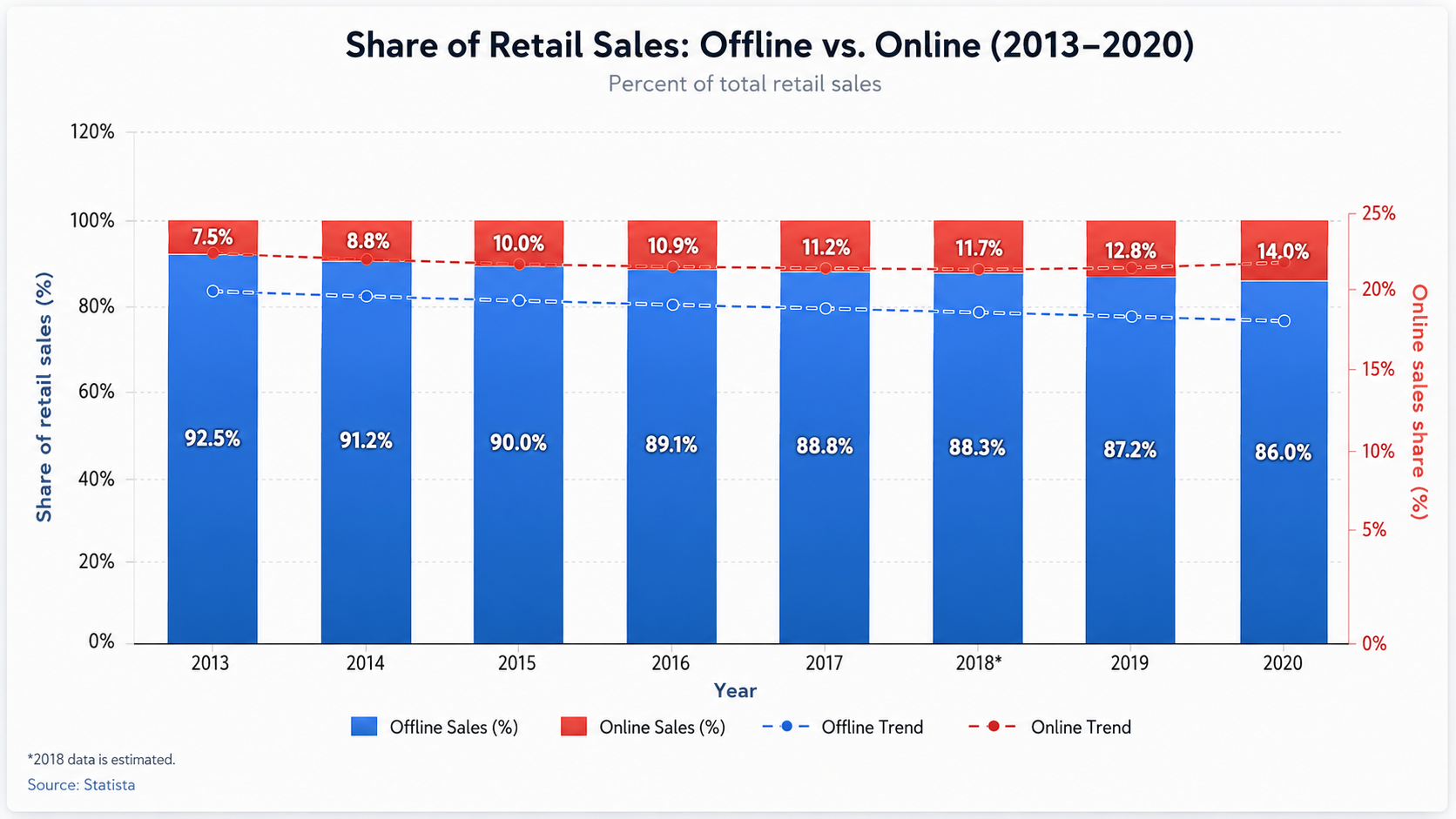}
    \caption{Case 3: edited}
    \label{fig:chartqa_case3_edit}
  \end{subfigure}

  \caption{\textbf{ChartQA edits increase difficulty through table-level changes.}
Each column shows one case before and after editing. The EnvRewriter modifies the CSV-level representation and re-renders the chart. For Tier~1 examples, $a'$ is recovered by entity-anchored lookup in the edited table; for Tier~2 examples, CSV values are kept fixed and $a'=a$ is preserved by construction.}
  \label{fig:chartqa_editing_examples}
\end{figure*}

We present the ChartQA edits shown in Figure~\ref{fig:chartqa_editing_examples}.
Each seed sample is associated with an underlying data table $\text{table}$, a question $q$, and an original answer $a$.
Conditioned on the current Actor pass rate $\hat{p}$, the EnvRewriter generates a table-level edit aimed at the desired difficulty.
To make the sample harder, it may add near-match rows, distractor series, or more confusable values; to make it easier, it may remove distractors or sharpen relevant contrasts.
The edit is applied to $\text{table}$ to produce $\text{table}'$, which is then re-rendered into a new chart.

We use the same two-tier verification rule as in Table~\ref{tab:verifiable_editing_protocol}.
For Tier~1 samples, the answer maps to a unique CSV cell
$(r^\star,j^\star)$. Let $e^\star$ denote the row entity at $r^\star$ in the
original table. After editing, the new answer is recovered by entity-anchored
lookup:
\[
a'=\mathrm{table}'[\mathrm{entity}=e^\star, j^\star].
\]
This avoids relying on the original row index after rows are inserted or
reordered.

For Tier~2 samples, the answer has no unique cell mapping, such as no match or
multiple matches. These cases often include derived quantities such as ratios,
differences, or percentages. For Tier~2, the EnvRewriter keeps CSV values fixed
and only varies rendering style or other answer-preserving visual factors, so
the label remains $a'=a$ by construction. In both tiers, no human relabeling or
LLM judgment is used.

\subsection{ZwZ Answer-Preserving Regional Editing Examples}
\label{appendix:zwz-case-study}

\begin{figure}[t]
  \centering
  \begin{tikzpicture}[
    img/.style={
      inner sep=0pt,
      draw=gray!35,
      rounded corners=2pt,
      line width=0.4pt,
      fill=white
    },
    title/.style={
      font=\scriptsize\bfseries,
      align=center
    },
    subtitle/.style={
      font=\tiny,
      align=center,
      text=gray!70!black
    }
  ]

  \def\imgw{3.55cm}
  \def\imgh{2.55cm}

  \node[img] (img1) {
    \includegraphics[width=\imgw,height=\imgh,keepaspectratio]{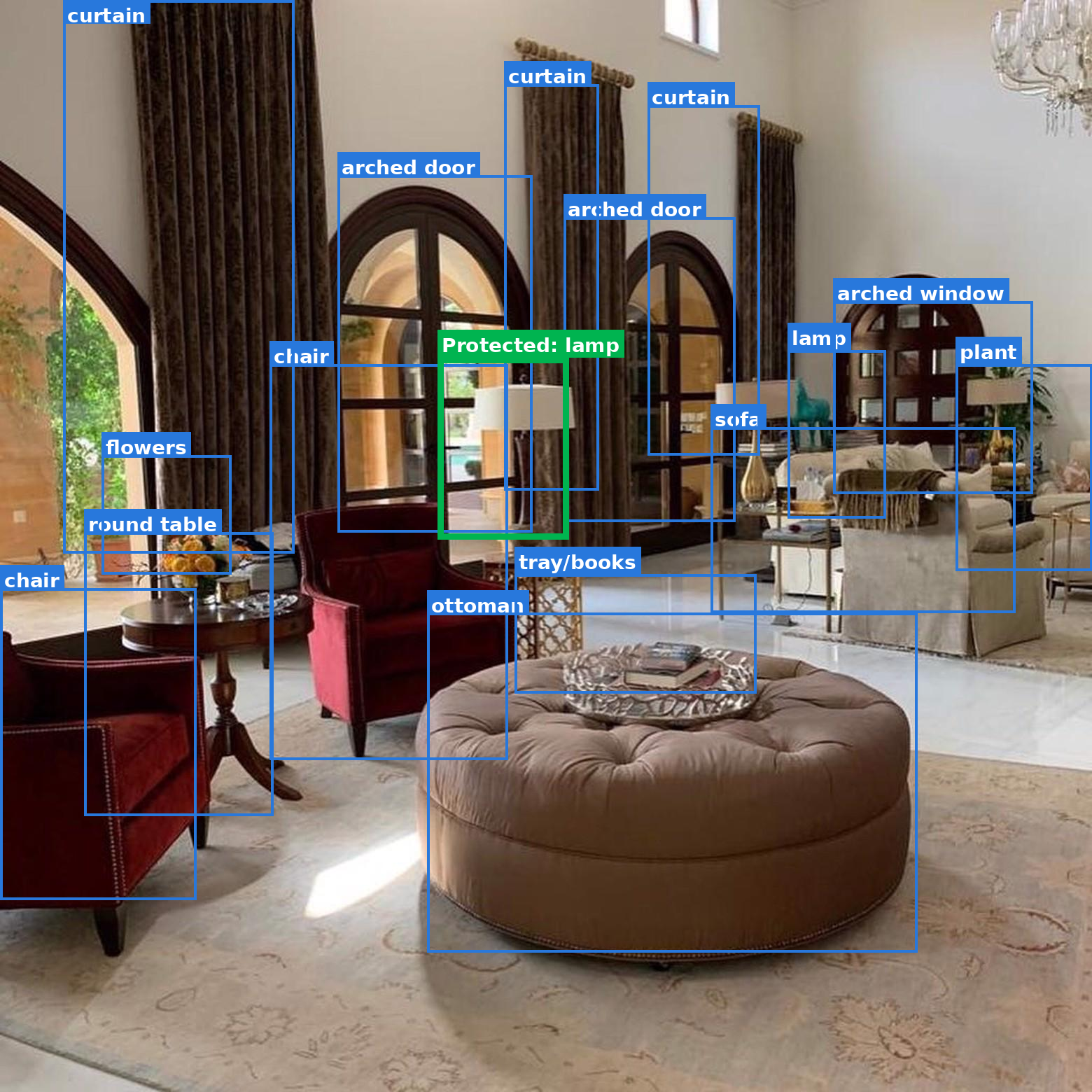}
  };
  \node[title, above=2pt of img1] {(a) protected target: lamp};
  \node[subtitle, below=3pt of img1, text width=\imgw] {green: answer-relevant \quad blue: other SAM3 regions};

  \node[img, right=0.55cm of img1] (img2) {
    \includegraphics[width=\imgw,height=\imgh,keepaspectratio]{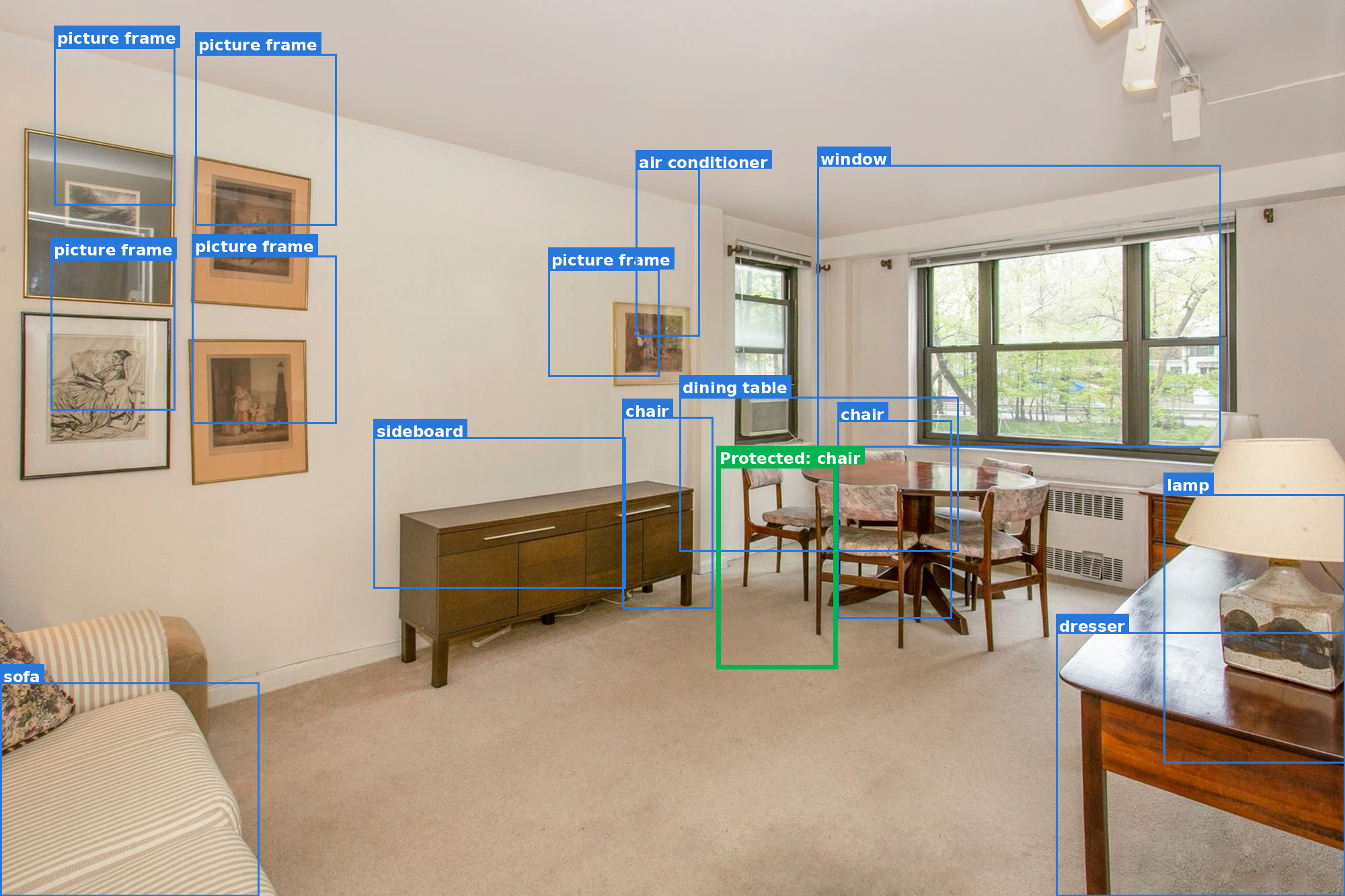}
  };
  \node[title, above=2pt of img2] {(b) protected target: chair};
  \node[subtitle, below=3pt of img2, text width=\imgw] {green: answer-relevant \quad blue: other SAM3 regions};

  \node[img, right=0.55cm of img2] (img3) {
    \includegraphics[width=\imgw,height=\imgh,keepaspectratio]{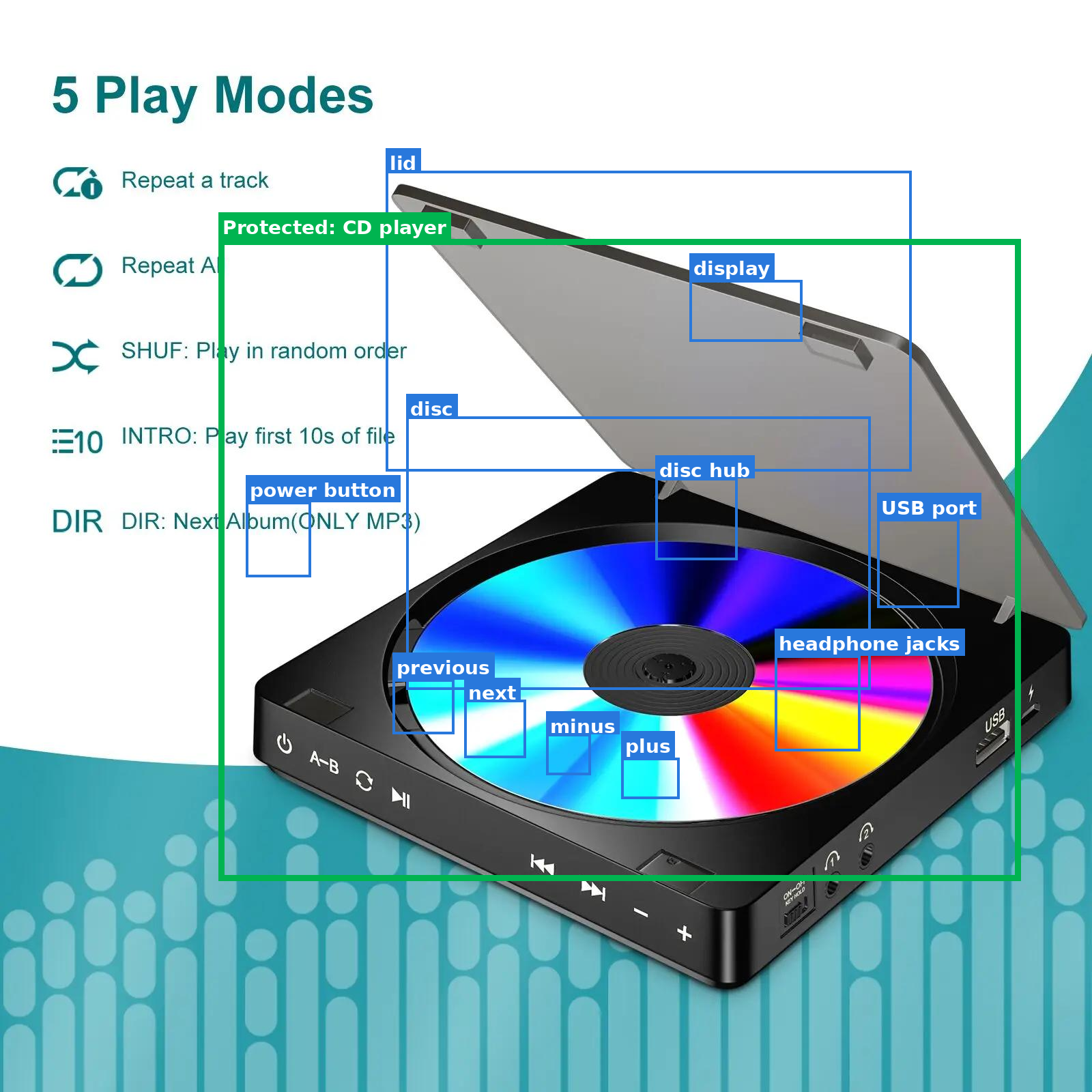}
  };
  \node[title, above=2pt of img3] {(c) protected target: device};
  \node[subtitle, below=3pt of img3, text width=\imgw] {green: answer-relevant \quad blue: other SAM3 regions};

  \end{tikzpicture}

  \caption{\textbf{ZwZ edits preserve answers by protecting target objects.}
  Green boxes mark answer-relevant targets, and blue boxes show other SAM3
  regions. \our{} keeps $(q'=q,a'=a)$ fixed and edits only non-target regions,
  allowing background difficulty to change while preserving the original answer.}
  \label{fig:zwz_editing_examples}
\end{figure}

Figure~\ref{fig:zwz_editing_examples} shows representative ZwZ examples.
Unlike CLEVR and ChartQA, ZwZ does not provide a complete symbolic state for deterministic relabeling.
Thus, \our{} keeps the original question-answer pair fixed, i.e., $q'=q$ and $a'=a$.
We use SAM3~\citep{carion2026sam3segmentconcepts} to identify object-level regions and mark the answer-relevant target objects as protected.
The EnvRewriter then edits only non-target regions, such as background areas or distractor objects, to adjust visual difficulty while preserving the evidence needed to answer the original question.

These examples highlight the difference between structured editing and
natural-image regional editing. For CLEVR, labels are recomputed by executing
the original question program on the edited scene graph. For ChartQA, labels are
handled by a two-tier rule: Tier~1 computes $a'$ by entity-anchored lookup in
the edited CSV table, while Tier~2 preserves $a'=a$ by keeping CSV values fixed
and only changing answer-preserving visual factors. For ZwZ, labels are
preserved by protecting answer-relevant regions and editing only non-target
visual context.

\subsubsection{What Happens without Answer Preservation?}
\label{appendix:zwz-without-answer-preservation}

\begin{figure*}[t]
  \centering
  \begin{tikzpicture}[
    img/.style={
      inner sep=0pt,
      draw=gray!35,
      rounded corners=2pt,
      line width=0.4pt,
      fill=white
    },
    title/.style={
      font=\scriptsize\bfseries,
      align=center
    },
    subtitle/.style={
      font=\tiny,
      align=center,
      text=gray!70!black
    },
    arrow/.style={
      -{Stealth[length=2mm]},
      thick,
      cyan!70!black
    },
    tag/.style={
      font=\tiny\bfseries,
      fill=cyan!12,
      draw=cyan!50!black,
      rounded corners=1pt,
      inner sep=1.5pt
    }
  ]

  \def\imgw{3.65cm}
  \def\imgh{2.65cm}

  \node[img] (easy) {
    \includegraphics[width=\imgw,height=\imgh,keepaspectratio]{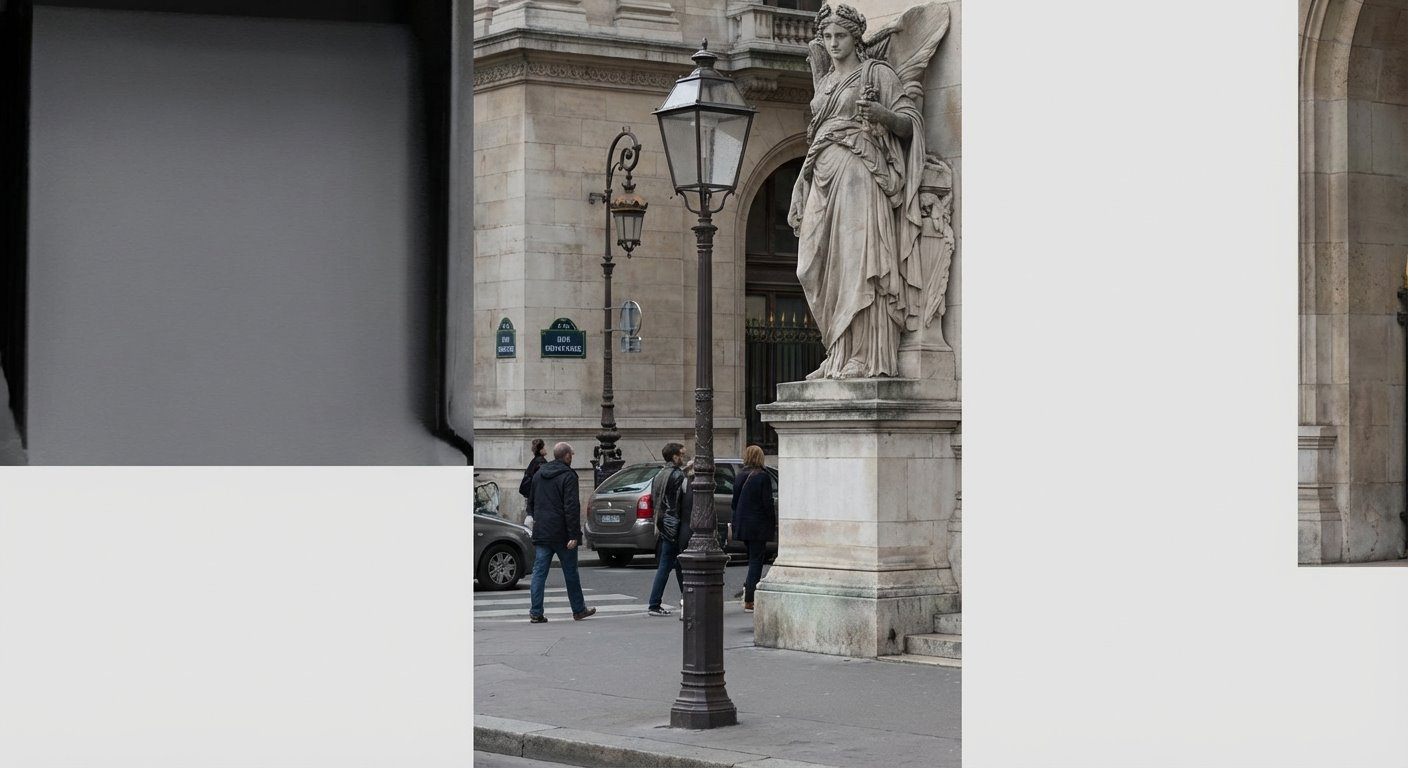}
  };
  \node[title, above=2pt of easy] {(a) easier};
  \node[subtitle, below=3pt of easy, text width=\imgw] {simplified scene};

  \node[img, right=1.25cm of easy] (orig) {
    \includegraphics[width=\imgw,height=\imgh,keepaspectratio]{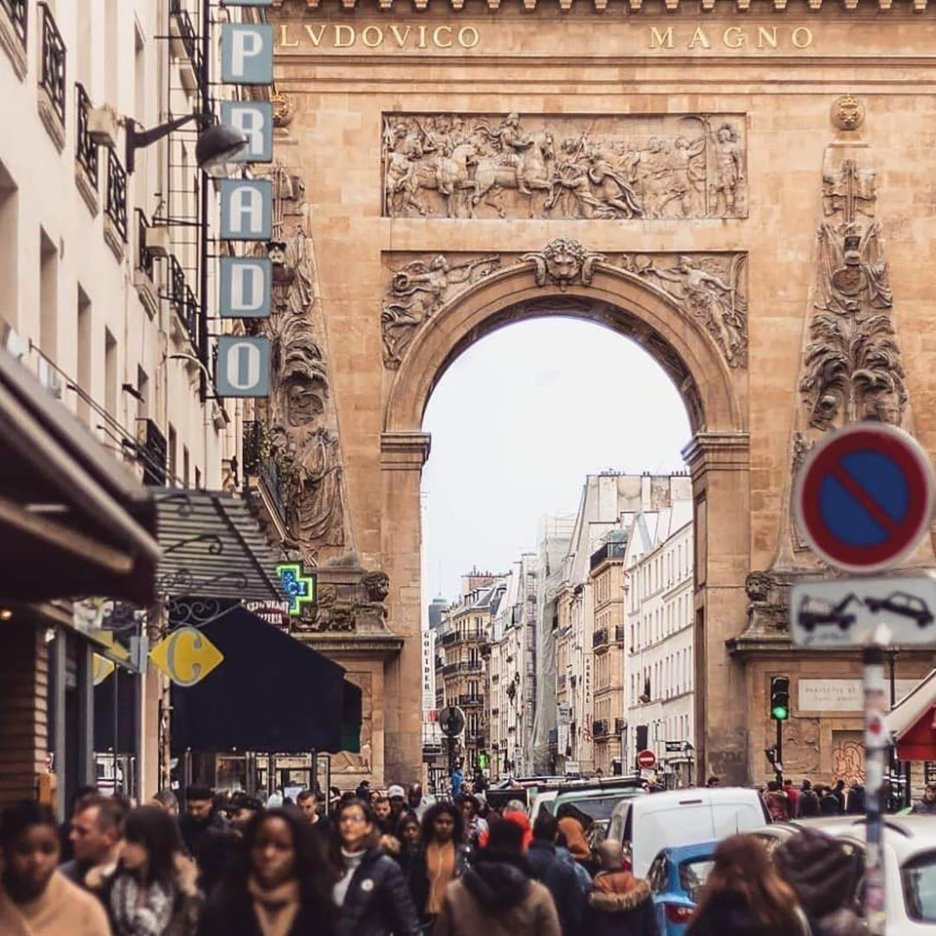}
  };
  \node[title, above=2pt of orig] {(b) original};
  \node[subtitle, below=3pt of orig, text width=\imgw] {source image};

  \node[img, right=1.25cm of orig] (hard) {
    \includegraphics[width=\imgw,height=\imgh,keepaspectratio]{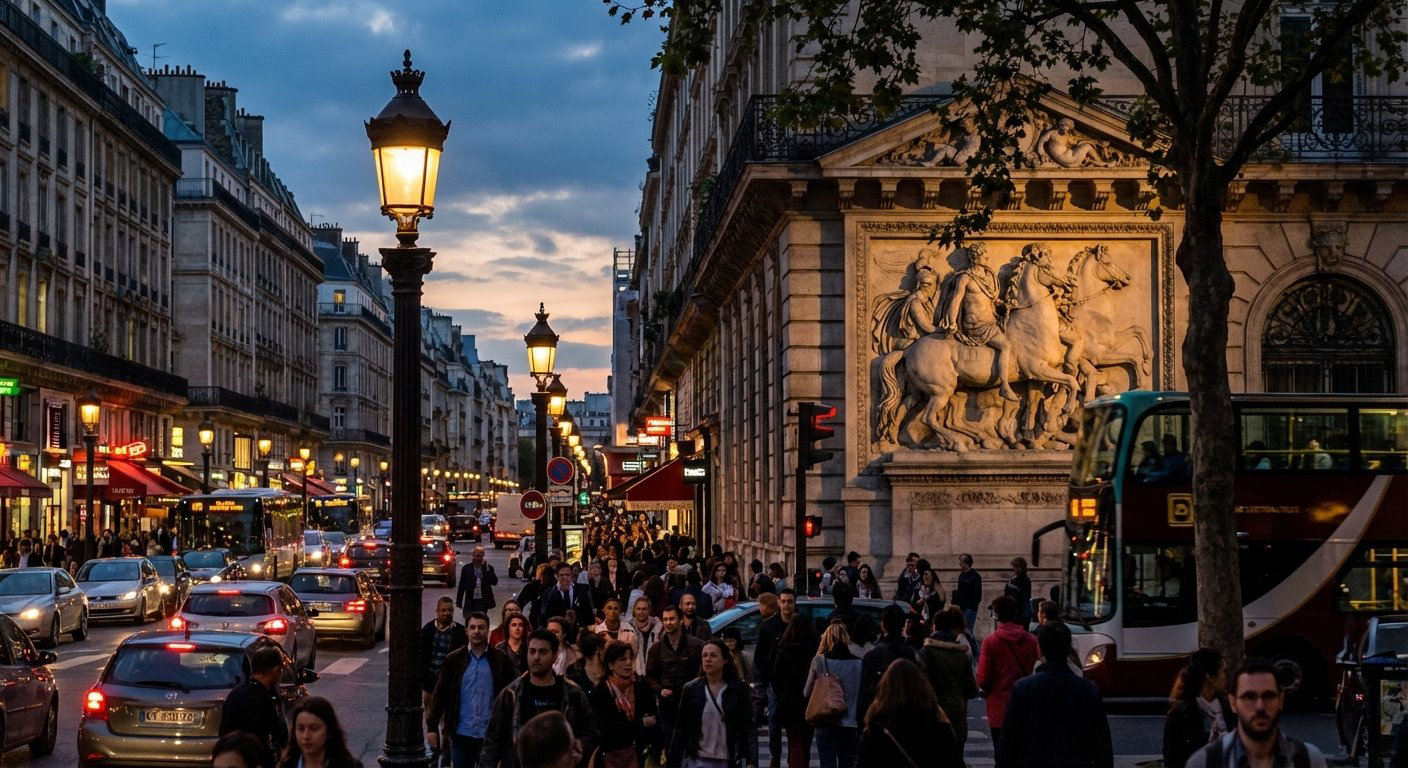}
  };
  \node[title, above=2pt of hard] {(c) harder};
  \node[subtitle, below=3pt of hard, text width=\imgw] {more visual ambiguity};

  \draw[arrow] (orig.west) -- node[above, tag] {make easier} (easy.east);
  \draw[arrow] (orig.east) -- node[above, tag] {make harder} (hard.west);

  \end{tikzpicture}

  \caption{\textbf{Direct Nano Banana editing can break answer validity.}
  Starting from the original image, direct generation can produce easier or harder variants, but the edits may change target relations or introduce ambiguous objects. As a result, the original question-answer pair may no longer be verifiable.}
  \label{fig:zwz_nanobanana_direct}
\end{figure*}

A natural alternative is to directly ask a generative image model to produce
an easier or harder image. However, this often breaks answer preservation.
Figure~\ref{fig:zwz_nanobanana_direct} shows one example. The original question
asks: ``Where is the street lamp in relation to the sculpture?'' The original
answer is ``left of.'' After direct generation with Nano Banana, the newly
generated answer becomes ``right of.''

This edit is problematic because the generated image no longer has a clearly
verifiable target relation. There are multiple street lamps and multiple
sculpture-like objects, so it is unclear which pair the answer refers to. The
image may look visually plausible, but the question-answer pair is no longer
label-valid. This illustrates why our editing pipeline explicitly enforces
answer preservation: the edited image should change visual difficulty while
keeping the answer-relevant objects and their relation verifiable.

\subsubsection{What Improves with \our{}}
\label{appendix:zwz_answer_with_preservation}
\begin{figure*}[t]
  \centering
  \begin{tikzpicture}[
    img/.style={
      inner sep=0pt,
      draw=gray!35,
      rounded corners=2pt,
      line width=0.4pt,
      fill=white
    },
    title/.style={
      font=\scriptsize\bfseries,
      align=center
    },
    subtitle/.style={
      font=\tiny,
      align=center,
      text=gray!70!black
    },
    arrow/.style={
      -{Stealth[length=2mm]},
      thick,
      cyan!70!black
    },
    tag/.style={
      font=\tiny\bfseries,
      fill=cyan!12,
      draw=cyan!50!black,
      rounded corners=1pt,
      inner sep=1.5pt
    }
  ]

  \def\imgw{3.25cm}
  \def\imgh{2.45cm}

  \node[img] (orig) {
    \includegraphics[width=\imgw,height=\imgh,keepaspectratio]{figures/ZWZ-case.PNG}
  };
  \node[title, above=2pt of orig] {(a) original};

  \node[img, right=0.78cm of orig] (allobj) {
    \includegraphics[width=\imgw,height=\imgh,keepaspectratio]{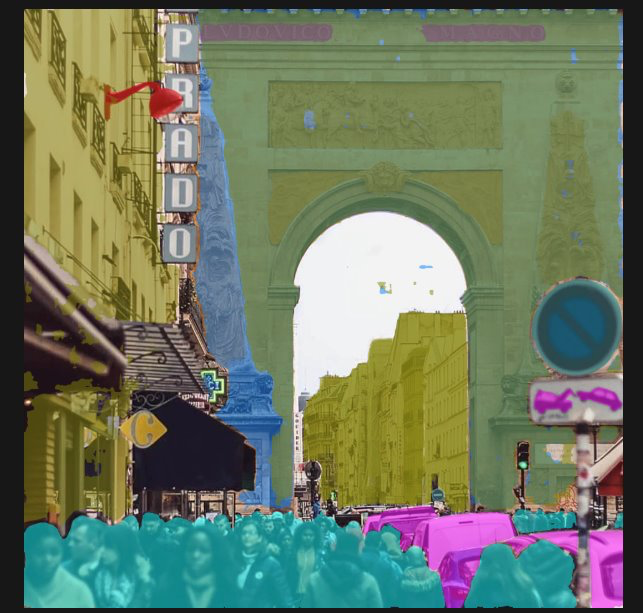}
  };
  \node[title, above=2pt of allobj] {(b) all objects};

  \node[img, right=0.78cm of allobj] (target) {
    \includegraphics[width=\imgw,height=\imgh,keepaspectratio]{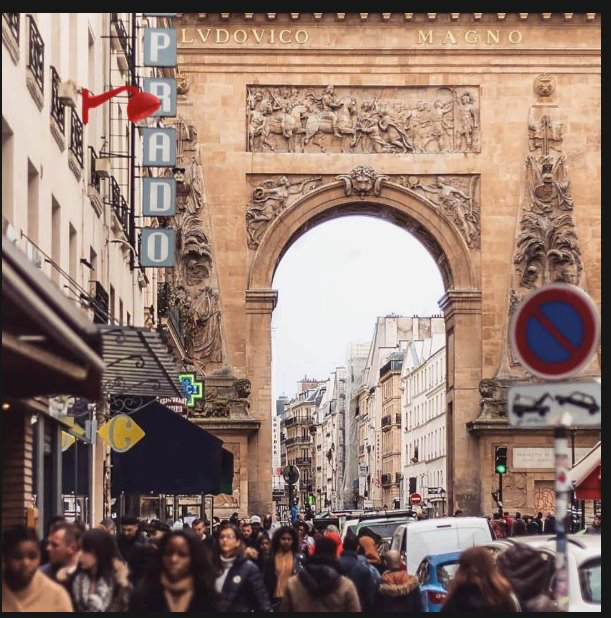}
  };
  \node[title, above=2pt of target] {(c) target objects};

  \node[img, right=0.78cm of target] (easy) {
    \includegraphics[width=\imgw,height=\imgh,keepaspectratio]{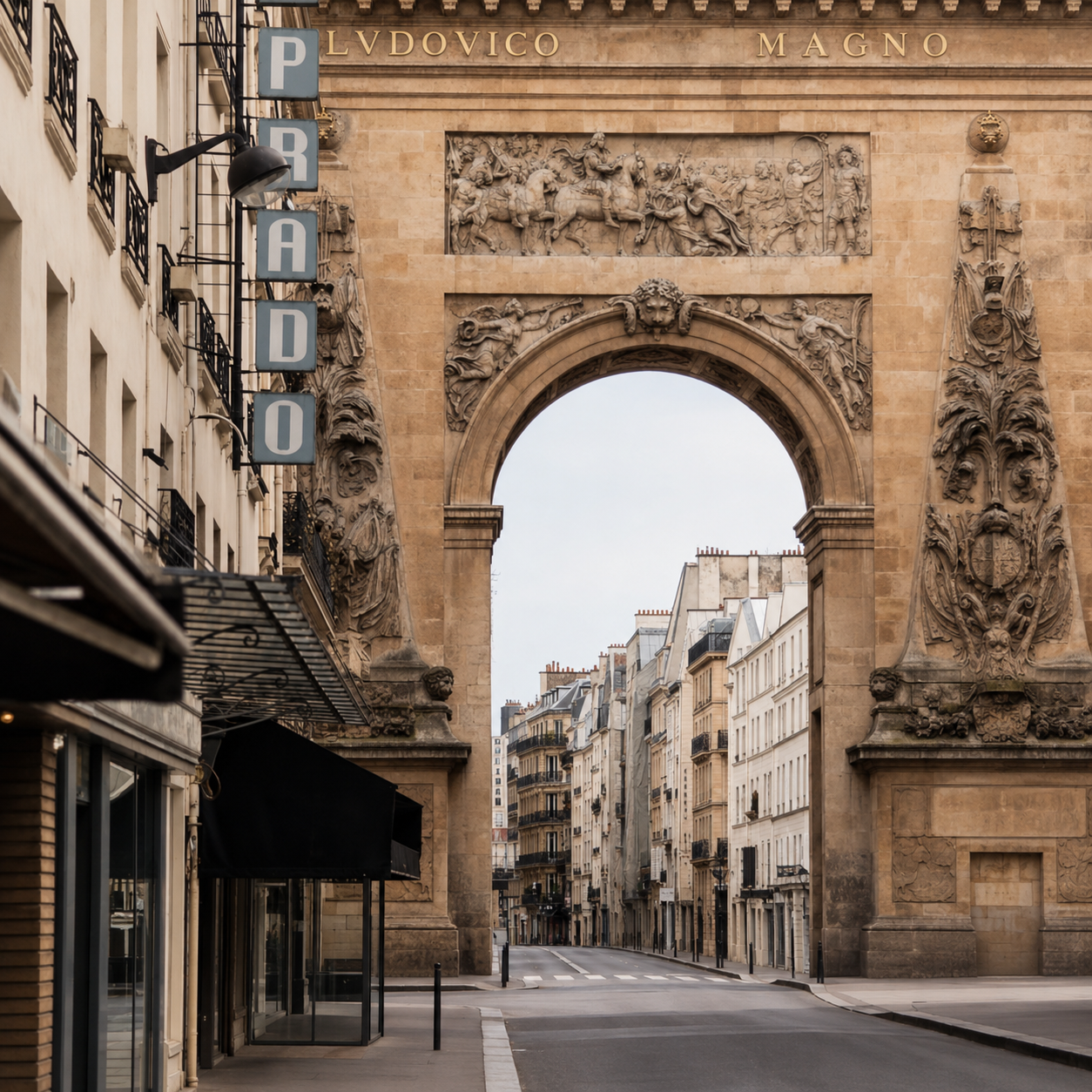}
  };
  \node[title, above=2pt of easy] {(d) easier with clean label};

  \draw[arrow] (orig.east) -- node[above, tag] {detect} (allobj.west);
  \draw[arrow] (allobj.east) -- node[above, tag] {lock} (target.west);
  \draw[arrow] (target.east) -- node[above, tag] {edit} (easy.west);

  \node[subtitle, below=3pt of orig, text width=\imgw] {input image};
  \node[subtitle, below=3pt of allobj, text width=\imgw] {SAM3 regions};
  \node[subtitle, below=3pt of target, text width=\imgw] {answer-relevant objects};
  \node[subtitle, below=3pt of easy, text width=\imgw] {non-target regions edited};

  \end{tikzpicture}

  \caption{\textbf{\our{} makes ZwZ easier while preserving the answer.}
  The pipeline detects object regions, locks the answer-relevant targets, and edits only non-target regions. This changes visual difficulty while keeping the original spatial relation and answer verifiable.}
  \label{fig:zwz_with-our-pipeline}
\end{figure*}

As shown in Figure~\ref{fig:zwz_with-our-pipeline}, this example illustrates
how \our{} edits a ZwZ image while preserving answer validity. Starting from the
original image in (d), \our{} first identifies all objects in the scene, as
shown in (c). It then selects and locks the answer-relevant target objects
mentioned in the question, shown in (b). In this example, the target objects
are the street lamp and the sculpture, whose relative relation determines the
answer.

After locking these target objects, \our{} edits only the non-target regions.
This allows the model to simplify the scene by removing distracting content,
such as the crowd and traffic signs, while keeping the answer-relevant objects
unchanged. The final edited image is shown in (a). The task becomes visually
easier, but the answer remains clean and verifiable because the key objects and
their spatial relations are preserved throughout the editing process.

\medskip
\section{Theoretical Analysis and Saturation}
\label{appendix:theory-saturation}
\label{appendix:theory-proofs}

This appendix presents the full derivation of the optimal difficulty target $\hat{p}^*_t$ that motivates the sweet-zone reward.

\subsection{Gradient-Signal Optimum}
\label{appendix:gradient-signal}
\label{appendix:theory-gradient-signal}

Under a REINFORCE-style policy gradient estimator~\citep{thomas2017policygradientmethodsreinforcement}, the expected gradient signal for a sample with Actor pass rate $\hat{p}$ is proportional to
\begin{equation}
  G(\hat{p}) = \hat{p}(1 - \hat{p}),
  \label{eq:gradient_signal}
\end{equation}
which is a concave quadratic maximized at $\hat{p}^* = 0.5$. Tasks that are too easy ($\hat{p} \to 1$) or too hard ($\hat{p} \to 0$) provide weak learning signals; half-correct tasks maximize gradient information. This motivates a single-point target $\hat{p}^* = 0.5$ in agent-environment co-evolution under standard assumptions.

\paragraph{Derivation of Eq.~\ref{eq:gradient_signal}.}
Let $y \sim \pi_\theta(\cdot \mid x)$ be a sampled solution, and let the verifier return a binary reward $R(y) \in \{0,1\}$. Define the actor pass rate as
\[
p = \Pr(R=1).
\]
Because $R$ only takes values $0$ and $1$, it follows a Bernoulli distribution:
\[
R \sim \mathrm{Bernoulli}(p), \qquad \mathbb{E}[R]=p.
\]
Using the REINFORCE estimator with the variance-minimizing baseline $b=\mathbb{E}[R]=p$~\citep{kaledin2022variancereductionpolicygradientmethods}, we have
\[
\nabla_\theta J(\theta)=\mathbb{E}\!\left[(R-p)\,\nabla_\theta \log \pi_\theta(y \mid x)\right].
\]
Since $R\sim\mathrm{Bernoulli}(p)$, we split the expectation into success and failure cases:
\[
\begin{aligned}
\nabla_\theta J(\theta)
&=\Pr(R{=}1)(1-p)\,\mathbb{E}\!\left[\nabla_\theta\log\pi_\theta(y\mid x)\mid R{=}1\right] \\
&\quad+\Pr(R{=}0)(0-p)\,\mathbb{E}\!\left[\nabla_\theta\log\pi_\theta(y\mid x)\mid R{=}0\right].
\end{aligned}
\]
Using $\Pr(R{=}1)=p$ and $\Pr(R{=}0)=1-p$, we obtain
\[
\begin{aligned}
\nabla_\theta J(\theta)
&=p(1-p)\,\mathbb{E}\!\left[\nabla_\theta\log\pi_\theta(y\mid x)\mid R{=}1\right] \\
&\quad+(1-p)(-p)\,\mathbb{E}\!\left[\nabla_\theta\log\pi_\theta(y\mid x)\mid R{=}0\right].
\end{aligned}
\]
Factoring out $p(1-p)$ gives
\[
\nabla_\theta J(\theta)=p(1-p)\left(\mathbb{E}\!\left[\nabla_\theta\log\pi_\theta(y\mid x)\mid R{=}1\right]-\mathbb{E}\!\left[\nabla_\theta\log\pi_\theta(y\mid x)\mid R{=}0\right]\right).
\]
Defining
\[
\Delta=\mathbb{E}\!\left[\nabla_\theta\log\pi_\theta(y\mid x)\mid R{=}1\right]-\mathbb{E}\!\left[\nabla_\theta\log\pi_\theta(y\mid x)\mid R{=}0\right],
\]
we conclude
\begin{equation}
  \nabla_\theta J(\theta)=p(1-p)\Delta.
  \label{eq:reinforce_pq}
\end{equation}
Under the trust-region KL constraint and gradient-clipping bias of GRPO-style updates~\citep{shao2024deepseekmathpushinglimitsmathematical}, the conditional log-policy gradients in $\Delta$ stay within a bounded range across $R\in\{0,1\}$, so $\|\Delta\|$ contributes only a bounded multiplicative constant. The strength of the learning signal is therefore proportional to $p(1-p)$, recovering Eq.~\ref{eq:gradient_signal}.

\subsection{Distribution Drift Cost}
\label{appendix:drift-cost}
\label{appendix:theory-drift}

The single-point analysis implicitly assumes that adjusting task difficulty incurs no cost. This holds for text-agent settings where generating a harder task does not change the task's distributional form. In visual reasoning, however, the EnvRewriter must edit images to shift $\hat{p}$ from its original value $\hat{p}_{\text{orig},t}$ toward the target. As the Actor strengthens and $\hat{p}_{\text{orig},t}$ rises, reaching $\hat{p}=0.5$ requires increasingly aggressive edits that push the training distribution away from the natural image distribution~\citep{borji2024noteshumailovetal,alemohammad2023selfconsuminggenerativemodelsmad}. The Actor may then learn patterns specific to these distorted images that do not transfer back to the real test set.

We model this trade-off by decomposing each round's expected accuracy gain into a learning-signal term and a distribution-drift penalty:
\begin{equation}
  \Delta\text{Acc}_t = C_t \cdot G(\hat{p}) - \lambda \cdot D_t(\hat{p}),
  \label{eq:drift_objective}
\end{equation}
where $C_t>0$ denotes the Actor's remaining capacity headroom at round $t$, $D_t(\hat{p})$ captures the generalization cost of shifting the training distribution, and $\lambda>0$ controls the Actor's sensitivity to distribution drift. The single-point analysis corresponds to the special case $D_t\equiv0$.

We adopt the simplest convex form for the drift cost:
\begin{equation}
  D_t(\hat{p})=\kappa_t\cdot(\hat{p}-\hat{p}_{\text{orig},t})^2,
  \label{eq:drift_cost}
\end{equation}
which satisfies three natural properties: zero cost when no editing occurs, monotonically increasing cost with larger deviations, and increasing marginal cost. More general convex forms $D_t=\kappa_t|\hat{p}-\hat{p}_{\text{orig},t}|^q$ with $q\geq1$ yield qualitatively identical conclusions.

\subsection{Optimal Difficulty Target}
\label{appendix:optimal-target}
\label{appendix:theory-optimum}

Substituting Eqs.~\ref{eq:gradient_signal} and~\ref{eq:drift_cost} into Eq.~\ref{eq:drift_objective} and taking the derivative with respect to $\hat{p}$,
\[
  \frac{d\,\Delta\text{Acc}_t}{d\hat{p}}=C_t(1-2\hat{p})-2\lambda\kappa_t(\hat{p}-\hat{p}_{\text{orig},t})=0.
\]
Solving yields
\begin{equation}
  \hat{p}^*_t=\frac{C_t+2\lambda\kappa_t\cdot\hat{p}_{\text{orig},t}}{2C_t+2\lambda\kappa_t}.
  \label{eq:optimal_raw}
\end{equation}
Defining $\rho_t=\frac{\lambda\kappa_t}{C_t+\lambda\kappa_t}\in[0,1]$, this rearranges to a convex combination:
\begin{equation}
  \boxed{\;\hat{p}^*_t=\tfrac{1}{2}+\rho_t\cdot\bigl(\hat{p}_{\text{orig},t}-\tfrac{1}{2}\bigr).\;}
  \label{eq:optimal_convex}
\end{equation}
The optimal target is a weighted interpolation between $0.5$ and the current $\hat{p}_{\text{orig},t}$, with the weight $\rho_t$ determined by the ratio of drift cost to remaining capacity.

\subsection{Two Limiting Regimes}
\label{appendix:theory-regimes}

Equation~\ref{eq:optimal_convex} unifies two extremes.

\paragraph{Early-stage ($\rho_t\to0$, i.e., $C_t\gg\lambda\kappa_t$).}
The Actor has ample capacity headroom, and light editing suffices to reach the target zone. The optimal target approaches $0.5$, recovering the gradient-signal optimum. In this regime, difficulty calibration should be aggressive.

\paragraph{Saturation regime ($\rho_t\to1$, i.e., $C_t\to0$).}
The Actor is near its capacity ceiling~\citep{yue2025doesreinforcementlearningreally}; reaching $\hat{p}=0.5$ would require extreme edits whose drift cost exceeds the learning benefit. The optimal target shifts toward $\hat{p}_{\text{orig},t}$, meaning the EnvRewriter should reduce its editing intensity. Continued training in this regime enters a drift-dominated phase where further editing degrades rather than improves performance.

\subsection{From Theory to the Sweet-Zone Reward}
\label{appendix:sweet-zone-reward}
\label{appendix:theory-sweetzone}

In practice, $C_t$ and $\kappa_t$ are not directly observable, making it infeasible to track $\hat{p}^*_t$ exactly. We instead define a sweet zone $[0.35,0.65]$ centered at $0.5$ as a robust approximation. The Gaussian reward $R_R(\hat{p})=\exp(-\beta(\hat{p}-0.5)^2)$ peaks at $0.5$ but assigns substantial weight to nearby pass rates, absorbing the drift of $\hat{p}^*_t$ across rounds without requiring direct estimation of $\rho_t$. The empirical saturation patterns in Appendix~\ref{appendix:saturation-evidence} provide direct evidence that \our{}'s training dynamics align with the regime predicted by Eq.~\ref{eq:optimal_convex}.

\subsection{Empirical Saturation Evidence}
\label{appendix:saturation-evidence}
\label{appendix:RQ9}
\label{sec:saturation}
\label{sec:saturation_rq}

\our{} keeps improving through R4 and reaches $89.12\%$ on ChartQA. At R5, accuracy slightly retreats to $88.94\%$. This is not a training failure but the saturation regime predicted by the drift-corrected analysis. Once the Actor approaches its capacity ceiling, the optimal target $\hat{p}^*_t$ drifts away from $0.5$, and forcing the original sweet zone begins to hurt rather than help. The analysis predicts four signatures of saturation, all of which are verified empirically.

\input{tables/tab-sweet-zone-decay}
\input{tables/tab-round-growth}

\paragraph{Signature 1: The optimal target drifts away from $0.5$.}
The EnvRewriter pulls $\hat{p}$ into the sweet zone in early rounds. But the target itself does not stay still. As the Actor strengthens, the optimal $\hat{p}^*_t$ shifts away from $0.5$ toward $\hat{p}_{\text{orig},t}$ because forcing a $50\%$ success rate would require edits whose drift cost outweighs their learning benefit. The EnvRewriter follows this shift: $\hat{p}$ moves from $0.502$ at R3 to $0.375$ at R5, tracing the predicted trajectory $\hat{p}^*_t\to\hat{p}_{\text{orig},t}$ as $\rho_t\to1$.

\paragraph{Signature 2: Useful training signals become scarce.}
The fraction of edits landing in $[0.35,0.65]$ decays monotonically from $53.4\%$ at R1 to $11.1\%$ at R5. Most edits no longer land in the trainable band, so each round delivers a weaker gradient signal even though the EnvRewriter still produces edits at the same rate.

\paragraph{Signature 3: Per-round gains decay smoothly.}
The full trajectory of round-over-round improvements is $+8.34$, $+2.26$, $+0.97$, $+0.65$, and $-0.18$. This traces the smooth deceleration expected as $C_t\to0$, rather than a sudden collapse.

\paragraph{Signature 4: Categories with the least headroom regress first.}
The main text identifies \texttt{value\_lookup} and \texttt{ratio\_pct} as two question types that gained most through R4. These large prior gains mean these categories have less remaining headroom, and at R5 they are the first to regress, while categories with smaller earlier gains remain stable. This selective regression is consistent with the target-drift mechanism in Eq.~\ref{eq:optimal_convex}.

Together, these four trajectories indicate that R5 marks entry into a drift-dominated regime. The fixed sweet zone $[0.35,0.65]$ that works through R4 is no longer aligned with the moving optimal target $\hat{p}^*_t$. This points to a concrete next step: rather than holding the difficulty band fixed, future work should schedule $\alpha$ adaptively or widen the trainable band as the Actor matures.

\subsection{Pass-Rate Distribution Shift}
\label{appendix:phat-distribution}

The empirical pass-rate distribution shifts across rounds. Early rounds concentrate near $\alpha=0.5$, while later rounds increasingly drift toward the extremes. This distributional shift explains why the sweet-zone hit rate decays and why additional rounds eventually yield diminishing returns.

\begin{figure}
  \centering
  \includegraphics[width=0.45\linewidth]{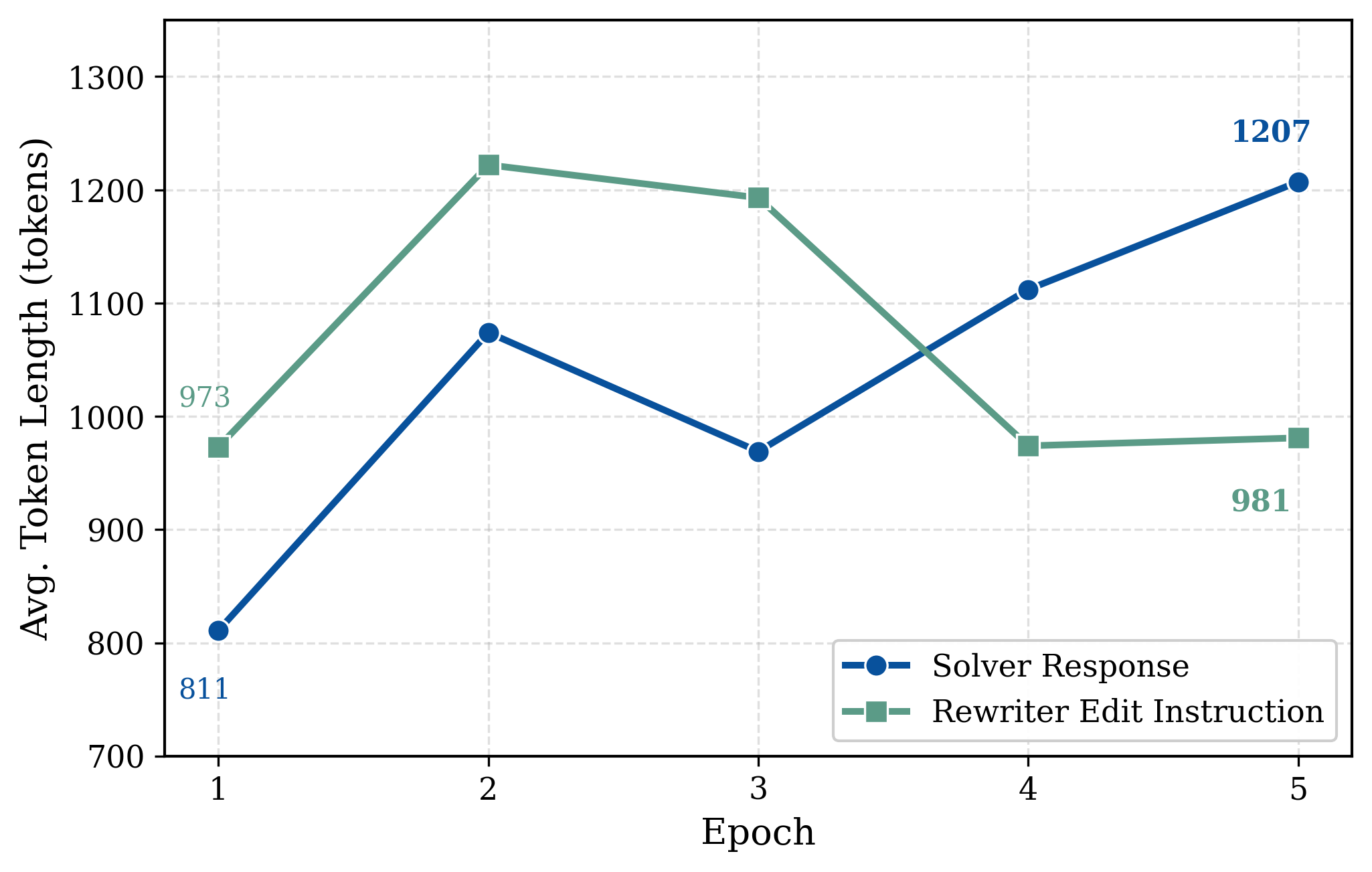}
  \vspace{-2.5pt}
  \caption{\textbf{EnvRewriter edit length stays stable.} Multi-round training improves edit strength without making edit instructions longer.}
  \label{fig:rewriter-edit-length}
\end{figure}

\section{Evaluation Protocols and Prompt Templates}
\label{appendix:evaluation-prompts}

\subsection{Evaluation Benchmarks and Metrics}
\label{appendix:evaluation-metrics}

We evaluate all models under the same zero-shot setting. For mathematical reasoning, we use MathVista, MathVision, MathVerse, WeMath, LogicVista, and ChartQA. For vision-centric understanding, we use MMMU, MMVP, and BLINK. Unless otherwise noted, the average in the main tables is computed as the equally weighted average of the mathematical-reasoning average and the vision-centric average.

\subsection{Answer Normalization and Scoring}
\label{appendix:answer-normalization}

For rule-based verification, the model response is first parsed into the required answer field. We normalize trivial formatting differences, including surrounding whitespace, capitalization for short textual answers, and common numerical formats. For ChartQA, we follow the official relaxed numerical accuracy protocol when reporting ChartQA-specific accuracy. For other benchmarks, we use each benchmark's standard answer matching or official evaluation script whenever available.

\subsection{Actor Rollout Prompt}
\label{appendix:actor-prompt}

The Actor is prompted to solve the visual question and return a response in the required reasoning-and-answer format. The key requirement is that the final answer can be parsed reliably by the verifier.

\begin{quote}\small
\textbf{Actor prompt template.}
Given the image and the question, reason step by step and provide the final answer in the required format. Put the final answer inside \texttt{<answer>} and \texttt{</answer>}. The answer should be concise and directly match the question.
\end{quote}

\subsection{EnvRewriter Prompt}
\label{appendix:rewriter-prompt}

The EnvRewriter is prompted to modify the visual environment while preserving verifiability. For structured domains, it must output an executable edit command over the scene graph or data table. For natural-image VQA, it must avoid changing the protected answer-relevant region.

\begin{quote}\small
\textbf{EnvRewriter prompt template.}
You are an environment rewriter. Given the original image metadata, question, answer, current Actor pass rate, and available edit tools, propose one valid edit that adjusts task difficulty toward the learnable zone. Return a structured command with the fields \texttt{tool}, \texttt{params}, and \texttt{reasoning}. The edit must remain executable and must preserve or programmatically update the answer.
\end{quote}

\subsection{Evaluation Prompt}
\label{appendix:evaluation-prompt}

During evaluation, we use a deterministic prompt format and do not expose the model to EnvRewriter metadata, pass-rate statistics, or training-time edit commands. The model only receives the test image and question.

\begin{quote}\small
\textbf{Evaluation prompt template.}
Answer the question based on the image. Provide the final answer in \texttt{<answer>} and \texttt{</answer>}. Do not include any information that is not supported by the image.
\end{quote}

\subsection{Baseline Implementation Details}
\label{appendix:baseline-details}

We compare against three categories of baselines: RLVR-trained reasoning models, self-evolution methods, and co-evolution methods that do not edit the visual environment. For published baselines, we use reported numbers when available under comparable settings. For in-house ablations, including Direct GRPO, Supervised GRPO, text-only editing, and static offline augmentation, we use the same Actor backbone and the same evaluation protocol as \our{} to isolate the effect of visual-environment co-evolution.

\subsection{Full ChartQA EnvRewriter Prompt}
\label{appendix:prompt_template}

\begin{lstlisting}[style=promptstyle, caption={rewriter editing planner prompt.}, label={lst:chart-planner-prompt}]
You are an expert chart editing planner.

Goal: choose one edit that moves the Actor pass rate ($\hat{p}$) closer to 0.5.

Tool:
- chart_rerender(edit_instruction): edit the chart's underlying data table and
  re-render it with matplotlib.

Rules:
- To make harder: add near-miss distractors, change the answer value, or increase
  chart density.
- To make easier: remove distractors, simplify the chart, or make the answer
  more distinct.
- Return one specific executable edit.

Output format:
<think>
your reasoning
</think>
<tool_call>
{"tool": "chart_rerender", "params": {"edit_instruction": "your edit"}}
</tool_call>
\end{lstlisting}

\section{Additional Main Results}
\label{appendix:main-results}
\label{appendix:main-results-supplementary}

\subsection{Full OOD Average Calculation}
\label{appendix:ood-average}

For the Chart-domain variant, we report the out-of-distribution average by excluding ChartQA, because ChartQA is the in-domain visual source for that variant. This avoids overstating generalization by including the benchmark most directly aligned with the training domain. For CLEVR and ZwZ variants, the same benchmark list is kept for consistency in the main comparison.

\subsection{Cross-Backbone Results}
\label{appendix:cross-backbone}
\label{appendix:backbone-models-results}

\our{} improves across different VLM backbones, suggesting that the co-evolution loop is not specific to a single model family. 

\subsection{Cost and Annotation Efficiency}
\label{appendix:cost-efficiency}
\label{sec:cost}

We provide a transparent accounting of \our{}'s computational and annotation costs to facilitate reproducibility and position the method against existing chart-specialized approaches.

\paragraph{Definition of label cost.}
We define ``labeled QA pairs'' as human-annotated or strong-LLM-generated question-answer pairs used as supervision targets during training. \our{} uses seed images from existing public collections but does not require additional human-written $(q,a)$ pairs during co-evolution: the EnvRewriter modifies a structured representation of each image, and the corresponding $(q',a')$ pair is computed programmatically from the modified representation. The seed images themselves are inputs to the loop and do not require new human labels.

\begingroup
\setlength{\abovecaptionskip}{3pt}
\setlength{\belowcaptionskip}{-4pt}
\input{figures/fig-cross-modality-per-round-acc}
\vspace{-0.6em}
\endgroup

\paragraph{Compute budget.}
We report the measured cost for the
\our{}-Qwen2.5-VL-7B chart-backbone setting, which requires approximately
$149$ A100 GPU hours. For the main Qwen3VL-8B setting, the full \our{}
pipeline runs through R5 for saturation analysis. The actor-training stage
through R4 requires approximately $305$ A100 GPU hours: $\sim101$ for
Round 1, $\sim85$ for Round 2, $\sim27$ for Round 3, and $\sim91$ for
Round 4, which trains for more steps. The R5 run is used to verify saturation.
Trajectory generation contributes an additional $\sim40$ A100 GPU hours and
is parallelizable across rounds.

\paragraph{No additional human annotation during training.}
Each edited training pair is derived from the EnvRewriter's modified structured
representation, eliminating the need for human-labeled $(q,a)$ pairs at any
stage of the co-evolution process.

\input{tables/tab-cost-comparison}

\subsection{Comparison with Chart-Specialized Methods}
\label{appendix:chart-specialized}
\label{appendix:chartqa-specialized-methods-comparison}

Table~\ref{tab:chart_specialized_comparison} compares \our{} against three recent chart-specialized methods: Chart-R1~\citep{chen2026chartr1chainofthoughtsupervisionreinforcement}, Bespoke-MiniChart-7B~\citep{movva2025enhancingscientificvisualquestion}, and ECD~\citep{yang2025effectivetrainingdatasynthesis}. \our{}-Qwen3VL-8B reaches $89.12\%$ on ChartQA using seed visual inputs and no additional labeled QA pairs, while remaining competitive with methods that use tens or hundreds of thousands of labeled pairs.

\begin{table}[t]
\caption{\textbf{\our{} reaches competitive ChartQA accuracy with far fewer labeled QA pairs.} All methods use ChartQA's official relaxed accuracy with $5\%$ numerical tolerance. \our{} uses seed inputs and avoids extra labeled QA supervision, while chart-specialized baselines require tens or hundreds of thousands of labeled pairs.}
\label{tab:chart_specialized_comparison}

\centering
\footnotesize
\begin{tabular}{lcc}
\toprule
Method & Labeled QA & ChartQA (\%) \\
\midrule
Chart-R1-7B~\citep{chen2026chartr1chainofthoughtsupervisionreinforcement} & $\sim$33k & $91.04$ \\
Bespoke-MiniChart-7B~\citep{movva2025enhancingscientificvisualquestion} & $\sim$91k & $89.5$ \\
ECD (Qwen2.5-VL-7B)~\citep{yang2025effectivetrainingdatasynthesis} & $\sim$320k & $85.32$ \\
\midrule
\textbf{\our{}-Qwen2.5-VL-7B} (Ours) & $2{,}000$ seed inputs & $87.96$ \\
\textbf{\our{}-Qwen3-VL-8B} (Ours) & $2{,}000$ seed inputs & \textbf{$89.12$} \\
\bottomrule
\end{tabular}
\end{table}

\begin{enumerate}[leftmargin=1.6em,itemsep=0.1em,topsep=0.2em]
    \item \emph{Strong cost-efficiency}: \our{} avoids additional labeled QA supervision and uses substantially fewer seed inputs than chart-specialized SFT/RL approaches.
    \item \emph{Practical paradigm}: \our{} scales to new domains without proportional annotation effort, making it attractive in domains where labeled QA is scarce.
\end{enumerate}

\subsection{Additional Per-Benchmark Results}
\label{appendix:additional-results}

Extended per-benchmark details are reported in the domain-specific tables and analyzes throughout this appendix.

\section{Ablation and Robustness Studies}
\label{appendix:ablation-robustness}

\subsection{EnvRewriter SFT as a Co-evolution Catalyst}
\label{appendix:rewriter-sft}

We examine the relationship between supervised initialization and co-evolution using four experimental settings (Table~\ref{tab:rewriter_init}). The EnvRewriter is initialized on verified GPT-5 edit demonstrations, which supervise both executable command generation and difficulty-adjustment strategies. These comparisons assess the benefit of combining this initialization with co-evolution.

\textbf{Direct Actor SFT baseline.} Applying supervised fine-tuning directly to the Actor, without co-evolution, improves ChartQA from $76.9\%$ to $79.4\%$, a gain of $2.5$ percentage points. This comparison measures the effect of direct Actor SFT in the evaluated setting; it does not isolate the contribution of the GPT-5 demonstrations used to initialize the EnvRewriter.

\textbf{Co-evolution without EnvRewriter SFT.} Running the co-evolution loop with a vanilla, non-SFT EnvRewriter reaches $78.6\%$ ($1.7$ percentage points above the base model). On manual inspection, the vanilla EnvRewriter frequently produces malformed or unexecutable edit commands, limiting the supply of valid training samples. This result highlights the importance of initializing an EnvRewriter that can produce executable edits.

\textbf{Combining EnvRewriter SFT and co-evolution.} Initializing the EnvRewriter with verified GPT-5 edit demonstrations and then updating it through co-evolution yields $89.1\%$ on ChartQA ($12.2$ percentage points above the base model). The combined setting outperforms both direct Actor SFT and co-evolution with a non-SFT EnvRewriter, supporting the complementary roles of supervised initialization and iterative actor--environment adaptation.

These results support combining EnvRewriter SFT with co-evolution. Because the initialization demonstrations supervise both command format and editing strategy, the present comparisons do not separately quantify these two contributions or isolate the contribution of GPT-5-generated supervision.

\subsection{Static Offline Augmentation Details}
\label{appendix:offline-augmentation}

To test whether the gain comes simply from stronger generated data, we compare \our{} against static offline augmentation without the iterative actor–environment loop. The offline variants generate a fixed corpus with self-generation, GPT-5, or Nano Banana, and then train the Actor on that fixed data. These variants improve over the base model but remain below iterative visual co-evolution, showing that the dynamic feedback loop contributes more than one-shot data quality.

\subsection{Text-Only Editing vs. Visual Editing}
\label{appendix:text-vs-image}

To isolate the role of the editing modality, we compare text-only perturbation with visual editing under the same CLEVR training source. Text-only perturbation changes the question or answer format while keeping the visual scene fixed. Visual editing modifies the image itself through the verifiable editing pipeline. The main results show that visual editing provides larger gains, especially on chart reasoning, supporting the claim that effective multimodal co-evolution requires adapting the visual environment rather than only rewriting text.

\subsection{Robustness to Imperfect Image Editing}
\label{appendix:editor-robustness}
\label{appendix:20-noise}

In practice, image editors may produce failures such as blank renders or mismatches between images and labels. To simulate partial editing failures, we corrupt $20\%$ of the R1 training data: $10\%$ of the samples are replaced with blank white images, and another $10\%$ use unedited original images paired with edited labels. As shown in Table~\ref{tab:20-noise}, the overall average performance drops only marginally from $66.8\%$ to $66.0\%$, less than one point, and the noisy model still outperforms the base model by $+2.4\%$. This shows that \our{} does not rely on a perfect image editor: failed edits do not fully destroy the training signal, and the remaining valid samples still support stable co-evolution.

\subsection{Stability Across Random Seeds}
\label{appendix:seed-stability}
\label{app:multi-seed-stability}

A natural concern for any RL-based training pipeline is whether the reported gains depend on a favorable random seed. We re-ran the first round of training with three independent random seeds on CLEVR and evaluated the resulting \our{}-8B (CLEVR) checkpoints on ChartQA and MathVista.

\begin{table}[h]
\caption{%
\textbf{\our{} is stable across random seeds.}
We re-run the first CLEVR training round with three seeds and evaluate on ChartQA and MathVista. Standard deviation is far below the reported gains, so improvements are not driven by favorable seed selection.
}
\label{tab:seed_robustness}

\centering
\footnotesize
\setlength{\tabcolsep}{12pt}
\renewcommand{\arraystretch}{1.2}
\begin{tabular}{@{}lcc@{}}
\toprule
\textbf{Seed} & \textbf{ChartQA (\%)} & \textbf{MathVista (\%)} \\
\midrule
Seed 1 (reported) & 85.2 & 73.8 \\
Seed 2 & 84.5 & 72.9 \\
Seed 3 & 85.2 & 71.7 \\
\midrule
Mean $\pm$ Std & 85.0 $\pm$ 0.4 & 72.8 $\pm$ 1.1 \\
\midrule
Base (Qwen3VL-8B) & 76.9 & 67.7 \\
Gain over base & $+8.1$ & $+5.1$ \\
\bottomrule
\end{tabular}
\end{table}

We focus this analysis on Round~1 because in \our{}, the seed data distribution is fixed once the first round is initialized; subsequent rounds operate on EnvRewriter-generated samples that no longer depend on the initial seed. As shown in Table~\ref{tab:seed_robustness}, the standard deviation of final accuracy is $0.4$ points on ChartQA and $1.1$ points on MathVista, well within the margin of our reported improvements. We therefore report single-seed results in the main text, following the convention adopted by prior self-evolution and co-evolution work.

\subsection{Robustness to Initial Data Difficulty}
\label{appendix:initial-difficulty}

A second source of variability is the initial task difficulty. Fixed-distribution
methods can be sensitive to this choice: easy seeds leave little room for improvement,
while hard seeds can stall training.

\our{} is less sensitive by design. The EnvRewriter is rewarded for producing
edits with a pass rate near $\alpha=0.5$. It makes easy samples harder and hard
samples easier, pushing both toward the sweet zone. As shown in
Appendix~\ref{appendix:bidirectional-calibration}, pass rates converge from both
directions, suggesting that \our{}'s gains come from the co-evolution loop rather
than a specially curated initial difficulty distribution.

\section{Co-Evolution Dynamics and Qualitative Analysis}
\label{appendix:dynamics}

\subsection{Bidirectional Difficulty Calibration}
\label{appendix:bidirectional-calibration}

The EnvRewriter calibrates difficulty from both directions. On ChartQA, the initial Actor is strong, so the EnvRewriter lowers the pass rate toward the sweet zone. On ZwZ, the initial Actor is weak, so the EnvRewriter raises the pass rate toward the sweet zone. This supports the claim that \our{} performs actor-relative calibration rather than random perturbation.

\input{tables/tab-passrate}

\subsection{Editing Intensity and Response Length}
\label{appendix:editing-intensity}
\label{appendix:appendix_editing_intensity_saturation}

\begin{figure}[!htbp]
  \centering
  \includegraphics[width=0.8\linewidth]{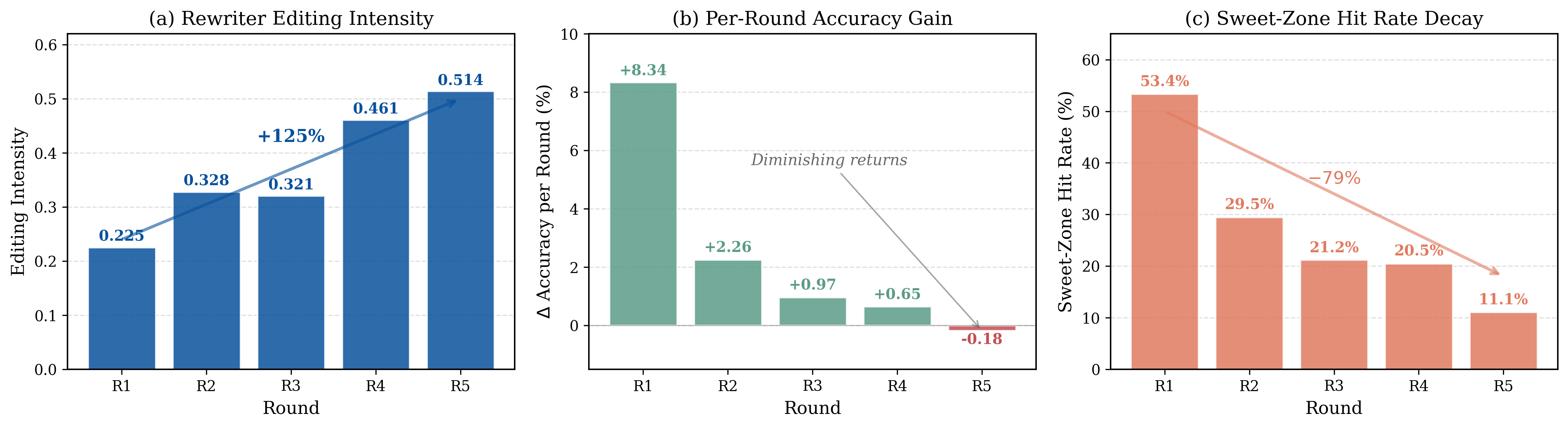}
  \vspace{-0.35em}
  \caption{\textbf{Saturation appears near the Actor's capacity ceiling.} Editing intensity grows by $+125\%$, per-round gains shrink from $+8.34\%$ to $-0.18\%$, and sweet-zone hit rate falls from $53.4\%$ to $11.1\%$, matching Eq.~\ref{eq:optimal_convex}.}
  \vspace{-0.6em}
\label{fig:saturation_evidence}
\end{figure}

We further find that editing intensity and output length evolve differently. As reported in Table~\ref{tab:token_length} and Figure~\ref{fig:saturation_evidence}(a), editing intensity increases from $0.22$ to $0.51$ across rounds, while the average EnvRewriter instruction length remains almost constant ($973 \to 981$ characters). This suggests that the EnvRewriter learns to apply stronger and more targeted edits without producing longer instructions. In contrast, the Actor's response length grows from $811$ to $1207$ characters ($+49\%$), indicating longer reasoning chains as the tasks become more challenging. Together, these trends highlight a clear co-evolutionary process: the EnvRewriter poses increasingly well-calibrated challenges, and the Actor correspondingly develops more elaborate reasoning to tackle them.

\begin{table}[h]
\caption{\textbf{Editing intensity grows without longer EnvRewriter instructions.} Across rounds, edits become stronger while instruction length stays nearly fixed; Actor responses grow as tasks become harder.}
\label{tab:token_length}

\centering
\footnotesize
\setlength{\tabcolsep}{8pt}
\renewcommand{\arraystretch}{1.2}
\begin{tabular}{@{}lccc@{}}
\toprule
\textbf{Metric} & \textbf{Early round} & \textbf{Late round} & \textbf{Change} \\
\midrule
Editing intensity $|\hat{p}_{\text{orig}}-\hat{p}|$ & $0.22$ & $0.51$ & $+125\%$ \\
EnvRewriter instruction length & $973$ & $981$ & $+0.8\%$ \\
Actor response length & $811$ & $1207$ & $+49\%$ \\
\bottomrule
\end{tabular}
\end{table}

This co-adaptation has limits. Figure~\ref{fig:saturation_evidence}(b,c) shows that the per-round accuracy improvement declines from $+8.34\%$ to $-0.18\%$, and the sweet-zone hit rate falls from $53.4\%$ to $11.1\%$. These patterns indicate that, over time, the EnvRewriter struggles to keep up with the Actor's capacity ceiling. We analyze this saturation phenomenon theoretically in Appendix~\ref{appendix:saturation-evidence}.

\subsection{Sample-Level Flip Analysis}
\label{appendix:sample-flip}
\label{appendix:capability-unlock}
To understand the Round~4 gain, we compare sample-level correctness across the
base model and R1--R4 Actors on the ChartQA test set. Wrong-to-correct flips
dominate correct-to-wrong regressions, showing that the gain reflects broad
capability improvement rather than error redistribution. Most newly solved
samples unlock early and remain stable, with stable correctness and R1 unlocks
covering the largest fraction of cases.

\input{tables/tab-flip}



\input{tables/appendix-tab-capability-unlock}

\subsection{Question-Type Breakdown}
\label{appendix:question-type}

The main text shows that gains concentrate on reasoning-heavy ChartQA question types, such as ratio/percentage, value lookup, and difference. These categories require multi-step reading and computation, so their larger improvements support the claim that \our{} improves visual reasoning rather than simply increasing recognition accuracy on easy cases.

\subsection{Chain-of-Thought Case Studies}
\label{appendix:cot-cases}
\label{appendix:cot-analysis}

\textbf{Setup.}
Starting from Qwen3VL-8B-Instruct ($76.9\%$ on ChartQA test), we run four rounds of \our{}. In each round, the EnvRewriter generates approximately $2{,}000$--$3{,}000$ modified training images; the Actor is trained via GRPO on these images together with a fixed regularization set of $2{,}000$ original ChartQA samples; and the resulting Actor checkpoint serves as the starting point for the next round.

\textbf{CoT case analysis.}
Representative cases show that the base model tends to produce direct, single-step answers that miss key constraints, whereas the trained model exhibits explicit step-by-step decomposition, such as identifying relevant chart regions, reading the y-axis, and computing the final answer. Manual inspection of $20$ randomly sampled base-wrong $\to$ R4-correct pairs confirms this pattern is systematic.

\section{Preliminary Experiments on Difficulty-Aware Visual Editing}
\label{appendix:preliminary-difficulty-editing}

These preliminary experiments explored how to construct visual edits whose
difficulty is calibrated to an Actor. They informed the final design but used
different generation and optimization procedures from the main experiments.
In particular, the online Rewriter GRPO experiments described below are distinct
from the final EnvRewriter SFT initialization and RWR updates.

\paragraph{Early image-editing pipeline.}
We first explored a pipeline in which a Rewriter proposed a natural-language
perturbation plan and an external image generator executed it while attempting
to preserve the original question and answer. Scene analysis and object
segmentation provided additional editing context. The Actor then answered the
edited question repeatedly to estimate its pass rate, and the resulting
difficulty signal was used to update the Rewriter with online GRPO. Although
the generated images could shift aggregate task difficulty, these runs did not
show stable calibration toward the target pass rate. Edited samples frequently
remained either too easy or too hard, leaving relatively few samples at
intermediate difficulty.

\paragraph{Answer matching and prompt consistency.}
Offline diagnosis exposed scoring errors that could obscure the effect of an
edit. Substring matching could incorrectly accept a prediction containing the
answer as part of a different value, while overly strict matching of an entire
response could reject a correct answer embedded in explanatory text. For
multiple-choice questions, omitting the choices from the Actor prompt also
created a mismatch between free-form predictions and option-letter labels.
Providing the complete choices and explicitly parsing the final option corrected
these mismatches. This diagnosis motivated consistent answer extraction and
normalization, with numerical precision and units aligned to the question's
requirements. The scoring protocol used in the final experiments is described
in Appendix~\ref{appendix:answer-normalization}.

\paragraph{Calibration requires distribution-level diagnostics.}
The mean pass rate alone did not adequately characterize generation quality.
A mixture of samples with pass rates near zero and one can have a mean near
$0.5$ without producing intermediate-difficulty samples. We therefore distinguish
the pass rate of an individual edited sample, its average across samples, and
the fraction of samples in the target band. Difficulty statistics must also
identify whether their denominator includes failed or invalid generations.
Repeated sampling yields a finite-resolution estimate of success probability;
an observed pass rate is consequently an empirical measurement under a specific
Actor and sampling configuration, rather than an exact difficulty label.

\paragraph{Reward and optimization observations.}
We explored both threshold-based rewards and a continuous improvement reward
$|\hat p_{\mathrm{base}}-0.5|-|\hat p_{\mathrm{edit}}-0.5|$.
The latter measures whether an edit moves the Actor closer to the target.
However, when both the original and edited samples have extreme pass rates,
their distances to the target are equal and this reward provides no preference
between the edits. The recorded runs exhibited limited reward differentiation
and did not consistently increase the supply of calibrated samples. These
observations concern the tested pipeline and do not establish a general
limitation of GRPO. A near-zero reported loss alone is also insufficient to
establish that parameter gradients vanish.

\paragraph{Implications for the final design.}
These experiments motivated separating edit validity from difficulty assessment,
using explicit editing interfaces, and training the EnvRewriter on verified
edit transitions with measured Actor feedback. The final method uses the
domain-specific validity rules in
Appendix~\ref{appendix:programmatic-data-generation} and the RWR procedure in
Appendix~\ref{appendix:rewriter-training}. The preliminary improvement reward
and online Rewriter GRPO procedure are not the optimization recipe used for
the main results.

\section{Case Study: Co-Evolution Matters More Than Image-Generation Strength}
\label{appendix:case_dynamic_editing}

This case study explains why \our{} uses verifiable, actor-conditioned editing: the goal is to adjust difficulty while keeping labels clean.

\paragraph{Why not simple perturbations or unconstrained generation?}
Noise, blur, and occlusion can lower the Actor's pass rate, but they mainly add perceptual difficulty, not reasoning difficulty. Unconstrained generation may produce plausible edits, but it can change answer-relevant content, introduce ambiguity, or invalidate labels. Such failures harm RLVR because rewards require reliable verification.

\our{} avoids this through explicit validity constraints. CLEVR recomputes answers from edited scene graphs. ChartQA uses a two-tier table rule: Tier~1 recovers $a'$ by entity-anchored lookup, while Tier~2 preserves $a'=a$ by fixing answer-bearing CSV values. ZwZ protects answer-relevant regions and edits only non-target regions.

\paragraph{Strong generators are complementary, not replacements.}
GPT-5 and Nano Banana can serve as strong editing backends, but they do not replace the actor-conditioned environment policy. Used once, they create a fixed corpus that can saturate, remain too hard, or expose shortcuts. Used inside the loop, they still need a policy to select valid edits and target the learnable zone. Thus, the key factor is dynamic, verifiable, actor-conditioned co-evolution, not image-generation capability alone.

%% file: tables/tab-sweet-zone-decay.tex
\begin{table}[t]
\centering
\small
\setlength{\tabcolsep}{14pt}
\renewcommand{\arraystretch}{1.2}
\caption{%
\textbf{Sweet-zone hits decline as the Actor approaches saturation.}
The fraction of EnvRewriter trajectories in $[0.35, 0.65]$ falls monotonically as the Actor matures, matching the saturation regime predicted by Eq.~\ref{eq:optimal_convex}.
}
\label{tab:sweet_zone_decay}
\begin{tabular}{@{}l
                S[table-format=2.1]@{}}
\toprule
\textbf{Round} & {\textbf{Sweet-zone hit rate (\%)}} \\
\midrule
R1 & 53.4 \\
R2 & 29.5 \\
R3 & 21.2 \\
R4 & 20.5 \\
R5 & 11.1 \\
\bottomrule
\end{tabular}
\end{table}

%% file: tables/tab-round-growth.tex
\begin{table}[h]
\centering
\small
\setlength{\tabcolsep}{12pt}
\renewcommand{\arraystretch}{1.2}
\caption{%
\textbf{ChartQA gains diminish as co-evolution saturates.}
Accuracy improves through R4, while round-over-round gain $\Delta$ shrinks and turns slightly negative at R5, consistent with the saturation regime predicted by our drift-corrected theory.
}
\label{tab:round_growth}
\begin{tabular}{@{}l
                S[table-format=2.2]
                S[table-format=+2.2]@{}}
\toprule
\textbf{Round} & {\textbf{ChartQA (Qwen3-VL-8B)}} & {$\bm{\Delta}$} \\
\midrule
Base & 76.90 & {---} \\
\addlinespace[2pt]
R1   & 85.24 &  +8.34 \\
R2   & 87.50 &  +2.26 \\
R3   & 88.47 &  +0.97 \\
\rowcolor{gray!10}
R4   & \bfseries 89.12 & \bfseries +0.65 \\
R5   & 88.94 & -0.18 \\
\bottomrule
\end{tabular}
\end{table}

%% file: figures/fig-cross-modality-per-round-acc.tex
\begin{figure}
  \centering
  \includegraphics[width=0.45\linewidth]{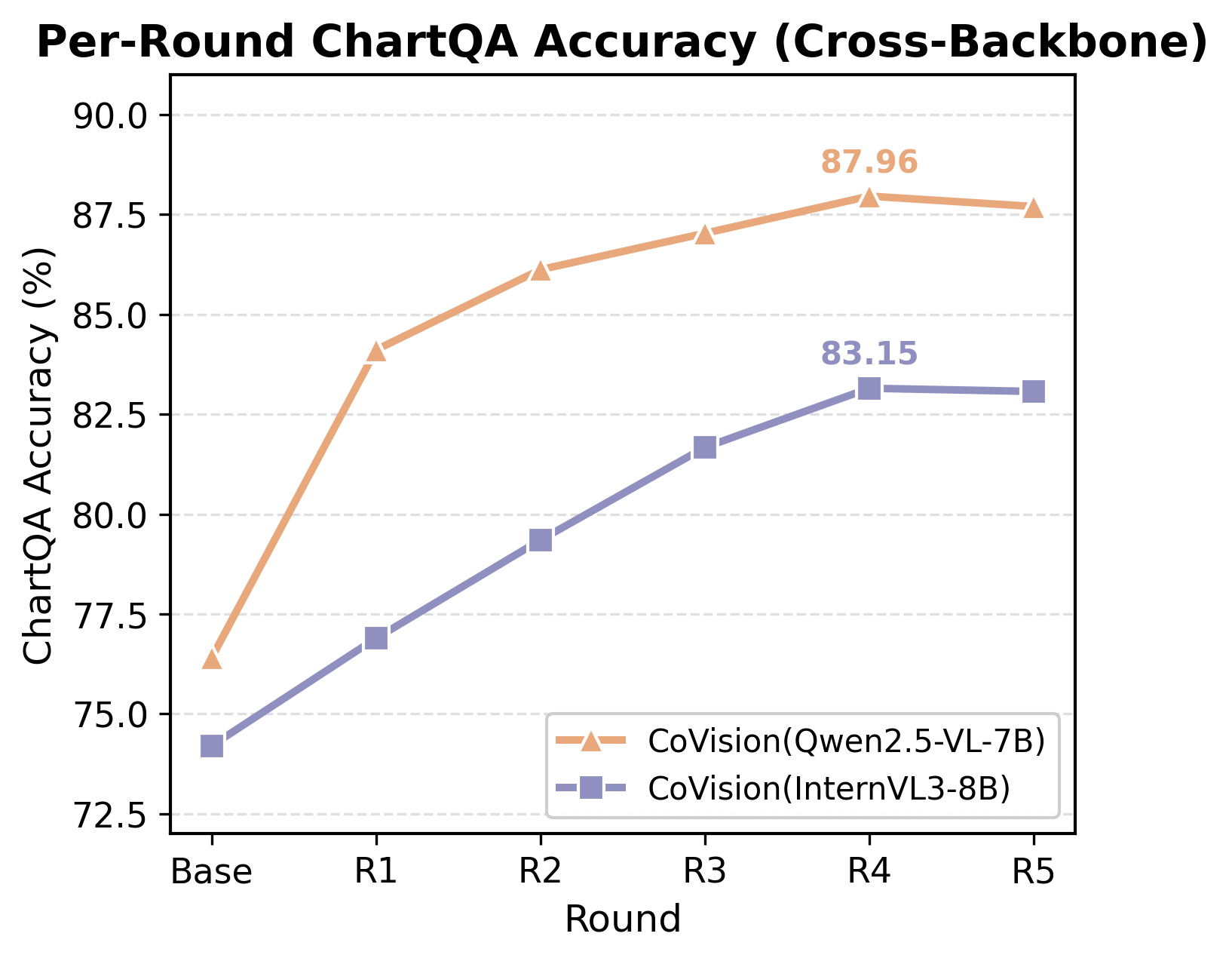}
  \caption{\textbf{\our{} improves accuracy across VLM backbones.} Per-round ChartQA accuracy rises for both tested models, showing that the emergent curriculum is not tied to one model family.}
  \label{fig:cross-modality-per-round-acc}
\end{figure}

%% file: tables/tab-cost-comparison.tex
\begin{table}[t]
\centering
\caption{\textbf{\our{} cuts labeling cost while staying competitive.} Label Cost counts teacher or judging LLM tokens with the Qwen2.5 tokenizer. Programmatic-label methods, including \our{}, use no LLM labeling tokens; baseline training costs are estimated from reported RL samples, while \our{} cost is directly measured.}
\label{tab:cost_comparison}
\resizebox{\textwidth}{!}{%
\begin{tabular}{lccccccccc}
\toprule
\multirow{2}{*}{\textbf{Method}} & \multicolumn{3}{c}{\textbf{Data Cost}} & \multicolumn{3}{c}{\textbf{Training}} & \multicolumn{2}{c}{\textbf{Performance}} \\
\cmidrule(lr){2-4} \cmidrule(lr){5-7} \cmidrule(lr){8-9}
 & \textbf{Prepare Method} & \textbf{Num (RL)} & \textbf{Label Cost (Tokens)} & \textbf{Method} & \textbf{Interact} & \textbf{Time Cost} & \textbf{MathVerse} & \textbf{Mathvista} \\
\midrule
Qwen2.5-VL-7B                    & --                              & --     & --      & --              & --     & --                       & 39.6 & 68.2 \\
\midrule
R1-OneVision-7B                  & Programmatic + human checks      & 10k    & $\geq$ 1.1 M  & SFT+GRPO        & \xmark & $\geq$ 170 A100-Hours    & 46.4 & 64.1 \\
VLAA-Thinker-7B                  & Teacher LLM-generated traces     & 25k    & 29.6 M  & SFT+GRPO        & \xmark & $\geq$ 120 A100-Hours    & 48.2  &68.0   \\
OpenVLThinker-7B                 & Teacher LLM-generated solutions  & 9k     & 5.7 M   & SFT+GRPO        & \xmark & $\geq$ 125 A100-Hours    & 47.9 & 70.2 \\
MM-Eureka-Qwen-7B                & Human-curated                    & 15k    & --      & GRPO            & \xmark & $\approx$ 700 A100-Hours & 50.3 & 73.0\\
ViGaL-Snake                      & Game environment via PPO         & 72k    & 0       & RLOO            & \xmark & $\approx$ 170 A100-Hours & 51.1 & 70.7  \\
ViGaL-Rotation                   & Game environment via PPO         & 72k    & 0       & RLOO            & \xmark & $\approx$ 170 A100-Hours & 50.4  & 71.2  \\
\midrule
\rowcolor{blue!8}
\textbf{\our{}-Qwen2.5VL-7B}(Chart)    & \textbf{Co-evolutionary editing} & \textbf{2000}   & \textbf{0}      & \textbf{Co-evolve+GRPO} & \cmark & \textbf{$\approx$ 149 A100-Hours} & \textbf{54.9} & \textbf{72.8} \\
\bottomrule
\end{tabular}%
}
\end{table}

%% file: tables/tab-passrate.tex
\begin{table}[t]
\centering
\caption{
\textbf{The EnvRewriter moves both strong and weak Actors toward the sweet zone.}
On ChartQA, it lowers the pass rate of a strong Actor; on ZwZ, it raises the pass rate of a weak Actor. Both converge toward $[0.35, 0.65]$, showing actor-relative difficulty calibration.
}
\label{tab:passrate}
\renewcommand\arraystretch{1.05}
\begin{tabular}{l c c c c}
\toprule
\textbf{Round} & \textbf{n} & $\hat{p}_{\text{orig}}$ & $\hat{p}_{\text{final}}$ & \textbf{Editing Intensity} \\
\midrule
\multicolumn{5}{l}{\textit{ChartQA (strong Actor $\rightarrow$ EnvRewriter pushes down)}} \\
\midrule
R1 & 2200 & 0.846 & 0.622 & +0.225 \\
R2 & 2200 & 0.829 & 0.502 & +0.328 \\
R3 & 2200 & 0.822 & 0.502 & +0.321 \\
R4 & 2200 & 0.848 & 0.387 & +0.461 \\
R5 & 2200 & 0.889 & 0.375 & +0.514 \\
\midrule
\multicolumn{5}{l}{\textit{ZwZ (weak Actor $\rightarrow$ EnvRewriter pulls up)}} \\
\midrule
R1 & 2000 & 0.209 & 0.651 & $-0.442$ \\
R2 & 2000 & 0.229 & 0.583 & $-0.354$ \\
R3 & 2000 & 0.223 & 0.524 & $-0.301$ \\
R4 & 2000 & 0.231 & 0.433 & $-0.203$ \\
R5 & 2000 & 0.227 & 0.442 & $-0.215$ \\
\bottomrule
\end{tabular}
\end{table}

%% file: tables/tab-flip.tex

\begin{table}[t]
\centering
\small
\setlength{\tabcolsep}{12pt}
\renewcommand{\arraystretch}{1.25}
\caption{%
\textbf{Positive capability flips dominate regressions.}
On ChartQA, wrong-to-correct flips from Base to R4 outnumber correct-to-wrong flips by $\mathbf{5.7\times 1}$, showing that \our{} improves capability rather than merely moving errors across samples.
}
\label{tab:flip}
\begin{tabular}{@{}lcc@{}}
\toprule
\textbf{Outcome} & \textbf{n} & \textbf{\%} \\
\midrule
Both correct (Base\,\ding{51}, R4\,\ding{51})       & 1863 & 74.5 \\
Improvement (Base\,\ding{55}\,$\to$\,R4\,\ding{51}) & 365  & 14.6 \\
Regression  (Base\,\ding{51}\,$\to$\,R4\,\ding{55}) & 64   & 2.6  \\
Both wrong (Base\,\ding{55}, R4\,\ding{55})         & 208  & 8.3  \\
\bottomrule
\end{tabular}
\end{table}

%% file: tables/appendix-tab-capability-unlock.tex
\begin{table}[t]
\centering
\caption{%
\textbf{Most new ChartQA capabilities unlock early and remain stable.}
Each sample is represented by a 5-bit correctness sequence across Base and R1--R4. The top panel shows first-correct rounds; the bottom lists common trajectories. The top five patterns cover $92.5\%$ of samples, led by stable correctness (\texttt{11111}) and stable R1 unlocks (\texttt{01111}).
}
\label{appendix-tab:capability_unlock}
\renewcommand\arraystretch{1.05}
\begin{tabular}{l c c}
\toprule
\textbf{State} & \textbf{n} & \textbf{\%} \\
\midrule
Already correct at Base         & 1927 & 77.1  \\
Newly unlocked at R1            & 347  & 13.9  \\
Newly unlocked at R2            & 37   & 1.5   \\
Newly unlocked at R3            & 15   & 0.6   \\
Newly unlocked at R4            & 3    & 0.1   \\
Persistent failures             & 171  & 6.8   \\
\midrule
\textbf{Mutually exclusive sum} & \textbf{2500} & \textbf{100.0} \\
\midrule
Non-monotonic (any regression)  & 197  & 7.9   \\
\bottomrule
\end{tabular}
\vspace{6pt}
\renewcommand\arraystretch{1.05}
\begin{tabular}{l c c}
\toprule
\textbf{Pattern} & \textbf{n} & \textbf{\%} \\
\midrule
\texttt{11111}             & 1796 & 71.8 \\
\texttt{01111}             & 306  & 12.2 \\
\texttt{00000}             & 171  & 6.8  \\
\texttt{10111}             & 22   & 0.9  \\
\texttt{10000}             & 18   & 0.7  \\
\midrule
\textit{Top 5}             & 2313 & 92.5 \\
\textit{Other 26 patterns} & 187  & 7.5  \\
\midrule
\textbf{Total}             & \textbf{2500} & \textbf{100.0} \\
\bottomrule
\end{tabular}
\end{table}

%% file: checklist.tex
\section*{NeurIPS Paper Checklist}

\begin{enumerate}

\item {\bf Claims}
    \item[] Question: Do the main claims made in the abstract and introduction accurately reflect the paper's contributions and scope?
    \item[] Answer: \answerYes{} 
    \item[] Justification: This work uses public visual reasoning benchmarks and does not involve human subjects, private personal data, or deceptive data collection. The research is intended for benchmark-style VLM reasoning and follows the NeurIPS Code of Ethics.{}

\item {\bf Limitations}
    \item[] Question: Does the paper discuss the limitations of the work performed by the authors?
    \item[] Answer: \answerYes{} 
    \item[] Justification: The paper discusses limitations related to verifiable image-side structures, imperfect image editing, compute overhead, and saturation in later co-evolution rounds. These limitations are addressed in the Limitations and Impact Statement and further supported by the robustness, compute, and saturation analyses in the appendix.

\item {\bf Theory assumptions and proofs}
    \item[] Question: For each theoretical result, does the paper provide the full set of assumptions and a complete (and correct) proof?
    \item[] Answer: \answerYes{} 
    \item[] Justification: The paper includes a theoretical analysis of difficulty calibration, including the gradient-signal argument based on $\hat{p}(1-\hat{p})$ and a drift-corrected target that explains saturation. The assumptions, derivations, and empirical saturation evidence are provided in Section~4 and Appendix~E.

    \item {\bf Experimental result reproducibility}
    \item[] Question: Does the paper fully disclose all the information needed to reproduce the main experimental results of the paper to the extent that it affects the main claims and/or conclusions of the paper (regardless of whether the code and data are provided or not)?
    \item[] Answer: \answerYes{} 
    \item[] Justification: The paper specifies model backbones, training datasets, evaluation benchmarks metrics, OOD protocol, baseline settings, training recipe, SFT and RL hyperparameters, baselines, reward design, editing interfaces, rollout settings, hyperparameters, and prompt templates. These details provide a reasonable path to reproduce or verify the main experimental results.


\item {\bf Open access to data and code}
    \item[] Question: Does the paper provide open access to the data and code, with sufficient instructions to faithfully reproduce the main experimental results, as described in supplemental material?
    \item[] Answer: \answerYes{} 
    \item[] Justification: We plan to publicly release the code, training dataset, prompt templates, and generated data with instructions for data preparation and reproduction of the main results.

\item {\bf Experimental setting/details}
    \item[] Question: Does the paper specify all the training and test details (e.g., data splits, hyperparameters, how they were chosen, type of optimizer) necessary to understand the results?
    \item[] Answer: \answerYes{} 
    \item[] Justification: The paper reports the training setup, model backbones, seed data sources, evaluation benchmarks, OOD setting, reward functions, rollout parameters, and inference settings. Additional hyperparameters, implementation details, and prompt templates are provided in the appendix.


\item {\bf Experiment statistical significance}
    \item[] Question: Does the paper report error bars suitably and correctly defined or other appropriate information about the statistical significance of the experiments?
    \item[] Answer: \answerNo{} 
    \item[] Justification: The paper reports deterministic benchmark accuracy results and extensive ablations, but does not include error bars, confidence intervals, or statistical significance tests. We acknowledge this as a limitation and plan to include repeated-run statistics in future versions.

\item {\bf Experiments compute resources}
    \item[] Question: For each experiment, does the paper provide sufficient information on the computer resources (type of compute workers, memory, time of execution) needed to reproduce the experiments?
    \item[] Answer: \answerYes{} 
    \item[] Justification:The paper reports the main hardware setup, including up to 8 A100-80GB GPUs, and provides compute estimates for the Qwen2.5-VL-7B chart-backbone run. Additional training cost and annotation-efficiency comparisons are provided in Appendix~G.3.
    
\item {\bf Code of ethics}
    \item[] Question: Does the research conducted in the paper conform, in every respect, with the NeurIPS Code of Ethics \url{https://neurips.cc/public/EthicsGuidelines}?
    \item[] Answer: \answerYes{} 
    \item[] Justification: The work uses public visual reasoning benchmarks and does not involve human subjects, private personal data, or deceptive data collection. The research is intended for benchmark-style VLM reasoning and follows the NeurIPS Code of Ethics.

\item {\bf Broader impacts}
    \item[] Question: Does the paper discuss both potential positive societal impacts and negative societal impacts of the work performed?
    \item[] Answer: \answerYes{} 
    \item[] Justification:  The paper discusses broader impacts in the Limitations and Impact Statement. On the positive side, \our{} can reduce reliance on large-scale human annotation for VLM post-training by turning existing visual data into adaptive, label-valid training samples. On the negative side, automatic image editing may introduce artifacts, label drift, or biased visual changes if not properly verified, and the paper discusses verification and filtering as mitigation strategies.
    
    
\item {\bf Safeguards}
    \item[] Question: Does the paper describe safeguards that have been put in place for responsible release of data or models that have a high risk for misuse (e.g., pre-trained language models, image generators, or scraped datasets)?
    \item[] Answer: \answerNA{} 
    \item[] Justification: The paper does not release high-risk pretrained models, scraped sensitive datasets, or data intended for identity, surveillance, or safety-critical deployment. The planned release focuses on code, prompts, verifiable editing procedures, and benchmark-style generated training data derived from public visual reasoning datasets.

\item {\bf Licenses for existing assets}
    \item[] Question: Are the creators or original owners of assets (e.g., code, data, models), used in the paper, properly credited and are the license and terms of use explicitly mentioned and properly respected?
    \item[] Answer: \answerYes{} 
    \item[] Justification: The paper cites the original sources for all external datasets, benchmarks, model backbones, and tools used in the experiments, including CLEVR, ChartQA, ZwZ, MathVista, MathVision, MathVerse, WeMath, LogicVista, MMMU, MMVP, BLINK, Qwen, InternVL, SAM3, and prior baseline methods. We respect the licenses and terms of use of these existing assets and use them only for benchmark-style research and evaluation.

\item {\bf New assets}
    \item[] Question: Are new assets introduced in the paper well documented and is the documentation provided alongside the assets?
    \item[] Answer:  \answerYes{} 
    \item[] Justification:The paper introduces new assets including the \our{} training pipeline, EnvRewriter prompts, verifiable editing procedures, generated edited samples, and training trajectories. These assets are documented in the methodology and appendices, including the editing protocols, reward functions, filtering rules, training hyperparameters, implementation details, and prompt templates, and will be released with accompanying documentation.
   

\item {\bf Crowdsourcing and research with human subjects}
    \item[] Question: For crowdsourcing experiments and research with human subjects, does the paper include the full text of instructions given to participants and screenshots, if applicable, as well as details about compensation (if any)? 
    \item[] Answer: \answerNA{} 
    \item[] Justification:The paper does not involve crowdsourcing experiments or research with human subjects.

\item {\bf Institutional review board (IRB) approvals or equivalent for research with human subjects}
    \item[] Question: Does the paper describe potential risks incurred by study participants, whether such risks were disclosed to the subjects, and whether Institutional Review Board (IRB) approvals (or an equivalent approval/review based on the requirements of your country or institution) were obtained?
    \item[] Answer: \answerNA{} 
    \item[] Justification: The paper does not involve crowdsourcing experiments or research with human subjects, so IRB approval or equivalent review is not applicable.

\item {\bf Declaration of LLM usage}
    \item[] Question: Does the paper describe the usage of LLMs if it is an important, original, or non-standard component of the core methods in this research? Note that if the LLM is used only for writing, editing, or formatting purposes and does \emph{not} impact the core methodology, scientific rigor, or originality of the research, declaration is not required.
    \item[] Answer:  \answerYes{} 
    \item[] Justification: The paper explicitly describes the use of large vision-language models in the core methodology: the Actor and EnvRewriter are initialized from VLM backbones and are jointly trained in the co-evolution loop. The paper also describes the use of GPT-5 and Gemini (Nano Banana) as stronger offline augmentation baselines, and reports the model backbones, prompts, reward functions, and training settings used in the experiments.


\end{enumerate}

%% file: refs.bib
@misc{carion2026sam3segmentconcepts,
      title={SAM 3: Segment Anything with Concepts}, 
      author={Nicolas Carion and Laura Gustafson and Yuan-Ting Hu and Shoubhik Debnath and Ronghang Hu and Didac Suris and Chaitanya Ryali and Kalyan Vasudev Alwala and Haitham Khedr and Andrew Huang and Jie Lei and Tengyu Ma and Baishan Guo and Arpit Kalla and Markus Marks and Joseph Greer and Meng Wang and Peize Sun and Roman Rädle and Triantafyllos Afouras and Effrosyni Mavroudi and Katherine Xu and Tsung-Han Wu and Yu Zhou and Liliane Momeni and Rishi Hazra and Shuangrui Ding and Sagar Vaze and Francois Porcher and Feng Li and Siyuan Li and Aishwarya Kamath and Ho Kei Cheng and Piotr Dollár and Nikhila Ravi and Kate Saenko and Pengchuan Zhang and Christoph Feichtenhofer},
      year={2026},
      eprint={2511.16719},
      archivePrefix={arXiv},
      primaryClass={cs.CV},
      url={https://arxiv.org/abs/2511.16719}, 
}

@misc{shao2024deepseekmathpushinglimitsmathematical,
      title={DeepSeekMath: Pushing the Limits of Mathematical Reasoning in Open Language Models}, 
      author={Zhihong Shao and Peiyi Wang and Qihao Zhu and Runxin Xu and Junxiao Song and Xiao Bi and Haowei Zhang and Mingchuan Zhang and Y. K. Li and Y. Wu and Daya Guo},
      year={2024},
      eprint={2402.03300},
      archivePrefix={arXiv},
      primaryClass={cs.CL},
      url={https://arxiv.org/abs/2402.03300}, 
}

@misc{li2026mmzeroselfevolvingmultimodelvision,
      title={MM-Zero: Self-Evolving Multi-Model Vision Language Models From Zero Data}, 
      author={Zongxia Li and Hongyang Du and Chengsong Huang and Xiyang Wu and Lantao Yu and Yicheng He and Jing Xie and Xiaomin Wu and Zhichao Liu and Jiarui Zhang and Fuxiao Liu},
      year={2026},
      eprint={2603.09206},
      archivePrefix={arXiv},
      primaryClass={cs.CV},
      url={https://arxiv.org/abs/2603.09206}, 
}

@misc{wang2025visionzero,
  title         = {Vision-Zero: Scalable VLM Self-Improvement via Strategic Gamified Self-Play},
  author        = {Wang, Qinsi and Liu, Bo and Zhou, Tianyi and Shi, Jing and Lin, Yueqian and Chen, Yiran and Li, Hai Helen and Wan, Kun and Zhao, Wentian},
  year          = {2025},
  eprint        = {2509.25541},
  archivePrefix = {arXiv},
  primaryClass  = {cs.CV},
  url           = {https://arxiv.org/abs/2509.25541}
}

@misc{guo2025genenvdifficultyalignedcoevolutionllm,
      title={GenEnv: Difficulty-Aligned Co-Evolution Between LLM Agents and Environment Simulators}, 
      author={Jiacheng Guo and Ling Yang and Peter Chen and Qixin Xiao and Yinjie Wang and Xinzhe Juan and Jiahao Qiu and Ke Shen and Mengdi Wang},
      year={2025},
      eprint={2512.19682},
      archivePrefix={arXiv},
      primaryClass={cs.CL},
      url={https://arxiv.org/abs/2512.19682}, 
}

@misc{lu2024mathvistaevaluatingmathematicalreasoning,
      title={MathVista: Evaluating Mathematical Reasoning of Foundation Models in Visual Contexts}, 
      author={Pan Lu and Hritik Bansal and Tony Xia and Jiacheng Liu and Chunyuan Li and Hannaneh Hajishirzi and Hao Cheng and Kai-Wei Chang and Michel Galley and Jianfeng Gao},
      year={2024},
      eprint={2310.02255},
      archivePrefix={arXiv},
      primaryClass={cs.CV},
      url={https://arxiv.org/abs/2310.02255}, 
}

@misc{wang2024measuringmultimodalmathematicalreasoning,
      title={Measuring Multimodal Mathematical Reasoning with MATH-Vision Dataset}, 
      author={Ke Wang and Junting Pan and Weikang Shi and Zimu Lu and Mingjie Zhan and Hongsheng Li},
      year={2024},
      eprint={2402.14804},
      archivePrefix={arXiv},
      primaryClass={cs.CV},
      url={https://arxiv.org/abs/2402.14804}, 
}

@misc{zhang2024mathversedoesmultimodalllm,
      title={MathVerse: Does Your Multi-modal LLM Truly See the Diagrams in Visual Math Problems?}, 
      author={Renrui Zhang and Dongzhi Jiang and Yichi Zhang and Haokun Lin and Ziyu Guo and Pengshuo Qiu and Aojun Zhou and Pan Lu and Kai-Wei Chang and Peng Gao and Hongsheng Li},
      year={2024},
      eprint={2403.14624},
      archivePrefix={arXiv},
      primaryClass={cs.CV},
      url={https://arxiv.org/abs/2403.14624}, 
}

@misc{qiao2024wemathdoeslargemultimodal,
      title={We-Math: Does Your Large Multimodal Model Achieve Human-like Mathematical Reasoning?}, 
      author={Runqi Qiao and Qiuna Tan and Guanting Dong and Minhui Wu and Chong Sun and Xiaoshuai Song and Zhuoma GongQue and Shanglin Lei and Zhe Wei and Miaoxuan Zhang and Runfeng Qiao and Yifan Zhang and Xiao Zong and Yida Xu and Muxi Diao and Zhimin Bao and Chen Li and Honggang Zhang},
      year={2024},
      eprint={2407.01284},
      archivePrefix={arXiv},
      primaryClass={cs.AI},
      url={https://arxiv.org/abs/2407.01284}, 
}

@misc{xiao2024logicvistamultimodalllmlogical,
      title={LogicVista: Multimodal LLM Logical Reasoning Benchmark in Visual Contexts}, 
      author={Yijia Xiao and Edward Sun and Tianyu Liu and Wei Wang},
      year={2024},
      eprint={2407.04973},
      archivePrefix={arXiv},
      primaryClass={cs.AI},
      url={https://arxiv.org/abs/2407.04973}, 
}

@misc{masry2022chartqabenchmarkquestionanswering,
      title={ChartQA: A Benchmark for Question Answering about Charts with Visual and Logical Reasoning}, 
      author={Ahmed Masry and Do Xuan Long and Jia Qing Tan and Shafiq Joty and Enamul Hoque},
      year={2022},
      eprint={2203.10244},
      archivePrefix={arXiv},
      primaryClass={cs.CL},
      url={https://arxiv.org/abs/2203.10244}, 
}

@misc{yue2024mmmumassivemultidisciplinemultimodal,
      title={MMMU: A Massive Multi-discipline Multimodal Understanding and Reasoning Benchmark for Expert AGI}, 
      author={Xiang Yue and Yuansheng Ni and Kai Zhang and Tianyu Zheng and Ruoqi Liu and Ge Zhang and Samuel Stevens and Dongfu Jiang and Weiming Ren and Yuxuan Sun and Cong Wei and Botao Yu and Ruibin Yuan and Renliang Sun and Ming Yin and Boyuan Zheng and Zhenzhu Yang and Yibo Liu and Wenhao Huang and Huan Sun and Yu Su and Wenhu Chen},
      year={2024},
      eprint={2311.16502},
      archivePrefix={arXiv},
      primaryClass={cs.CL},
      url={https://arxiv.org/abs/2311.16502}, 
}

@misc{fu2024blinkmultimodallargelanguage,
      title={BLINK: Multimodal Large Language Models Can See but Not Perceive}, 
      author={Xingyu Fu and Yushi Hu and Bangzheng Li and Yu Feng and Haoyu Wang and Xudong Lin and Dan Roth and Noah A. Smith and Wei-Chiu Ma and Ranjay Krishna},
      year={2024},
      eprint={2404.12390},
      archivePrefix={arXiv},
      primaryClass={cs.CV},
      url={https://arxiv.org/abs/2404.12390}, 
}

@misc{bai2025qwen3vltechnicalreport,
      title={Qwen3-VL Technical Report}, 
      author={Shuai Bai and Yuxuan Cai and Ruizhe Chen and Keqin Chen and Xionghui Chen and Zesen Cheng and Lianghao Deng and Wei Ding and Chang Gao and Chunjiang Ge and Wenbin Ge and Zhifang Guo and Qidong Huang and Jie Huang and Fei Huang and Binyuan Hui and Shutong Jiang and Zhaohai Li and Mingsheng Li and Mei Li and Kaixin Li and Zicheng Lin and Junyang Lin and Xuejing Liu and Jiawei Liu and Chenglong Liu and Yang Liu and Dayiheng Liu and Shixuan Liu and Dunjie Lu and Ruilin Luo and Chenxu Lv and Rui Men and Lingchen Meng and Xuancheng Ren and Xingzhang Ren and Sibo Song and Yuchong Sun and Jun Tang and Jianhong Tu and Jianqiang Wan and Peng Wang and Pengfei Wang and Qiuyue Wang and Yuxuan Wang and Tianbao Xie and Yiheng Xu and Haiyang Xu and Jin Xu and Zhibo Yang and Mingkun Yang and Jianxin Yang and An Yang and Bowen Yu and Fei Zhang and Hang Zhang and Xi Zhang and Bo Zheng and Humen Zhong and Jingren Zhou and Fan Zhou and Jing Zhou and Yuanzhi Zhu and Ke Zhu},
      year={2025},
      eprint={2511.21631},
      archivePrefix={arXiv},
      primaryClass={cs.CV},
      url={https://arxiv.org/abs/2511.21631}, 
}

@misc{yang2025r1onevisionadvancinggeneralizedmultimodal,
      title={R1-Onevision: Advancing Generalized Multimodal Reasoning through Cross-Modal Formalization}, 
      author={Yi Yang and Xiaoxuan He and Hongkun Pan and Xiyan Jiang and Yan Deng and Xingtao Yang and Haoyu Lu and Dacheng Yin and Fengyun Rao and Minfeng Zhu and Bo Zhang and Wei Chen},
      year={2025},
      eprint={2503.10615},
      archivePrefix={arXiv},
      primaryClass={cs.CV},
      url={https://arxiv.org/abs/2503.10615}, 
}

@misc{meng2025mmeurekaexploringfrontiersmultimodal,
      title={MM-Eureka: Exploring the Frontiers of Multimodal Reasoning with Rule-based Reinforcement Learning}, 
      author={Fanqing Meng and Lingxiao Du and Zongkai Liu and Zhixiang Zhou and Quanfeng Lu and Daocheng Fu and Tiancheng Han and Botian Shi and Wenhai Wang and Junjun He and Kaipeng Zhang and Ping Luo and Yu Qiao and Qiaosheng Zhang and Wenqi Shao},
      year={2025},
      eprint={2503.07365},
      archivePrefix={arXiv},
      primaryClass={cs.CV},
      url={https://arxiv.org/abs/2503.07365}, 
}

@misc{deng2025openvlthinkercomplexvisionlanguagereasoning,
      title={OpenVLThinker: Complex Vision-Language Reasoning via Iterative SFT-RL Cycles}, 
      author={Yihe Deng and Hritik Bansal and Fan Yin and Nanyun Peng and Wei Wang and Kai-Wei Chang},
      year={2025},
      eprint={2503.17352},
      archivePrefix={arXiv},
      primaryClass={cs.CV},
      url={https://arxiv.org/abs/2503.17352}, 
}

@misc{wang2026vzeroselfimprovingmultimodalreasoning,
      title={V-Zero: Self-Improving Multimodal Reasoning with Zero Annotation}, 
      author={Han Wang and Yi Yang and Jingyuan Hu and Minfeng Zhu and Wei Chen},
      year={2026},
      eprint={2601.10094},
      archivePrefix={arXiv},
      primaryClass={cs.CV},
      url={https://arxiv.org/abs/2601.10094}, 
}

@inproceedings{Sheng_2025, series={EuroSys ’25},
   title={HybridFlow: A Flexible and Efficient RLHF Framework},
   url={http://dx.doi.org/10.1145/3689031.3696075},
   DOI={10.1145/3689031.3696075},
   booktitle={Proceedings of the Twentieth European Conference on Computer Systems},
   publisher={ACM},
   author={Sheng, Guangming and Zhang, Chi and Ye, Zilingfeng and Wu, Xibin and Zhang, Wang and Zhang, Ru and Peng, Yanghua and Lin, Haibin and Wu, Chuan},
   year={2025},
   month=Mar, pages={1279–1297},
   collection={EuroSys ’25} }

@misc{zhu2025internvl3exploringadvancedtraining,
      title={InternVL3: Exploring Advanced Training and Test-Time Recipes for Open-Source Multimodal Models}, 
      author={Jinguo Zhu and Weiyun Wang and Zhe Chen and Zhaoyang Liu and Shenglong Ye and Lixin Gu and Hao Tian and Yuchen Duan and Weijie Su and Jie Shao and Zhangwei Gao and Erfei Cui and Xuehui Wang and Yue Cao and Yangzhou Liu and Xingguang Wei and Hongjie Zhang and Haomin Wang and Weiye Xu and Hao Li and Jiahao Wang and Nianchen Deng and Songze Li and Yinan He and Tan Jiang and Jiapeng Luo and Yi Wang and Conghui He and Botian Shi and Xingcheng Zhang and Wenqi Shao and Junjun He and Yingtong Xiong and Wenwen Qu and Peng Sun and Penglong Jiao and Han Lv and Lijun Wu and Kaipeng Zhang and Huipeng Deng and Jiaye Ge and Kai Chen and Limin Wang and Min Dou and Lewei Lu and Xizhou Zhu and Tong Lu and Dahua Lin and Yu Qiao and Jifeng Dai and Wenhai Wang},
      year={2025},
      eprint={2504.10479},
      archivePrefix={arXiv},
      primaryClass={cs.CV},
      url={https://arxiv.org/abs/2504.10479}, 
}

@misc{bai2025qwen25vltechnicalreport,
      title={Qwen2.5-VL Technical Report}, 
      author={Shuai Bai and Keqin Chen and Xuejing Liu and Jialin Wang and Wenbin Ge and Sibo Song and Kai Dang and Peng Wang and Shijie Wang and Jun Tang and Humen Zhong and Yuanzhi Zhu and Mingkun Yang and Zhaohai Li and Jianqiang Wan and Pengfei Wang and Wei Ding and Zheren Fu and Yiheng Xu and Jiabo Ye and Xi Zhang and Tianbao Xie and Zesen Cheng and Hang Zhang and Zhibo Yang and Haiyang Xu and Junyang Lin},
      year={2025},
      eprint={2502.13923},
      archivePrefix={arXiv},
      primaryClass={cs.CV},
      url={https://arxiv.org/abs/2502.13923}, 
}

@misc{lambert2025tulu3pushingfrontiers,
      title={Tulu 3: Pushing Frontiers in Open Language Model Post-Training}, 
      author={Nathan Lambert and Jacob Morrison and Valentina Pyatkin and Shengyi Huang and Hamish Ivison and Faeze Brahman and Lester James V. Miranda and Alisa Liu and Nouha Dziri and Shane Lyu and Yuling Gu and Saumya Malik and Victoria Graf and Jena D. Hwang and Jiangjiang Yang and Ronan Le Bras and Oyvind Tafjord and Chris Wilhelm and Luca Soldaini and Noah A. Smith and Yizhong Wang and Pradeep Dasigi and Hannaneh Hajishirzi},
      year={2025},
      eprint={2411.15124},
      archivePrefix={arXiv},
      primaryClass={cs.CL},
      url={https://arxiv.org/abs/2411.15124}, 
}

@article{Guo_2025,
   title={DeepSeek-R1 incentivizes reasoning in LLMs through reinforcement learning},
   volume={645},
   ISSN={1476-4687},
   url={http://dx.doi.org/10.1038/s41586-025-09422-z},
   DOI={10.1038/s41586-025-09422-z},
   number={8081},
   journal={Nature},
   publisher={Springer Science and Business Media LLC},
   author={Guo, Daya and Yang, Dejian and Zhang, Haowei and Song, Junxiao and Wang, Peiyi and Zhu, Qihao and Xu, Runxin and Zhang, Ruoyu and Ma, Shirong and Bi, Xiao and Zhang, Xiaokang and Yu, Xingkai and Wu, Yu and Wu, Z. F. and Gou, Zhibin and Shao, Zhihong and Li, Zhuoshu and Gao, Ziyi and Liu, Aixin and Xue, Bing and Wang, Bingxuan and Wu, Bochao and Feng, Bei and Lu, Chengda and Zhao, Chenggang and Deng, Chengqi and Ruan, Chong and Dai, Damai and Chen, Deli and Ji, Dongjie and Li, Erhang and Lin, Fangyun and Dai, Fucong and Luo, Fuli and Hao, Guangbo and Chen, Guanting and Li, Guowei and Zhang, H. and Xu, Hanwei and Ding, Honghui and Gao, Huazuo and Qu, Hui and Li, Hui and Guo, Jianzhong and Li, Jiashi and Chen, Jingchang and Yuan, Jingyang and Tu, Jinhao and Qiu, Junjie and Li, Junlong and Cai, J. L. and Ni, Jiaqi and Liang, Jian and Chen, Jin and Dong, Kai and Hu, Kai and You, Kaichao and Gao, Kaige and Guan, Kang and Huang, Kexin and Yu, Kuai and Wang, Lean and Zhang, Lecong and Zhao, Liang and Wang, Litong and Zhang, Liyue and Xu, Lei and Xia, Leyi and Zhang, Mingchuan and Zhang, Minghua and Tang, Minghui and Zhou, Mingxu and Li, Meng and Wang, Miaojun and Li, Mingming and Tian, Ning and Huang, Panpan and Zhang, Peng and Wang, Qiancheng and Chen, Qinyu and Du, Qiushi and Ge, Ruiqi and Zhang, Ruisong and Pan, Ruizhe and Wang, Runji and Chen, R. J. and Jin, R. L. and Chen, Ruyi and Lu, Shanghao and Zhou, Shangyan and Chen, Shanhuang and Ye, Shengfeng and Wang, Shiyu and Yu, Shuiping and Zhou, Shunfeng and Pan, Shuting and Li, S. S. and Zhou, Shuang and Wu, Shaoqing and Yun, Tao and Pei, Tian and Sun, Tianyu and Wang, T. and Zeng, Wangding and Liu, Wen and Liang, Wenfeng and Gao, Wenjun and Yu, Wenqin and Zhang, Wentao and Xiao, W. L. and An, Wei and Liu, Xiaodong and Wang, Xiaohan and Chen, Xiaokang and Nie, Xiaotao and Cheng, Xin and Liu, Xin and Xie, Xin and Liu, Xingchao and Yang, Xinyu and Li, Xinyuan and Su, Xuecheng and Lin, Xuheng and Li, X. Q. and Jin, Xiangyue and Shen, Xiaojin and Chen, Xiaosha and Sun, Xiaowen and Wang, Xiaoxiang and Song, Xinnan and Zhou, Xinyi and Wang, Xianzu and Shan, Xinxia and Li, Y. K. and Wang, Y. Q. and Wei, Y. X. and Zhang, Yang and Xu, Yanhong and Li, Yao and Zhao, Yao and Sun, Yaofeng and Wang, Yaohui and Yu, Yi and Zhang, Yichao and Shi, Yifan and Xiong, Yiliang and He, Ying and Piao, Yishi and Wang, Yisong and Tan, Yixuan and Ma, Yiyang and Liu, Yiyuan and Guo, Yongqiang and Ou, Yuan and Wang, Yuduan and Gong, Yue and Zou, Yuheng and He, Yujia and Xiong, Yunfan and Luo, Yuxiang and You, Yuxiang and Liu, Yuxuan and Zhou, Yuyang and Zhu, Y. X. and Huang, Yanping and Li, Yaohui and Zheng, Yi and Zhu, Yuchen and Ma, Yunxian and Tang, Ying and Zha, Yukun and Yan, Yuting and Ren, Z. Z. and Ren, Zehui and Sha, Zhangli and Fu, Zhe and Xu, Zhean and Xie, Zhenda and Zhang, Zhengyan and Hao, Zhewen and Ma, Zhicheng and Yan, Zhigang and Wu, Zhiyu and Gu, Zihui and Zhu, Zijia and Liu, Zijun and Li, Zilin and Xie, Ziwei and Song, Ziyang and Pan, Zizheng and Huang, Zhen and Xu, Zhipeng and Zhang, Zhongyu and Zhang, Zhen},
   year={2025},
   month={Sept}, pages={633–638} }

@misc{johnson2016clevrdiagnosticdatasetcompositional,
      title={CLEVR: A Diagnostic Dataset for Compositional Language and Elementary Visual Reasoning}, 
      author={Justin Johnson and Bharath Hariharan and Laurens van der Maaten and Li Fei-Fei and C. Lawrence Zitnick and Ross Girshick},
      year={2016},
      eprint={1612.06890},
      archivePrefix={arXiv},
      primaryClass={cs.CV},
      url={https://arxiv.org/abs/1612.06890}, 
}

@misc{zhou2025r1zerosahamomentvisual,
      title={R1-Zero's "Aha Moment" in Visual Reasoning on a 2B Non-SFT Model}, 
      author={Hengguang Zhou and Xirui Li and Ruochen Wang and Minhao Cheng and Tianyi Zhou and Cho-Jui Hsieh},
      year={2025},
      eprint={2503.05132},
      archivePrefix={arXiv},
      primaryClass={cs.AI},
      url={https://arxiv.org/abs/2503.05132}, 
}

@misc{yang2025visionthinksmartefficientvision,
      title={VisionThink: Smart and Efficient Vision Language Model via Reinforcement Learning}, 
      author={Senqiao Yang and Junyi Li and Xin Lai and Bei Yu and Hengshuang Zhao and Jiaya Jia},
      year={2025},
      eprint={2507.13348},
      archivePrefix={arXiv},
      primaryClass={cs.CV},
      url={https://arxiv.org/abs/2507.13348}, 
}

@misc{liang2025swsselfawareweaknessdrivenproblem,
      title={SwS: Self-aware Weakness-driven Problem Synthesis in Reinforcement Learning for LLM Reasoning}, 
      author={Xiao Liang and Zhong-Zhi Li and Yeyun Gong and Yang Wang and Hengyuan Zhang and Yelong Shen and Ying Nian Wu and Weizhu Chen},
      year={2025},
      eprint={2506.08989},
      archivePrefix={arXiv},
      primaryClass={cs.LG},
      url={https://arxiv.org/abs/2506.08989}, 
}

@misc{dai2025cdecuriositydrivenexplorationefficient,
      title={CDE: Curiosity-Driven Exploration for Efficient Reinforcement Learning in Large Language Models}, 
      author={Runpeng Dai and Linfeng Song and Haolin Liu and Zhenwen Liang and Dian Yu and Haitao Mi and Zhaopeng Tu and Rui Liu and Tong Zheng and Hongtu Zhu and Dong Yu},
      year={2025},
      eprint={2509.09675},
      archivePrefix={arXiv},
      primaryClass={cs.CL},
      url={https://arxiv.org/abs/2509.09675}, 
}

@misc{wang2025vagenreinforcingworldmodel,
      title={VAGEN: Reinforcing World Model Reasoning for Multi-Turn VLM Agents}, 
      author={Kangrui Wang and Pingyue Zhang and Zihan Wang and Yaning Gao and Linjie Li and Qineng Wang and Hanyang Chen and Chi Wan and Yiping Lu and Zhengyuan Yang and Lijuan Wang and Ranjay Krishna and Jiajun Wu and Li Fei-Fei and Yejin Choi and Manling Li},
      year={2025},
      eprint={2510.16907},
      archivePrefix={arXiv},
      primaryClass={cs.AI},
      url={https://arxiv.org/abs/2510.16907}, 
}

@misc{yue2025doesreinforcementlearningreally,
      title={Does Reinforcement Learning Really Incentivize Reasoning Capacity in LLMs Beyond the Base Model?}, 
      author={Yang Yue and Zhiqi Chen and Rui Lu and Andrew Zhao and Zhaokai Wang and Yang Yue and Shiji Song and Gao Huang},
      year={2025},
      eprint={2504.13837},
      archivePrefix={arXiv},
      primaryClass={cs.AI},
      url={https://arxiv.org/abs/2504.13837}, 
}

@misc{he2025visplayselfevolvingvisionlanguagemodels,
      title={VisPlay: Self-Evolving Vision-Language Models from Images}, 
      author={Yicheng He and Chengsong Huang and Zongxia Li and Jiaxin Huang and Yonghui Yang},
      year={2025},
      eprint={2511.15661},
      archivePrefix={arXiv},
      primaryClass={cs.CV},
      url={https://arxiv.org/abs/2511.15661}, 
}

@misc{xie2025playgeneralizelearningreason,
      title={Play to Generalize: Learning to Reason Through Game Play}, 
      author={Yunfei Xie and Yinsong Ma and Shiyi Lan and Alan Yuille and Junfei Xiao and Chen Wei},
      year={2025},
      eprint={2506.08011},
      archivePrefix={arXiv},
      primaryClass={cs.CV},
      url={https://arxiv.org/abs/2506.08011}, 
}

@misc{tong2025gamerlsynthesizingmultimodalverifiable,
      title={Game-RL: Synthesizing Multimodal Verifiable Game Data to Boost VLMs' General Reasoning}, 
      author={Jingqi Tong and Jixin Tang and Hangcheng Li and Yurong Mou and Ming Zhang and Jun Zhao and Yanbo Wen and Fan Song and Jiahao Zhan and Yuyang Lu and Chaoran Tao and Zhiyuan Guo and Jizhou Yu and Tianhao Cheng and Zhiheng Xi and Changhao Jiang and Zhangyue Yin and Yining Zheng and Weifeng Ge and Guanhua Chen and Tao Gui and Xipeng Qiu and Qi Zhang and Xuanjing Huang},
      year={2025},
      eprint={2505.13886},
      archivePrefix={arXiv},
      primaryClass={cs.CL},
      url={https://arxiv.org/abs/2505.13886}, 
}

@misc{zhang2025viperempoweringselfevolutionvisual,
      title={ViPER: Empowering the Self-Evolution of Visual Perception Abilities in Vision-Language Model}, 
      author={Juntian Zhang and Song Jin and Chuanqi Cheng and Yuhan Liu and Yankai Lin and Xun Zhang and Yufei Zhang and Fei Jiang and Guojun Yin and Wei Lin and Rui Yan},
      year={2025},
      eprint={2510.24285},
      archivePrefix={arXiv},
      primaryClass={cs.CV},
      url={https://arxiv.org/abs/2510.24285}, 
}

@misc{wen2026selfevolvingvisionlanguagemodelsimage,
      title={Self-Evolving Vision-Language Models for Image Quality Assessment via Voting and Ranking}, 
      author={Wen Wen and Tianwu Zhi and Kanglong Fan and Yang Li and Xinge Peng and Yabin Zhang and Yiting Liao and Junlin Li and Li Zhang},
      year={2026},
      eprint={2509.25787},
      archivePrefix={arXiv},
      primaryClass={cs.CV},
      url={https://arxiv.org/abs/2509.25787}, 
}

@misc{deng2025selfrefinementvisionlanguagemodelstriangular,
      title={Towards Self-Refinement of Vision-Language Models with Triangular Consistency}, 
      author={Yunlong Deng and Guangyi Chen and Tianpei Gu and Lingjing Kong and Yan Li and Zeyu Tang and Kun Zhang},
      year={2025},
      eprint={2510.10487},
      archivePrefix={arXiv},
      primaryClass={cs.CV},
      url={https://arxiv.org/abs/2510.10487}, 
}

@misc{he2025seeingbelievingmitigatingocr,
      title={Seeing is Believing? Mitigating OCR Hallucinations in Multimodal Large Language Models}, 
      author={Zhentao He and Can Zhang and Ziheng Wu and Zhenghao Chen and Yufei Zhan and Yifan Li and Zhao Zhang and Xian Wang and Minghui Qiu},
      year={2025},
      eprint={2506.20168},
      archivePrefix={arXiv},
      primaryClass={cs.CV},
      url={https://arxiv.org/abs/2506.20168}, 
}

@misc{chen2025c2evocoevolvingmultimodaldata,
      title={C2-Evo: Co-Evolving Multimodal Data and Model for Self-Improving Reasoning}, 
      author={Xiuwei Chen and Wentao Hu and Hanhui Li and Jun Zhou and Zisheng Chen and Meng Cao and Yihan Zeng and Kui Zhang and Yu-Jie Yuan and Jianhua Han and Hang Xu and Xiaodan Liang},
      year={2025},
      eprint={2507.16518},
      archivePrefix={arXiv},
      primaryClass={cs.CV},
      url={https://arxiv.org/abs/2507.16518}, 
}

@misc{wei2026zoomingzoomingregiontoimagedistillation,
      title={Zooming without Zooming: Region-to-Image Distillation for Fine-Grained Multimodal Perception}, 
      author={Lai Wei and Liangbo He and Jun Lan and Lingzhong Dong and Yutong Cai and Siyuan Li and Huijia Zhu and Weiqiang Wang and Linghe Kong and Yue Wang and Zhuosheng Zhang and Weiran Huang},
      year={2026},
      eprint={2602.11858},
      archivePrefix={arXiv},
      primaryClass={cs.CV},
      url={https://arxiv.org/abs/2602.11858}, 
}

@misc{zhang2024mmvpmultimodalmocapdataset,
      title={MMVP: A Multimodal MoCap Dataset with Vision and Pressure Sensors}, 
      author={He Zhang and Shenghao Ren and Haolei Yuan and Jianhui Zhao and Fan Li and Shuangpeng Sun and Zhenghao Liang and Tao Yu and Qiu Shen and Xun Cao},
      year={2024},
      eprint={2403.17610},
      archivePrefix={arXiv},
      primaryClass={cs.CV},
      url={https://arxiv.org/abs/2403.17610}, 
}

@misc{borji2024noteshumailovetal,
      title={A Note on Shumailov et al. (2024): `AI Models Collapse When Trained on Recursively Generated Data'}, 
      author={Ali Borji},
      year={2024},
      eprint={2410.12954},
      archivePrefix={arXiv},
      primaryClass={cs.LG},
      url={https://arxiv.org/abs/2410.12954}, 
}

@misc{alemohammad2023selfconsuminggenerativemodelsmad,
      title={Self-Consuming Generative Models Go MAD}, 
      author={Sina Alemohammad and Josue Casco-Rodriguez and Lorenzo Luzi and Ahmed Imtiaz Humayun and Hossein Babaei and Daniel LeJeune and Ali Siahkoohi and Richard G. Baraniuk},
      year={2023},
      eprint={2307.01850},
      archivePrefix={arXiv},
      primaryClass={cs.LG},
      url={https://arxiv.org/abs/2307.01850}, 
}

@misc{kaledin2022variancereductionpolicygradientmethods,
      title={Variance Reduction for Policy-Gradient Methods via Empirical Variance Minimization}, 
      author={Maxim Kaledin and Alexander Golubev and Denis Belomestny},
      year={2022},
      eprint={2206.06827},
      archivePrefix={arXiv},
      primaryClass={cs.LG},
      url={https://arxiv.org/abs/2206.06827}, 
}

@misc{thomas2017policygradientmethodsreinforcement,
      title={Policy Gradient Methods for Reinforcement Learning with Function Approximation and Action-Dependent Baselines}, 
      author={Philip S. Thomas and Emma Brunskill},
      year={2017},
      eprint={1706.06643},
      archivePrefix={arXiv},
      primaryClass={cs.AI},
      url={https://arxiv.org/abs/1706.06643}, 
}

@misc{chen2026chartr1chainofthoughtsupervisionreinforcement,
      title={Chart-R1: Chain-of-Thought Supervision and Reinforcement for Advanced Chart Reasoner}, 
      author={Lei Chen and Xuanle Zhao and Zhixiong Zeng and Jing Huang and Yufeng Zhong and Lin Ma},
      year={2026},
      eprint={2507.15509},
      archivePrefix={arXiv},
      primaryClass={cs.AI},
      url={https://arxiv.org/abs/2507.15509}, 
}

@misc{movva2025enhancingscientificvisualquestion,
      title={Enhancing Scientific Visual Question Answering through Multimodal Reasoning and Ensemble Modeling}, 
      author={Prahitha Movva and Naga Harshita Marupaka},
      year={2025},
      eprint={2507.06183},
      archivePrefix={arXiv},
      primaryClass={cs.CV},
      url={https://arxiv.org/abs/2507.06183}, 
}

@misc{yang2025effectivetrainingdatasynthesis,
      title={Effective Training Data Synthesis for Improving MLLM Chart Understanding}, 
      author={Yuwei Yang and Zeyu Zhang and Yunzhong Hou and Zhuowan Li and Gaowen Liu and Ali Payani and Yuan-Sen Ting and Liang Zheng},
      year={2025},
      eprint={2508.06492},
      archivePrefix={arXiv},
      primaryClass={cs.CV},
      url={https://arxiv.org/abs/2508.06492}, 
}

@article{Williams1992,
  author   = {Williams, Ronald J.},
  title    = {Simple statistical gradient-following algorithms for connectionist reinforcement learning},
  journal  = {Machine Learning},
  year     = {1992},
  month    = {May},
  volume   = {8},
  number   = {3},
  pages    = {229--256},
  issn     = {1573-0565},
  doi      = {10.1007/BF00992696},
  url      = {https://doi.org/10.1007/BF00992696}
}

@misc{he2026activezeroselfevolvingvisionlanguage,
      title={Active Zero: Self-Evolving Vision-Language Models through Active Environment Exploration}, 
      author={Jinghan He and Junfeng Fang and Feng Xiong and Zijun Yao and Fei Shen and Haiyun Guo and Jinqiao Wang and Tat-Seng Chua},
      year={2026},
      eprint={2602.11241},
      archivePrefix={arXiv},
      primaryClass={cs.CV},
      url={https://arxiv.org/abs/2602.11241}, 
}

@misc{liu2023visualinstructiontuning,
      title={Visual Instruction Tuning}, 
      author={Haotian Liu and Chunyuan Li and Qingyang Wu and Yong Jae Lee},
      year={2023},
      eprint={2304.08485},
      archivePrefix={arXiv},
      primaryClass={cs.CV},
      url={https://arxiv.org/abs/2304.08485}, 
}

@misc{chen2023sharegpt4vimprovinglargemultimodal,
      title={ShareGPT4V: Improving Large Multi-Modal Models with Better Captions}, 
      author={Lin Chen and Jinsong Li and Xiaoyi Dong and Pan Zhang and Conghui He and Jiaqi Wang and Feng Zhao and Dahua Lin},
      year={2023},
      eprint={2311.12793},
      archivePrefix={arXiv},
      primaryClass={cs.CV},
      url={https://arxiv.org/abs/2311.12793}, 
}

@misc{openai_gpt5_2025,
  title = {GPT-5 Technical Overview},
  author = {{OpenAI}},
  year = {2025},
  howpublished = {\url{https://openai.com}},
  note = {Accessed: 2026-05}
}

@misc{fortin2025gemini25flashimage,
  title        = {Introducing Gemini 2.5 Flash Image, our state-of-the-art image model},
  author       = {Fortin, Alisa and Vernade, Guillaume and Kampf, Kat and Reshi, Ammaar},
  year         = {2025},
  month        = aug,
  howpublished = {\url{https://developers.googleblog.com/en/introducing-gemini-2-5-flash-image/}},
  note         = {Accessed: 2026-05-06}
}
